\documentclass[a4paper,onesided,12pt]{report}

\usepackage[utf8x]{inputenc} 

\usepackage{amsmath, amsthm, amssymb}
\usepackage[bottom]{footmisc}
\usepackage{cite}
\usepackage{graphicx}
\usepackage{longtable}
\graphicspath{{figures/}} 

\usepackage{caption}
\providecommand{\shortcite}[1]{\cite{#1}}
\usepackage{textcomp}

\usepackage{booktabs}
\usepackage{mwe}

\makeatletter

\newcommand*\PrintSkips[1]{%
  \typeout{In #1:}%
  \typeout{\@spaces above: \the\abovecaptionskip}%
  \typeout{\@spaces below: \the\belowcaptionskip}%
}

\makeatother

\usepackage{multirow}
\usepackage[T1]{fontenc}

\newlength\mylength
\usepackage{array}

\usepackage{rotating}
\usepackage{pbox}
\usepackage{graphicx}
\usepackage{verbatim}
\usepackage{booktabs}

\def\BibTeX{{\rm B\kern-.05em{\sc i\kern-.025em b}\kern-.08em
T\kern-.1667em\lower.7ex\hbox{E}\kern-.125emX}}

\usepackage{array}
\newcolumntype{P}[1]{>{\centering\arraybackslash}p{#1}}
\usepackage{multirow}
\usepackage{multicol}
\usepackage{pbox}
\def\arrvline{\hfil\kern\arraycolsep\vline\kern-\arraycolsep\hfilneg}

\usepackage{graphicx}
\graphicspath{ {images/} }
\usepackage{amsmath}

\newcolumntype{P}[1]{>{\centering\arraybackslash}p{#1}}

\usepackage{subfigure}
\usepackage{algorithm}
\usepackage{algorithmicx}
\usepackage[noend]{algpseudocode}
\usepackage{styles/fbe_tez}

\title{DEVELOPING A COMPREHENSIVE FRAMEWORK FOR SENTIMENT ANALYSIS IN TURKISH}
\turkcebaslik{TÜRKÇE İÇİN KAPSAMLI BİR DUYGU ANALİZİ ÇATISI GELİŞTİRME}
\degree{B.S., Computer Engineering, Bahçeşehir University, 2011\\
	M.S., Computer Engineering, Boğaziçi University, 2013}
\author{Cem Rıfkı Aydın}
\program{Computer Engineering}
\subyear{2020}

\supervisor{Prof. Tunga Güngör}
\examineri{Assist. Prof. Tevfik Aytekin}
\examinerii{Prof. Fikret S. Gürgen}
\examineriii{Assoc. Prof. Arzucan Özgür}
\examineriv{Assist. Prof. Reyyan Yeniterzi}

\dateofapproval{28.07.2020}

\begin{document}
\sloppy
\pagenumbering{roman}
\makephdtitle 
\makeapprovalpage

\begin{acknowledgements}
I am more than grateful to my splendid supervisor, Tunga Güngör for his great contributions to this thesis, approaching me always with caring, support, encouragement and understanding, and invaluable guidance in both my M.Sc. and Ph.D. studies, and in my life overall.

I would like to thank the members of my thesis committee, Fikret Gürgen, Tevfik Aytekin, Arzucan Özgür, and Reyyan Yeniterzi for their invaluable feedbacks and contributions to this work. I would like to also thank my professor Arzucan Özgür again, also Haşim Sak and Cumali Türkmenoğlu for providing me with data and tools.

I would especially like to thank my family members, my mother, father, and sister, without whom I would not have succeeded in anything in life. I also thank my late aunt, my friends, relatives, professors, and every other person, whom I have met and who has contributed to my life in terms of love, sharing, learning, teaching, and making me myself.

This work was supported in part by the Boğaziçi University Research Fund (BAP) under Grant 6980D, and in part by the Turkish Directorate of Strategy and Budget under the TAM Project number 2007K12-873. The work of Cem Rıfkı Aydın was supported by the Scientific and Technological Research Council of Turkey (TÜBİTAK) under Grant BİDEB 2211.

\end{acknowledgements}

\begin{abstract}
In this thesis, we developed a comprehensive framework for sentiment analysis that takes its many aspects into account mainly for Turkish. We have also proposed several approaches specific to sentiment analysis in English only. We have accordingly made five major and three minor contributions. We generated a novel and effective feature set by combining unsupervised, semi-supervised, and supervised metrics. We then fed them as input into classical machine learning methods, and outperformed neural network models for datasets of different genres in both Turkish and English. We created a polarity lexicon with a semi-supervised domain-specific method, which has been the first approach applied for corpora in Turkish. We performed a fine morphological analysis for the sentiment classification task in Turkish by determining the polarities of morphemes. This can be adapted to other morphologically-rich or agglutinative languages as well. We have built a novel neural network architecture, which combines recurrent and recursive neural network models for English. We built novel word embeddings that exploit sentiment, syntactic, semantic, and lexical characteristics for both Turkish and English. We also redefined context windows as subclauses in modelling word representations in English. This can also be applied to other linguistic fields and natural language processing tasks. We have achieved state-of-the-art and significant results for all these original approaches. Our minor contributions include methods related to aspect-based sentiment in Turkish, parameter redefinition in the semi-supervised approach, and aspect term extraction techniques for English. This thesis can be considered the most detailed and comprehensive study made on sentiment analysis in Turkish as of July, 2020. Our work has also contributed to the opinion classification problem in English.
\end{abstract}
\begin{ozet}
Bu çalışmada, Türkçe için geniş kapsamlı duygu analizi çatıları oluşturulmuştur. Ayrıca, İngilizce'de duygu analizine özel bazı yaklaşımlar da geliştirilmiştir. Bu kapsamda bu problemin bir sürü yönü göz önünde bulundurulmuştur. Beş tane ana katkı ve üç tane küçük katkıda bulunulmuştur. Denetimsiz, yarı denetimli ve denetimli yöntemlerle orijinal ve etkili öznitelik kümeleri oluşturulmuştur. Bundan sonra, bu öznitelikler klasik makine öğrenme metotlarına girdi olarak verilmiştir ve Türkçe ile İngilizce dillerindeki veri setleri için, yapay sinir ağlarının performansı geçilmiştir. Türkçe dili için ilk defa yarı denetimli yöntemlerle duygu sözlükleri oluşturulmuştur. Sondan eklemeli Türkçe dili için detaylı bir biçimbilimsel analiz gerçekleştirilmiş ve eklerin duygu skorları tespit edilmiştir. Bu, biçimsel olarak zengin diğer dillere de uygulanabilmektedir. Tekrarlamalı ve yinelemeli sinir ağı modellerini birleştiren özgün bir yapay sinir ağı mimarisi geliştirilmiştir. Duygusal, sözdizimsel, anlamsal ve sözlüksel öznitelikler hesaba katılarak, hem Türkçe hem İngilizce için özgün kelime vektörleri oluşturulmuştur. Ayrıca, bağlam pencereleri yancümlecik olarak belirlenerek İngilizce için kısmen orijinal bir yöntemle kelime vektörleri modellenmiştir. Bu, diğer dilbilimsel alanlara ve doğal dil işleme alanlarına uyarlanabilmektedir. Bütün bu özgün yaklaşımlar için, ölçüt olarak alınan çalışmaların başarı oranları geçilmiştir. Küçük çaplı katkılarımız, Türkçe için yön bazlı duygu analizi, yarı denetimli metot için parametrelerin değiştirilmesi ve İngilizce'de duygu analizi için yorumlarda yönlerin tespit edilmesi olarak belirtilebilir. Bu tez, Temmuz, 2020 itibariyle Türkçe için geliştirilmiş en kapsamlı çalışma olarak atfedilebilir. Bu çalışma aynı zamanda İngilizce'de duygu analizi alanı için önemli katkılarda bulunmuştur.
\end{ozet}
\tableofcontents
\listoffigures
\listoftables
\begin{symbols}
%
\sym{$a_{ij}$}{Aspect $j$ of entity ${i}$}
\sym{$c_s$}{Coefficient of the supervised score}
\sym{$c_u$}{Coefficient of the unsupervised score}
\sym{$cos(.)$}{Cosine similarity}
\sym{$\textbf{E}$}{Edge matrix representing cosine similarities between words}
\sym{$e_i$}{Entity $i$}
\sym{$f_{w,d}$}{Frequency of the word $w$ in document $d$} 
\sym{$g$}{Target constituting of entity and aspect}
\sym{$\mathbf{g}$}{Global propagation factor in the semi-supervised approach}
\sym{$h_k$}{Opinion holder $k$}
\sym{$\textbf{M}$}{Positive pointwise mutual information matrix}
\sym{$\textbf{M}_{i,j}$}{Positive pointwise mutual information between $word_i$ and $word_j$}
\sym{$N$}{Corpus of positive polarity}
\sym{$N'$}{Corpus of negative polarity}
\sym{$\mathbf{P_P}$}{Vector of positive polarity word scores}
\sym{$\mathbf{P_N}$}{Vector of negative polarity word scores}
\sym{$s_{ijkl}$}{Sentiment expressed towards the aspect $j$ of entity $i$ by the person $k$ in the time period $l$}
\sym{$\mathbf{s}$}{Seed polarity vector used in the semi-supervised approach}
\sym{$t_l$}{Time $l$}
\sym{$\mathbf{U}$}{Real or complex unitary matrix of the singular value decomposition}
\sym{$\mathbf{v}$}{Vocabulary}
\sym{$\mathbf{V}$}{Real or complex unitary matrix of the singular value decomposition}

\sym{}{}

\sym{$\boldsymbol{\Sigma}$}{Diagonal matrix of the singular value decomposition}

\end{symbols}

\begin{abbreviations}
\sym{3-feats}{Minimum, mean, and maximum sentiment scores per review}
\sym{AASR}{Aspect-Aware Sentiment Representation}
\sym{ABSA}{Aspect-Based Sentiment Analysis}
\sym{BOW}{Bag-of-Words}
\sym{CML}{Classical Machine Learning}
\sym{CNN}{Convolutional Neural Networks}
\sym{combSC}{Combination Score}
\sym{CRF}{Conditional Random Fields}
\sym{C-RNN}{Convolutional Neural Networks - Recurrent Neural Networks}
\sym{DNNs}{Deep Neural Networks}
\sym{IARM}{Inter-Aspect Relation Modelling}
\sym{idf}{Inverse Document Frequency}
\sym{IR}{Information Retrieval}
\sym{kNN}{k-Nearest Neighbour}
\sym{LDA}{Latent Dirichlet Allocation}
\sym{LS}{Log Scoring}
\sym{LSA}{Latent Semantic Analysis}
\sym{LSTM}{Long Short-Term Memory}
\sym{MA}{Multiple Aspects}
\sym{MWE}{Multi-Word Expression}
\sym{NB}{Naïve Bayes}
\sym{NER}{Named Entity Recognition}
\sym{NLP}{Natural Language Processing}
\sym{OOV}{Out-of-Vocabulary}
\sym{PMI}{Pointwise Mutual Information}
\sym{PPMI}{Positive Pointwise Mutual Information}
\sym{POS}{Part-of-Speech}
\sym{R-CNN}{Recurrent Neural Networks - Convolutional Neural Networks}
\sym{RecNN}{Recursive Neural Networks}
\sym{RNN}{Recurrent Neural Networks}
\sym{RST}{Rhetorical Structure Theory}
\sym{SA}{Single Aspect}
\sym{SC}{Sentiment Score}
\sym{SVD}{Singular Value Decomposition}
\sym{SVM}{Support Vector Machines}
\sym{TDK}{Türk Dil Kurumu (Turkish Language Institution)}
\sym{tf-idf}{Term Frequency—Inverse Document Frequency}
\sym{URL}{Uniform Resource Locator}
\sym{VSM}{Vector Space Model}
\end{abbreviations}

\chapter{INTRODUCTION}
\label{chapter:introduction}
\pagenumbering{arabic}

Sentiment analysis is the field of study, where subjective information in source materials is identified and extracted, referring to the use of natural language processing (NLP), machine learning, text analysis and computational linguistics. This sentiment classification task is heavily employed in many fields, such as marketing, social media analyses, and customer service, making use of reviews. Relevant studies in the literature have mostly been conducted for the sentiment analysis problem in English and other most commonly spoken languages. There is a lack of bulky datasets and sentiment lexicons in Turkish to be useful for this classification problem. Hence, this topic can be considered immature for Turkish and this gap needs to be filled by generating comprehensive datasets and lexicons. Also, developing approaches to generating even more robust and novel machine learning algorithms, and feature sets for this language may be helpful, which can be adapted to other languages and even other NLP problems as well. 

In this thesis, we present several contributions, which make our work one of the most detailed analyses performed on the sentiment classification task for Turkish. Our several approaches are cross-domain and also portable to other languages. We also developed some models specific to sentiment analysis in English only, which may be applied to different languages with minor changes. A brief introductory overview of this domain is given below. We have not modelled all of the below scenarios in our study. Our contributions will be discussed in Section~\ref{section:contributions}.

\section{Sentiment Analysis Applications}

Sentiment analysis has been one of the hottest and most alluring topics in the recent decades in NLP, since it helps us understand and analyse the trends, movements, product likeability, and several other implicit and explicit factors at play. By using the techniques specific to this domain, companies can increase their profits through marketing analyses or even political parties may take actions, as was the case in the Arab Spring \cite{ste:11}. In the past, when people needed to ask an opinion, they used to consult their friends or families. On the other hand, when an organisation needed to gather consumer or public opinions, they resorted to conducting surveys and opinion polls. However, due to the explosive growth of the social media (e.g. Facebook and Twitter) people are mostly making decisions, based on what they come across in the reviews, blogs, and posts. However, extracting sentiments from these kinds of texts manually on a large scale is virtually impossible. This is why automated sentiment analysis techniques have to be developed. While data on the Web is mostly utilised for detecting polarities of the reviews, organisations employ the internal data as well for this task. For example, these can be the data collected from call centres, e-mails sent to them by their customers, or other feedback media, such as surveys. Taking into account the importance of this domain, many start-up companies are built, and several companies, such as Microsoft and Google, have built their own in-house capabilities \cite{liu:12}.

\section{Sentiment Analysis Research}

Sentiment analysis, as an NLP task, has become popular in the recent years, and is also widely being researched in the Web mining and information retrieval (IR) fields of study. This task has in fact started spreading from computer science domain into management sciences. The following subsections represent the overall aspects and characteristics of sentiment analysis.

\subsection{Levels of Analysis}

Different levels in sentiment analysis can be taken into consideration, which are document-level, sentence-level, and entity-levels, in a coarse-to-fine order. In document-level analysis, the whole review, which can consist of one or several sentences, or even several paragraphs, is assigned a sentiment score or a polarity label. On the other hand, in sentence-level analysis, factual information (i.e. objective sentence(s)) is distinguished from the subjective information \cite{wie:99}. In entity-level analysis, aspects in the sentence(s) are assigned sentiment scores or labels. For example, the below sentence is given to exemplify this process:

\begin{center}
``Samsung S20+ is packed with a good screen resolution and an awesome video quality overall. However, I observed that it has a bad call quality.”
\end{center} 

Here, the screen aspect and video quality features are each assigned a positive opinion. However, it is the opposite case for the call quality. Another challenging task is that we can prefer using comparative opinions over regular opinions. In regular opinions, a comment may be made only on an aspect or an entity. On the other hand, in the comparative type, two aspects or entities are compared to each other, where disambiguation may occur. For example, a reviewer may make a comment like: 

\begin{center}
``I have recently bought an iPhone and my girlfriend preferred to buy an Android phone despite my contrary opinions. However, I should confess that the sound quality of her smartphone has turned out to be much better than that of mine.”
\end{center}

Here, the aspect \textit{``Android phone’s sound quality”} is represented by the word \textit{``hers”}, and it is a laborious task to extract that information \cite{jin:06}. Therefore, techniques from different NLP areas can also be employed to help extract the corresponding target aspects and their polarities.

\subsection{The Use of Sentiment Lexicons}

Sentiment lexicons are some of the major resources we may make use of when extracting sentiments from source materials, since mostly unigram words represent opinions. To name a few, \textit{``good''}, \textit{``excellent''}, and \textit{``amazing''} are positive words.  On the other hand, some other words, such as \textit{``terrible''}, \textit{``poor''}, and \textit{``repulsive''} are of negative polarity. Nevertheless, taking into account only unigrams may be considered insufficient, since idioms may represent a sentiment independently of its constituent words, such as \textit{``cost someone an arm and leg''}. Another problematic issue is that some words may express a positive sentiment for a domain and a negative sentiment for some other domains. For example, the word \textit{``suck''} carries a negative sentiment in the sentence \textit{``This camera sucks!”} Nonetheless, the same word is of the opposite polarity (i.e. positive) in the sentence \textit{``This vacuum cleaner sucks well.”} Another word exemplifying this case can be \textit{``loud''}, which may have a positive sentiment in a comment made on a headphone. But it could be of negative polarity in a car review, as in \textit{``The car I have just bought is too loud and has not met my expectations.''}

Another challenging issue related to this domain is as follows. Even if a sentiment word appears in a sentence, this does not exactly mean that it has a sentiment. For example, in the sentence \textit{``Do you know any good laptop I can buy?”}, the reviewer does not express a positive or negative opinion. In the other example \textit{``If you did not like the screen quality of your phone, sell it on eBay and buy this Huawei phone with an impressive AMOLED screen instead.”}, a positive sentiment is expressed about the Huawei smartphone. However, a negative comment is in fact not made on the phone the user is currently using (the owner of the phone may be thinking negatively about it). Thus, interrogative and conditional sentences may bring problems along with it, and much further research is needed in this field.

We can detect the polarity of a review through the words, which appear in sentiment lexicons or training corpora. Additionally, it is also possible that an opinion may express a sentiment without the use of a polarity word. For instance, in the sentence \textit{``This washer uses up a lot of water.”}, a polarity word does not appear, while it clearly has a negative opinion. Also, in \textit{``Most of the people in modern society sustain a life with copy-pasted and learnt emotions instead of innate feelings.”}, a word of negative polarity does not occur. However, in fact, the word \textit{``copy-pasted''} indirectly and implicitly expresses a negative sentiment. Identifying such cases is difficult.


\enlargethispage{-\baselineskip}

Another issue that is prevalently observed in this domain is the use of sarcastic opinions. For instance, in the review \textit{``OMG, I am so impressed with the car I have just bought! It stopped working just in two days.''}, a positive sentiment seems like to be expressed in the first sentence when taking account of the word \textit{``impressed''}. However, we can observe that this is not the case when we look at the second sentence. Sarcasm generally occurs in political discussions, but not as frequently in product or service reviews. This problem is one of the hardest tasks to handle in the sentiment classification task.

\section{Opinion Spamming Detection}

Social media has enabled people to express their ideas and opinions without needing to disclose their true identity. This way of expressing a person's own views is highly valuable, since we may be informed of the ``true” sentiments and what he or she ``really” thinks. However, we should not overlook the fact that people may fake identities so as to make undesirable comments about products. The plausible reasons may be to discredit or promote a target product, service, or organisation. For instance, even fake profiles are being created to defame the products of fierce competitor companies. A profile generated for a mobile phone company can defame the product of another cell phone company in several ways, although most of the customers of the second company can be satisfied with nearly all of the characteristics of this product \cite{jin:08}. This problem can be solved by analysing the posting behaviours of individuals. For example, a reviewer may give a score of 10/10 to almost all the products of a company. The usage of these products may not be closely related to each other (e.g. computers, mobile cell phones, televisions, refrigerators). Also, if the same reviewer gives rather low scores to the products of a competitor company, irrespective of what the product is, it could be considered a suspicious act. Another case would be that in which an individual posts the same (or ``weirdly'', very similar) negative review for all the computer products of a corporation. Even several tactics are followed in generating spam opinions. For example, so as to make a fake comment (of positive or negative polarity) the top comment, fake ratings can also be generated. 

That is, other fake profiles can favour that review and \textit{``find it useful”}. Another way to detect spamming is that in which we analyse how frequently and periodically an individual makes reviews or gives star ratings to the comments of other reviewers. If he or she makes many (specifically similar) comments about the product of a company in a span of a few seconds, that can be thought of as another suspicious behaviour.

\section{This Thesis}

Given the background, in this thesis, we present a comprehensive framework for sentiment analysis in Turkish. We propose several approaches to performing document-level and aspect-level sentiment analysis in Turkish. We perform a novel morphological analysis for this agglutinative language and boost the classification performance. We adapted our several novel methods to English as well and achieved state-of-the-art results for both languages. We have also developed a novel aspect-based polarity detection framework for English only and have outperformed all the participating teams in the most reputable competition on sentiment analysis, which has taken place in 2014. Additionally, we define novel word vector models that uses the sentiment information in the reviews. Apart from these, we also propose a novel rule set for defining context windows, which are not only adaptable to sentiment analysis, but also to other NLP tasks. These are all explained briefly below.

\subsection{Motivations, Research Questions, and Contributions}
\label{section:contributions}

In this work, we mainly addressed the problems in the sentiment classification task for Turkish. Although several studies for sentiment analysis in Turkish handle some aspects of it, these are not as comprehensively taken into account as in English and other widely-used languages. In our study, our main goal has been to fill this gap for Turkish by developing novel approaches and achieve better performances than the existing approaches. We list the main contributions and the corresponding research questions below.

\begin{enumerate}

\item \textit{Combination of Unsupervised and Supervised Features:} In this sub-approach, we generate a feature set which combines unsupervised, semi-supervised, and supervised techniques. We simply assign averaged scores to each word, whereby supervised scores are weighed more heavily compared to semi-supervised scores. Using only these word scores and after generating three polarity scores (mean, maximum, and minimum values) for each review, we feed these features as input into an ensemble classifier. Accordingly, we achieve the best results and even outperform neural network models for several corpora of different genres and languages. This proves that classical machine learning algorithms fed with effective feature engineering techniques can yield better results than deep learning frameworks. Employing semi-supervised scores instead of unsupervised scores and combining them with supervised features yields the best results. 

The two \textit{research questions} we addressed for this sub-approach are as follows. Can we achieve a better performance when combining supervised, semi-supervised, and unsupervised techniques? Is it possible that classical machine learning models fed with effective features perform better than neural network approaches?

\item \textit{Morphological Analysis for Turkish:} Turkish is an agglutinative language, whereby morphemes attached to the root forms of words can also be indicative of polarity in sentiment analysis and other classification tasks. Hence, we performed a fine-grained analysis, in which morphemes are also assigned polarity scores relying on the labels of reviews. These reviews consist of words, which are in surface forms. Incorporating the sentiment information of these morphemes as well helps us boost the performance even more rather than when making use of only the root or surface forms of words. We think that this sub-approach can be adopted for other morphologically-rich and agglutinative languages, and other NLP tasks as well with minor changes. 

The corresponding \textit{research question} is: Does comprehensive morphological analysis contribute to the sentiment classification task compared to using only the root or surface forms of words for morphologically-rich languages?

\item \textit{Combining Recurrent and Recursive Neural Networks for Sentiment Analysis Using Inter-Aspect Relations:} Most of the works employ either recurrent or recursive neural network models when performing aspect-based sentiment analysis (ABSA). Recurrent neural networks (RNN) can capture and represent the temporal information. On the other hand, recursive neural networks (RecNN) can model the grammatical and syntactic information more robustly. However, only a few studies combine these two models in an ensemble form. In this approach, we extend a baseline study, which employs only a recurrent model. Their work performs aspect-based sentiment analysis, whereby aspects are provided in advance. They also model inter-aspect relations. That is, it is assumed that the sentiment of an aspect can affect those of preceding and following aspects. Our approach does not only use this recurrent model, but also incorporates recursive model into the framework. We achieve this by splitting each review into sub-reviews (or subclauses) in an original way and training them separately by utilising the sentiment information. We then incorporate these vectors containing sentiment, grammatical, and syntactic information into the baseline recurrent network. By relying on this ensemble neural network model, we outperform the baseline study with a significant margin for two domains. 

\textit{Research questions} addressed for this approach are as follows. Can we enhance recurrent neural networks by including distant sentiment information? Can recursive and recurrent neural network models be merged in an ensemble framework? If so, would it boost the performance? What would be the reason for an ensemble form to perform better than non-ensemble frameworks of recurrent and recursive neural network models? Can reviews be partitioned into sub-reviews, which contain only the relevant sentiments about the aspect(s)? Which one can be considered a better-performing and robust parsing method for the sentiment classification task, constituency or dependency parsing?

\item \textit{Generating Sentiment and Semantic Embeddings:} Most studies in the literature use off-the-shelf word vectors, such as word2vec \cite{mik:13} or GloVe vectors \cite{pen:14}, when implementing NLP tasks. These embeddings can represent the syntactic and semantic information of tokens. However, these kinds of vectors cannot capture the sentiment information as robustly. Therefore, in this approach, we generate several novel embeddings techniques, which incorporate the sentiment information into the vector models. When we feed these embeddings into a sentiment classifier, we achieve the best performances for several corpora, which are of different genres and in two different languages. We also observe that, in several cases, even employing unsupervised techniques can lead to better performances than when utilising supervised methods in creating word vectors. We empirically show that this original approach is cross-lingual and portable to other languages as well. 

Here, the relevant \textit{research questions} are as follows. Does including the sentiment information in word embeddings lead to better performances rather than when incorporating only co-occurrence statistics? Is it possible for vectors generated by unsupervised techniques to outperform those created with supervised methods in a classification task?

\item \textit{Generating Context Windows for NLP Tasks:} In the literature, context windows are generally defined as sliding windows, which consist of a limited number of words, around the target words. In this approach, we propose a novel rule set for generating sub-sentences as context windows and employ this information when generating word vectors. Accordingly, we define the context window to be subclauses, in which a word occurs, rather than to be a fixed number of words in their left and right contexts. Thus, words in the same subclauses are in general more related to each other compared to those in others. We utilise this context window information when modelling the off-the-shelf embedding algorithms, which are the word2vec and GloVe methods, and then feed these vectors into two classifiers. These two NLP classification tasks are sentiment analysis and spam detection problems. We outperform the baseline approach, which makes use of the sliding windows technique, for these tasks. That is, this approach is adaptable to several embedding generation models and other NLP tasks as well. We think that this window context definition can also be used in other models, such as convolutional neural networks (CNN), in lieu of sliding windows. 

The related \textit{research questions} are as follows. Can we redefine context windows for them to model the sequences more robustly compared to sliding windows of fixed-length? Are the words in the same subclause are syntactically and semantically more similar to each other compared to those in sliding windows, even if they are physically more distant in the first technique? If so, why?

\end{enumerate}

In summary, in this thesis, these are the five main contributions for the sentiment analysis task. These are not only restricted to proposing novel approaches for Turkish and specific domains, some of them are also contributive to languages other than Turkish, are cross-domain, and can even be adapted to other NLP tasks with minor changes. In addition to these five contributions, we can list three minor contributions as follows.

\begin{enumerate}

\item \textit{Generating Polarity Scores Using Search Engine Method and Tweaking the Parameters of a Semi-Supervised Approach:} Polarities of words can be detected by querying search engines. In this respect, using syntactic relations can help identify the sentiment and semantic similarities between words. In a study, seed polarity words are defined in advance and the sentiment scores of other words are detected by their co-occurrence statistics. For example, if the word \textit{``awesome''} co-occurs with the word \textit{``good''} more frequently rather than with the word \textit{''annoying''} in context windows (e.g. sliding windows), this word can be assigned a positive score. Several different syntactic contexts can be defined in this respect. For example, words of the same polarities are more frequently connected to each other with the conjunction \textit{``and''}. In our study, we define several syntactic relations specific to Turkish when querying the search engine and employ them. That is, our contribution in this sub-approach is minor in this regard.

In a baseline study, domain-specific polarities are computed by using statistical information. In their work, researchers define a seed set of words, which are specific to a corpus, and then find the sentiments of other words in a corpus by employing propagation methods. The motivation behind their study is that some words may have opposite polarities for different domains. For example, the word \textit{``unpredictable''} may have a negative sentiment in the automobile domain, as in \textit{``This car has an unpredictable steering.''} On the other hand, the same word may express a positive polarity for movie reviews, as in \textit{``This film's plot was really unpredictable. This totally got me!''} In our approach, we have adopted this method as well and adapted it to Turkish by performing minor changes. To our knowledge, our work is the first to employ this semi-supervised approach for the sentiment  classification task in Turkish. When we change several formulae when applying the propagation technique, we achieved better results for this language. However, apart from it, we have not made a broad contribution to this approach.

\textit{Research question} addressing the two above sub-approaches is: What is the effect of employing sentiment lexicons generated in a semi-supervised manner on the performance of the classifier? 

\item \textit{Aspect Term Extraction Method for English:} In this thesis, we also extracted aspects from raw text in English for sentiment analysis. We generated a set of features to identify whether or not a target word is aspect. Among these features are the GloVe embeddings of words, part-of-speech (POS) vectors, and several others as will be discussed later. When we fed the concatenated forms of these word vectors into a support vector machine (SVM) classifier, we achieved a similar performance as that of the baseline. The features we employed alone are not very novel and our contribution for this sub-approach in this respect can be considered minor.

\item \textit{Aspect-Based Sentiment Analysis in Turkish:} In this sub-approach, we extract aspects from a corpus in Turkish for the sentiment classification task. However, it is mostly the same as in a baseline study \cite{deh:15}. Our minor contributions in this respect are that (1) we take into account negation, amplifiers and downtoners and (2) we rely on sentiment lexicons generated by a semi-supervised approach when assigning polarities to aspects.  These two techniques are not original in their way, since they are adapted to the sentiment classification task in English. However, to the best of our knowledge, they have not been adapted to Turkish thus far in such a fine-grained analysis.

\end{enumerate}

\section{Outline of the Thesis}

The remainder of the thesis is organised as follows. In Chapter~\ref{chapter:basic_concepts}, we give a brief overview on sentiment analysis and its main concepts. In Chapter~\ref{chapter:related-work}, we present and discuss the existing works on different approaches used in the sentiment classification task for Turkish and other languages. We describe the proposed approaches and show the experimental results in Chapter~\ref{chapter:methods-and-experiments}. We also discuss the main contributions of the proposed approaches in the same section. We incorporate these different parts of the thesis work into this same section, since we developed several different comprehensive approaches and we do not want to discuss experiments in a faraway chapter. That is, we include the results of experimentations and their contributions immediately after each corresponding sub-approach. In Chapter~\ref{chapter:conclusion}, we conclude the paper.

\chapter{SENTIMENT ANALYSIS AND ITS BASIC CONCEPTS}
\label{chapter:basic_concepts}

If we analyse a problem by breaking it down into sub-problems, we would understand the relevant topic better with respect to its underlying structure. By doing so, researchers can create a more maintainable, consistent, accurate, and robust framework, which can provide solutions and be enhanced more easily later. Therefore, several major sub-components of the sentiment analysis domain and the main problems existing in the current techniques related to this domain are discussed below.


\section{Opinion Definition}

In order to represent the definition of what is an opinion, the below example will be referred to:


\begin{center}
``(1) I bought this iPhone about six months ago. (2) I definitely loved it. (3) Its screen resolution is fascinating and is what I have been looking for for ages. (4) Its battery life is also long. (5) Although I liked it so much, my sister keeps telling that it is too lousy.”
\end{center}

The important points about this review are as follows:
\begin{enumerate}
\item In the second sentence, a positive sentiment is expressed about the phone overall. In the third sentence, an opinion of positive polarity is stated about its screen resolution. Sentence (4) expresses another positive sentiment towards its battery life. Lastly, in sentence (5), a negative opinion is stated. Hence, the following observation can be made.

\textit{Observation:} An opinion can consist of two components, which are a target \textit{g} and a sentiment \textit{s}, that is, 

\enlargethispage{-\baselineskip}
\enlargethispage{-\baselineskip}

\begin{center}
($g$, $s$),
\end{center}

where \textit{g} can be an entity, or an aspect of an entity, whereas \textit{s} can be the sentiment expressed as negative, neutral or positive, or as a rating score, which typically ranges from 1 through 5 (or 10), in increments of 0.5 or 1. In sentence (2), target is the entity iPhone, whereas in sentence (4), it is an aspect (i.e. battery life) of the entity iPhone.

\item The sentiments in this review are expressed by two people, called opinion holders or opinion sources. In sentences (2-4), the opinion holder is the person who makes this review, whereas in sentence (5), it is the sister of the reviewer.

\item The exact date of this review is stated explicitly (\textit{``six months ago''}) and this information is also leveraged in the sentiment classification task. This important aspect can be taken into account, when keeping track of the trends. Here, we can detect how differently people can react regarding a topic in various periods. For example, reviewers may start making negative reviews on some politicians after some corruption scandals come out, or a wave of demonstrations and civil unrest, such as \textit{``Gezi Parkı''} protests, happen. These all can be used in the context of social and political analyses.

\end{enumerate}

To summarise these, the following definition can be referred to:

\textit{Definition (Opinion):} An opinion is a quadruple consisting of four key components as follows:

\begin{center}
($g$, $s$, $h$, $t$),
\end{center}

\noindent
where \textit{g} stands for target, \textit{s} for sentiment, \textit{h} for the opinion holder, \textit{t} for the time when the opinion was expressed.

Although sentence (3) just makes mention of screen resolution, the entity referred to is \textit{``iPhone’s screen resolution”}. A target can therefore be broken down into an entity and an attribute thereof, and be represented as a pair as follows:

\begin{center}

(iPhone, screen resolution)

\end{center}

Entity can be defined as the target object as a whole and its aspects. A relevant example to this is given above. Aspects are generally considered to have a meronymy relation in the sense that these are parts of the entity. Aspects are also called features. However, it should not be confused with feature concept used in the machine learning domain. In this regard, the opinion definition can be refined as follows:

\begin{center}
($e_i$, $a_{ij}$, $s_{ijkl}$, $h_k$, $t_l$),
\end{center}

\noindent
where the only difference to the previously defined set is that the aspect component is added and the description is more fine-grained. Among these subscripts/modules, \textit{i} corresponds to a specific entity, \textit{j} to an aspect, \textit{k} to an opinion holder, and \textit{l} to a specific time, when the opinion is expressed. For instance, $s_{ijkl}$ stands for the sentiment expressed towards the aspect \textit{j} of the entity \textit{i} by the person \textit{k} in the time period \textit{l}. It is a more sensible notation, since we may be interested in detecting which components of a product are of good quality and liked, and which features are not. For example, many people can make a positive comment on the call quality of Google Pixel’s new smartphone, but they can also express negative sentiments on its camera quality. Hence, the aspect components and their implicit or explicit polarities should be analysed separately. The companies may thereby start ameliorating the disliked or flawed features of their products, based on a finer analysis of the reviewers' comments.

\section{Extraction of the Opinion Information}

First, the opinion quintuples ($e_i$, $a_{ij}$, $s_{ijkl}$, $h_k$, $t_l$) need to be extracted from each review (or called document). The first key component we have to obtain is entity, which can be performed by named entity recognition (NER) or other several techniques \cite{hob:10, moo:05, sar:08}. A problematic scenario here is that an entity name can be written in different ways. For instance,  \textit{``McDonald’s''} can be expressed as \textit{``Mc''}, \textit{``McDo''}, etc. Specific normalisation techniques can be employed in this case, which is another NLP task. Secondly, aspects have to be extracted, where different tools, such as dependency parsers, can also be utilised \cite{hu:04}. The following definitions can be referred to while carrying out the aspect extraction task:

\textit{Definition (explicit aspect):} Those are the aspects that can be expressed by nouns or noun phrases. For instance, \textit{``battery life”} in \textit{``Battery life of this laptop is rather long.”} represents an explicit aspect.

\textit{Definition (implicit aspect):} Those are the aspects, which are not defined by nouns or noun phrases. For example, in \textit{``This camera cannot fit in a pants pocket.”}, when expressing the phrase \textit{``fit in the pocket”}, the reviewer refers to the size of the camera.

Out of the remaining three components ($s_{ijkl}$, $h_k$, $t_l$), opinion holders also play a key role when performing social analyses. For example, the location of the reviewers (a metadatum about the opinion source) may be detected and this information can be used and analysed in the case of political turmoils or other significant events. For instance, when Russia had invaded Crimea in 2014, most of the negative reviews have been made by the Ukrainians, whereas the supportive comments were mostly made by the Russian people. Polarities and their target aspects are considered the main components in the sentiment analysis domain, since the services, organisations, or people may be informed of whether or not people are, in general, dissatisfied with their products or some of their features in general. The below example is given to exemplify these processes:

\begin{center}
``My friend adores his camera and its picture quality.”
\end{center}

The following quintuples are generated from it:

\begin{center}
(Canon, GENERAL, positive, Cem’s friend, 08/05/2020)

(Canon, picture quality, positive, Cem’s friend, 08/05/2020)
\end{center}

Disambiguation occurring here stems from the use of \textit{``its”}, which can be solved by the anaphora resolution task in NLP. That is, we should know that it refers to the Canon camera.

\section{Types of Sentiments}

Opinions, which have thus far been discussed are called regular opinions. Additionally, comparative opinions are another type of opinions. Based on how they are expressed, these can be called explicit or implicit opinions.

\subsection{Regular and Comparative Opinions}

Regular opinions are those, in which a comparison does not exist. The two subtypes are direct and indirect opinions. The first type is easy to handle, since the opinion is directly expressed on the entity or its aspects, as in \textit{``The sound quality is amazing.”} Analysing the latter comparison type is more difficult compared to the former type, since it is not as clearly stated. In the sentence \textit{``After injection, my elbows felt much better.”}, the effect of the injection indirectly made the patient feel better. However, some disambiguation may occur in indirect opinions, as in the sentence \textit{``Since I had headaches, the doctor put me on these drugs.”} Here, there is not a sentiment expressed on the drugs, because these are prescribed after the headaches started happening.

Comparative opinions, as it is clarified by its name itself, express ideas, which represent the similarities or differences between entities or their target aspects. These can be in comparative forms, as in \textit{``better”} or \textit{``worse”}, or superlative, as in \textit{``best”} or \textit{``worst”}. Apart from these, verbs are also sometimes utilised to express these kinds of opinions, as in \textit{``I prefer the books of Nietzsche over those of Jean-Paul Sartre among modern philosophers.”}

\subsection{Explicit and Implicit Opinions}

As mentioned, explicit opinions can be regular and comparative opinions. They clearly express what the reviewer intends to say, as shown in the following examples:

\begin{center}
``Lipton Ice Tea tastes great.”

``Nestea tastes even better than Lipton Ice Tea.”
\end{center}

Implicit (or called implied) opinions imply the sentiments, which are not voiced explicitly. This is exemplified by the following comments:

\begin{center}
``This ultrabook has a long battery life.”

``However, this laptop has a longer battery life than the other one.”
\end{center}

In comparison to the explicit opinions, much less research has been conducted for the implicit opinions \cite{zha:11, gre:09}.

\section{Author and Reader ``Confliction”}

A reviewer may make a positive comment on something, but other readers may perceive it negatively. For instance, in the sentence \textit{``I am so happy that housing prices went down.”}, those prospective readers who are likely to rent or buy an apartment may consider this comment as a positive review. On the other hand, for home sellers, the implied sentiment would be negative. In another example, which is \textit{``Trump is most likely going to be re-elected in 2020, I cannot help being in euphoria!”}, a positive opinion is likely indicated for those who are pro-neoliberal globalisation, but not for social democrats.

This chapter was given to provide readers with a general overview about the sentiment classification task. Among the sub-modules discussed here, we have not taken into account comparative opinions, opinion holder and time components in this thesis. Only regular opinions, entities, aspects, and polarities discussed in this chapter have been leveraged in this thesis. Additionally, we have developed several other approaches, which will be given in Chapter~\ref{chapter:methods-and-experiments} in detail.

\chapter{RELATED WORK}
\label{chapter:related-work}

Since we have performed a comprehensive sentiment analysis, where we take its several aspects into account in this thesis, we discuss the related works in separate subsections based on their topics to make them more readable for the reader. We categorised them with respect to the different aspects of the sentiment analysis problem (their application in different languages, the effects of using different machine learning models, feature sets, etc.). Although opinion holder and time components for the sentiment analysis problem are described and explained in Chapter~\ref{chapter:basic_concepts}, we do not review any study here, which models these scenarios. The reason is that we have not performed and implemented these scenarios in this thesis. This should be noted that, in the following subsections, several related works can be repeatedly discussed and overlap, because they handle more than one component of the quintuple model and subtasks of the sentiment classification domain.

\section{A General Overview of Sentiment Analysis at Word- and Document-Level}

In this section, we discuss related works, which mostly perform sentiment analysis at word- or document-level, and employ machine learning algorithms or simpler statistical methods. Most of these conduct binary or ternary sentiment classification, in which a finer analysis (e.g. aspect-based analysis) is not performed. Most of those studies focus on opinion mining in English.

\subsection{Unsupervised Approaches}
Several studies predict the polarities of sentences or reviews based on the sentiment orientations of words, which are generated through unsupervised techniques. In \shortcite{tur:02}, sentiments of words and phrases are detected by relying on a search engine. Then, sentiment of a document is predicted by averaging all the scores of the words therein. The performance is reported to be better for car reviews (80-84\%) than for movie reviews (66\%). We also adopted the same approach by performing minor changes. We have tried out several different query words and operators, and evaluated this technique on the movie and Twitter datasets in Turkish. It is also reported that making use of static corpora may be more robust and less erratic compared to the use of search engines \cite{tab:06}. We also utilised static corpora for our semi-supervised approach, albeit not for the unsupervised approach. In \cite{hat:97}, a graph of adjective words is generated by employing conjunctional relationships. For example, if two words are connected with the conjunction \textit{``and}'' too frequently, they are assumed to be closer in the graph. Lastly, they perform clustering and assign each word to either positive or negative class in the end.

\subsection{Semi-Supervised Approaches}

Annotators can specify a limited number of seed sentiment words and their effect can be propagated across a graph built of corpus words in a semi-supervised manner. In \cite{ham:16}, several sentiment words specific to each genre are chosen and a graph is generated in the sense that words that co-occur frequently have weightier links, that is, they are closer to each other. The sentiment scores of these seed words propagate through this graph and those nodes who are closer to these seeds are assigned similar polarities. The main advantage of this approach is that domain-specific sentiment lexicons are induced in lieu of generic polarity lexicons, such as SentiWordNet \cite{bac:10}. In \cite{mar:14}, unsupervised and supervised approaches are combined by using polarity lexicons. However, their approach does not generate domain-specific lexicons as we do.

\subsection{Supervised Approaches}

Although there is a large body of work, which relies on the use of unsupervised and semi-supervised approaches, the majority of the related studies employ supervised techniques at document-level. Most of the supervised approaches utilise Boolean and term frequency—inverse document frequency (tf-idf) techniques as features when making use of classical machine learning methods \cite{li:10,yil:14}. Apart from these, the delta tf-idf technique proves to be more effective, since sentiment information is taken into account on a word-basis \cite{mar:09}. In this thesis, we model these scenarios as feature engineering techniques as well, in addition to several others. In \cite{far:13}, three polarity scores are extracted from each review, and these are fed as input into machine learning classifiers. We also employ a similar technique in that we induce the minimum, mean, and maximum sentiment scores from each document. We thereby achieve the best performances with this effective feature set. Reference \cite{san:14} generates a feature set, which consists of character $n$-grams and several other metrics, and empirically show that it solves the data sparsity problem for sentiment analysis for the Twitter datasets. Several other features (e.g. POS tags and the number of emoticons or emojis) are also leveraged in another study \cite{lan:16}.

It is reported that punctuation marks and repeated letters in the reviews boost the strength of polarities of the immediately preceding sentiment \cite{the:12}. In \cite{wan:12}, bigrams are stated to capture modified verbs and nouns. Hence, modelling this scenario produces better results for this classification task compared to the bag-of-words (BOW) metric. Another study \cite{guh:15} takes into account the presence and absence of specific words for aspect-based sentiment classification. For instance, conditional words and the presence of \textit{wh} words are mostly indicative of expressing a negative polarity.

In \cite{jia:17}, a fine-grained sentiment analysis is performed employing the financial microblogs and news headlines datasets. For this domain-specific approach, they employ four feature sets, which are linguistic features, sentiment lexicon features, domain-specific features, and word vectors. They choose punctuation marks, numbers, and metadata as domain-specific features. For example, if a word is preceded by the ``-'' mark, it is indicative of a negative polarity for the finance domain. They achieve the best results when relying on ensemble regression classifiers. We instead only choose a limited number of seed words in this thesis and do not define any domain-specific rules for each review. In \shortcite{sar:18}, authors create sentiment-specific vectors, which are also language-independent. They utilise a supervised approach over the word2vec model layer. Emoticons and emojis are processed in a distant-supervised manner such that reviews are automatically annotated. They outperform the word2vec model by also incorporating this sentiment information into their approach. Sentiment lexicons are induced by propagating the effects of those polarity labels across the graph.

SVMs are generally the best-performing classical machine learning model, since these can avoid the overfitting problem through the use of the kernel trick. The other reason is that these are defined by a convex optimisation problem, in which local minima do not exist \cite{wan:12}. However, deep neural networks (DNNs) have more recently gained much more popularity, since they provide better models in terms of accuracy, efficiency, and flexibility compared to classical machine learning classifiers and regressors. Although it can take a great effort and intense time for a person to manually generate features which are fed into classical machine learning classifiers, DNNs extract features automatically. As inputs, in general, word embeddings, such as word2vec, are employed. In a study \cite{maa:11}, sentiment-aware vectors are generated by using the sentiment information of movie reviews. The word2vec embeddings capture only the syntactic and semantic characteristics of words. Therefore, incorporating sentiment information into those word representations as well is reported to boost the classification performance. We also combine unsupervised, semi-supervised, and supervised features on a word-basis. However, in the corresponding approach, we generate scalar values, not sentiment-aware multi-dimensional embeddings. In \cite{fel:17}, emojis are generated by using a distant supervision technique, whereafter a bi-directional long short-term memory (bi-LSTM) is employed to conduct multi-class sentiment analysis. In the same study, sarcasm is also handled. Reference \shortcite{baz:17} builds a framework, in which message-level and topic-based sentiment analyses in a hybrid form are performed. In their study, a two-layer bi-LSTM is utilised and an attention mechanism is employed over the last layer for the message-level sentiment analysis. In the topic-based classification module, a \textit{``Siamese”} bi-directional LSTM framework is modelled by using context-aware attention. No sentiment lexicons or hand-crafted features are relied on in their study. However, they rank first (tie) in the Subtask A of the SemEval-2017 competition.

\section{Studies on Sentiment Analysis in Turkish}

For the sentiment classification task in Turkish, most of the studies in the literature employ supervised techniques. In \cite{kay:12}, political columns in Turkish are subjected to the classification task, using $n$-gram language model, maximum entropy, naïve Bayes (NB), and SVM approaches. The performances they obtain range from 65\% to 77\%. In another study \cite{cet:13}, active learning is employed for sentiment analysis for the Twitter data in Turkish. They use a similar weighting technique as we have done. In \cite{tur:14}, the researchers translate a polarity lexicon in English into Turkish. They also utilise several supervised methods. In their study, employing presence and absence suffixes in Turkish increases the accuracy of their models. The accuracy they obtain is 89.5\% in their study. Other studies on sentiment analysis in Turkish, which rely on unsupervised or semi-supervised approaches, translate sentiment lexicons in English into Turkish \cite{vur:13} or induce sentiment lexicons using corpus statistics \cite{deh:16}. However, the approach that generates polarities based on the use of search engines \cite{tur:02} has thus far not been implemented for the Turkish language to the best of our knowledge. Our study, in this respect, is the first to use this technique as well. Apart from these, most of the studies in the literature take into account only the root forms of words and a limited number of affixes (e.g. the negation morpheme) for Turkish. Nonetheless, we conduct a fine-grained morphological analysis for sentiment analysis in Turkish and leverage every morpheme attached to the root forms. That is, we generate polarities for each morpheme. Our approach can be adapted to other agglutinative languages and even to other NLP tasks with minor changes.

\section{Morphological Analysis for Sentiment Analysis}

In a study \cite{abd:11}, for the opinion mining problem in Arabic, inflectional and derivational morphemes are employed in addition to the root forms of words. With this comprehensive feature set, they have outperformed the baseline model, which only utilises the stems of words as features. In \cite{jos:10}, authors conduct sentiment analysis in the Hindi language by relying on a three-stage system. In their study, if there is an annotated dataset in this language, they perform classification by using supervised techniques. If not, they translate the reviews into English and conduct the classification task in this language. If none of those two cannot be applied, they induce a sentiment lexicon also by taking into account the morphologically-rich structure of the Hindi language and perform classification by utilising a majority voting scheme. In \cite{yan:15}, when classifying movie reviews, the researchers use only seed tokens and Chinese morphemes that can be mono-syllabic characters. They do not employ any external sentiment resources. None of these studies feed the surface forms of words as input into their classifier models, since they report that this can increase the size of features to a great extent and can thereby hamper the performance. They instead choose a specific subset of morphemes (mostly suffixes) as features in addition to root forms. In our study, we also expand our feature set by taking into account all morphemes at first. However, after detecting the polarities of these morphemes, we keep the most discriminative ones and eliminate those which do not express much of a sentiment.

A sentiment classification framework is developed for the Korean language in \cite{jan:10}. In their work, they employ morphemes, which express sentiment, in addition to the root forms for this morphologically-rich language. They feed the movie reviews as input into the sentiment analysis model and leverage several sentiment lexicons. Additionally, a chunking technique which is based on the dependency relationships between morphemes is used in their study. For example, negation morphemes have an impact on the sentiments of words and morphemes only in the same sub-chunks. Similarly, in our study, we only utilise a subset of suffix set, which consists of the most sentimentally informative morphemes. We also leverage negation-shifters in addition to other contextual features. In \shortcite{med:16}, sentiment classification is conducted for the Sinhala language, which is also a morphologically-rich language. In his study, the researcher analyses the impact of the selection of specific adjectives and adverbs on the classification performance. Several weighting schemes are employed to predict the polarity scores. In another study \cite{med:17}, sentiment lexicons for the Sinhala language are generated. In that study, the researcher models a three-stage scenario as follows. First, a cross-lingual approach is followed such that a sentiment lexicon in English is translated into Sinhala. Second, morphological structures are handled and the lexicon's features are expanded. Third, a graph-based approach is proposed by employing the polarity lexicon in the source language (i.e. English). In this study, the researcher defines rule-based methods, where negations and intensifiers are handled. This study is reported to be adaptable to other morphologically-rich or agglutinative languages as well. In this thesis, we do not translate the sentiment lexicons into Turkish. We instead follow a semi-supervised approach, which requires only a minor effort to be made by researchers. This is adaptable to other languages and domains with minor changes. Here, only a set of seed words specific to a domain has to be determined in advance. We also determine the polarities of morphemes in Turkish by employing two corpora in this language \cite{ayd2:20}.

We summarise the related works, which have been covered up to this point in Table~\ref{table:Sentiment-analysis-general-related-work} with respect to their methods and feature engineering techniques. Here, we include only the settings that lead to the best performances. In this table, except \cite{ham:16}, all the studies conduct the binary sentiment classification task. We show only a subset of the related works in the table, since there is a large overlap between these studies in that most of them employ the same or very similar features and methods. We only intend to make the reader to get an overview here. However, most of these studies are not directly comparable, since these are evaluated on datasets of different genres and in various languages.

\begin{table}[thbp]
	\caption[Summary of the related works covered thus far.]{Summary of the related works covered thus far.}
	\label{table:Sentiment-analysis-general-related-work}
\begin{center}
	\addtocounter{table}{-1}
	\begin{longtable}{|
>{\raggedright\arraybackslash}m{2.5cm}|>{\centering\arraybackslash}m{2.1cm}| 
>{\centering\arraybackslash}m{3cm}|>{\centering\arraybackslash}m{1.8cm}|>{\centering\arraybackslash}m{3.22cm}|}\hline
		\endfirsthead
		\multicolumn{5}{c}{Table \ref{table:Sentiment-analysis-general-related-work}. Summary of the related works covered thus far. (cont.)\vspace{1em}} \\\hline
		\endhead
		\endfoot
		\endlastfoot
		
		\textbf{Study}& \textbf{Language} & \textbf{Techniques} & \textbf{Domain} & \textbf{Performance (\%)}\\\hline
		 
\cite{tur:14}& Turkish& tf-idf + SVM & Movie&89.50\\\hline

\cite{tur:02}& English & Co-occurrence statistics through the use of search engine (unsupervised) & Movie & 66.00\\\hline
                   \cite{ham:16}& English & Domain-specific (semi-supervised) for ternary classification & Sports & 63.10\\\hline
                    \cite{kay:12}& Turkish& $n$-gram + SVM&Politics&77.00\\\hline
                   \cite{mar:09}& English & Delta tf-idf + SVM & Movie & 91.26\\\hline
                    \cite{maa:11}&English& Sentiment-aware vectors + Logistic regression&Movie&88.90\\\hline
                    \cite{abd:11} &Arabic& Sentiment lexicon + Morphological analysis&Newswire&73.43\\\hline
                   \cite{jan:10} &Korean& Morphological analysis + Parsers&Movie&94.75\\

\hline
	\end{longtable}
\end{center}
\end{table}



\section{Aspect-Based Sentiment Analysis}

Polarities are, in general, predicted on a review/document basis. However, it is also possible to predict the labels of aspects. In another scenario, aspects and their corresponding orientations can be identified together in a hybrid approach. Most studies leverage statistical methods, knowledge-based techniques, or their hybrid combinations in this regard \cite{cam:16}.

\enlargethispage{-\baselineskip}

In this section, relevant and recent studies, which conduct the aspect-based sentiment classification task, are discussed. Again, these are given in separate subsections based on different underlying machine learning models and structures. There are two scenarios in this case: 1. Aspects can be extracted from the reviews. 2. Only the polarities of the given aspects can be detected. We model the second scenario comprehensively in this scenario \cite{ayd:20}.

\subsection{Aspect Extraction from Reviews}

In the literature, it is stated that aspects and their sentiments can be learned separately or in a joint model \cite{sch:16}. In \cite{por:16}, aspects are extracted by employing a 7-layer deep CNN model. Features fed as input into the system are word embeddings, POS tags, and also several linguistic patterns. In another study \cite{he:18}, the semantic meanings of target aspects are captured and these model aspect representations. The researchers utilise an LSTM model over their framework, which also takes into account the syntactic information. Reference \cite{ma:18} leverages target-level and sentence-level attention mechanism to predict aspect words from the reviews. They also incorporate common sense knowledge into their framework. In this thesis, we also tried to identify aspects in English reviews. We generate a limited number of features for each word, such as POS embeddings, a Boolean value indicating whether or not the word is a noun, and a few others. We then fed them as input into an SVM classifier. Although we outperformed the simple baseline approach, this approach we developed is not very original and the results are not much significant and state-of-the-art. That is, we cannot say that we have made a significant contribution in terms of the aspect extraction task.

\subsection{Aspect-Based Polarity Detection}

In this section, we review studies, which predict the polarities of the aspects given in advance. That is, most of these studies do not focus on extracting aspect words from reviews. The papers are discussed in separate subsections based on the models and structures used. Since we evaluate our aspect-based sub-approach only on SemEval-2014, Task 4 datasets, we first give an overview about this task.

\subsubsection{SemEval-2014, Task 4 - Aspect Polarity Detection}

Task 4 of the SemEval-2014 competition has four tasks, which are (1) aspect term detection, (2) aspect term polarity prediction, (3) aspect category detection, and (4) aspect category polarity detection \cite{sem:14}. For this approach, we have only focussed on aspect term polarity prediction by evaluating our methods on the laptop and restaurant datasets \cite{pon:14}. 26 teams have participated in this subtask. Both of the two top-performing teams \cite{kir:14, wag:14} generate features, such as parse trees and $n$-grams, and feed them as input into an SVM classifier. In \cite{kir:14}, private sentiment lexicons are used. One of them is an in-domain sentiment lexicon, which is generated through the Amazon laptop reviews. The others are out-of-domain polarity lexicons, comprehensive Twitter sentiment lexicons and three manually-curated lexicons. On the other hand, \cite{wag:14} makes use of publicly available sentiment lexicons. Some of the words in the lexicon are eliminated by the authors. Some words are also manually added to the lexicons, and these are adapted to the restaurant and laptop datasets. In our corresponding approach, we rely on deep neural networks and not on hand-crafted feature engineering techniques. That is, we do not conduct feature engineering tasks for this approach. None of the participating teams develop a novel ensemble neural network framework as we do.

\subsubsection{Recurrent Neural Network Models}

In the sentiment analysis problem, many studies employ recurrent neural network models, since they can capture the sequential model and temporal information effectively. As mentioned, those models can help predict the label of preceding or succeeding words more successfully as compared to other neural model approaches. LSTM and gated recurrent units (GRUs) are the two of the most preferred recurrent neural network models \cite{gol:17}.

The baseline study for this approach \cite{maj:18} conducts the aspect-based sentiment classification task by employing inter-aspect relations. They only predict the labels of the given aspects. That is, they do not detect the aspects separately. To model this scenario, they propagate the effect of sentiments through the text. They average aspect term embeddings in each aspect group and they feed these into a bi-directional GRU model. In our approach, we develop a novel ensemble neural network model, such that recursive and recurrent neural models are combined. Recurrent model as in the baseline is employed to capture temporal knowledge. We enrich it with a recursive neural model, such that syntactic, grammatical, and distant sentiment information are also incorporated into the model.

In \cite{che:17}, a recurrent neural network model with an attention mechanism is relied on for aspect-based sentiment analysis. In that study, the sentiments of target aspects are predicted by employing position-weighted memory and recurrent attention memory. They perform better than the baseline approaches. Reference \cite{arr:17} leverages layer-wise relevance propagation in recurrent neural network models. The authors apply a limited number of rules for the propagation process. They conduct five-class sentiment classification and outperform the baseline methods. In \cite{ma:17}, a novel model is proposed, in which aspect representations are merged with those of their contexts using an attention mechanism. In another study \cite{ara:17}, authors perform sentiment analysis for tweets in Spanish. They generate a hybrid approach such that sentiment information from polarity lexicons are incorporated into a recurrent neural network model (LSTM). We also rely on sentiment lexicons, but not in the same manner. In lieu of using the sentiment lexicon scores in a recurrent neural network, we train constituency and dependency parse trees by using a recursive neural network model. Then, we employ the root vectors of these trees in the ensemble framework. Reference \cite{baz:17} employs a 2-layer bi-directional LSTM network for the message-level sentiment detection task. The researchers boost their performance by using an attention mechanism over the last layer. A Siamese bi-directional model is also used for the detection of topic-level sentiments. Their approach rank first in the SemEval-2017 competition, Task 4 \textit{``Sentiment Analysis in Twitter''}.

\subsubsection{Recursive Neural Network Models}

In contrast to recurrent neural networks, RecNNs can model the syntactic, grammatical, and other relational representations in the text. In \cite{soc:13}, the authors decompose reviews into their chunks using a constituency parser and the Stanford sentiment treebank (SST). Each node in the trees can be assigned a sentiment score, whose value ranges from 0 (very negative) to 4 (very positive) in increments of 1. The values at these nodes are updated while training the recursive tree models. This model is more representative and robust than the BOW or $n$-gram techniques, since it can capture the structural relations effectively (e.g. subclauses can be affected with the conjunctions \textit{``although''} and \textit{``however''}). These can play a role in shifting the sentiment of a phrase, subclause, or the whole sentence. We leverage this model when generating our recursive model with a constituency parser. In \cite{kor:17}, a dependency parser is employed for the detection of sentiments. They propose a model, which is especially adapted to morphologically-rich languages (e.g. Polish and Turkish). Their proposed approach is based on ternary sentiment classification. A joint model co-extracts aspects and opinions in another study \cite{wan:16}. The researchers employ a recursive neural network model and the conditional random fields (CRF) approach. They arrive at state-of-the-art results in this study. They model a scenario, in which discriminative features are employed and sentiment information is propagated between related opinions and aspects through a dependency parser. The authors also make use of hand-crafted feature engineering techniques to have a better performance. Reference \cite{ngu:15} defines a specific set of rules to combine constituency and dependency parsers. They create target-dependent binary phrase dependency trees. The researchers predict the sentiments of aspects by training these ensemble forms of neural models and evaluate their approach on the SemEval datasets. We also combine recurrent and recursive neural network models, but in a different manner. We do not merge the output of the two models. We instead train those models separately and feed the outputs of the recursive sub-model per aspect term group as input into the recurrent sub-model. In advance, we also model a novel scenario, in which a set of rules are used to generate sub-review(s) per review.

\subsubsection{Rhetorical Structure Models}

Although recursive and recurrent neural network models can represent the syntactic, semantic, and sentimental characteristics of a review or the aspects therein, these may function on a sentence-basis only. If a document/review consists of two or more sentences, rhetorical structures can model the scenario for the sentiment classification task more effectively and comprehensively. In these structures, nuclei are considered the main components of the text, whereas the satellites are secondary, contributive to the nuclei. In \cite{hoo:16}, rhetorical structures are employed, where polarities of words are given in advance by relying on sentiment lexicons. In these structures, nucleus spans are weighed more heavily in terms of the sentiment score and they use a genetic algorithm to determine the optimal weights of their model. 

In \shortcite{hee:11}, the authors use a parser called \textit{``Sentence-level PArsing for DiscoursE''} to perform polarity detection at document-level. Several rhetorical relations are reported to have more significance for the sentiment analysis problem. They also state that some specific relations, such as the contrast relation, may shift the overall polarity of the review. Another study \cite{hog:15} models rhetorical structure theory (RST) for sentiment classification by employing sentiment lexicons and propagating polarities across sentences or even paragraphs. They use the \textit{``HIgh-Level Discourse Analyzer''} when conducting the sentiment classification task on a review-basis. In \shortcite{tab:08}, the researchers identify the sentiments of words through pointwise mutual information. For that goal, they query search engines. An SVM classifier is relied on to detect on- and off-topics, after which a weighting scheme is employed. In their study, negation, intensifying, and downtoning are handled as well. A decision tree algorithm is used by them to conduct sentiment classification at document-level in the end.

\subsubsection{Ensemble Models}

In ensemble frameworks, two or more sub-models are combined such that these each compensate for what the others lack. In \cite{van:17}, first, a convolutional neural network approach is employed over the word embedding layer. Accordingly, the context information of words are captured and these are convolved into vectors. Over this layer resides a recursive neural network that uses a constituency parser. In this sub-model, grammatical, syntactic, semantic, and sentimental characteristics are modelled, taking the convolved embeddings as input. They evaluate their methods on the SST. In our approach, we first decompose each review into sub-reviews for each aspect group. We then train the recursive neural network model by relying on constituency and dependency parsers. Lastly, we feed their inputs separately into the recurrent gated recurrent unit model. 

In \cite{yan:19}, a bi-LSTM model is trained and then a CNN model is applied over this layer (R-CNN). Also, the opposite mechanism is leveraged (C-RNN). The authors combine these two methods (i.e. R-CNN and C-RNN) under the name of \textit{``fusion gates''}. That is, both temporal and local features are both taken into account in their study. In another study \cite{min:19}, CNN and LSTM frameworks are separately trained and the authors compute the averaged probability scores of the outputs generated by these sub-models. The cut-off threshold value between positive and negative sentiments was set as 0.5. We do not follow such an approach. We instead use the outputs of recursive neural trees per aspect group as the sentiment embeddings in the recurrent model. Reference \cite{che2:17} proposes a model, in which, first, CNN is applied to capture the local information. Then they employ an LSTM model over this layer. The effect of each word is propagated through the sequence. They state that this approach can be expanded for use in other NLP tasks as well.

Apart from the ensemble frameworks discussed above, some studies model hybrid approaches combining ontology-based learning schemes with deep learning neural networks. A study \cite{wal:19} uses conceptual values. These conceptual values refer to the three cases that are (1) polarity orientation (positive or negative), (2) aspect mention (e.g. \textit{``atmosphere''} is associated with the aspect category/concept \textit{``ambiance''}), or (3) polarity mention (e.g. the word \textit{``cheap''} has a negative connotation in the sentence \textit{``This place has a cheap atmosphere.''}, whereas the opposite applies in \textit{``It has a cheap price.''}). If these conceptual frameworks cannot model the sentiment information robustly, the authors use bi-LSTMs. They can therefore capture the most informative words in the right and left context windows of a target word or phrase. In \cite{mes:19}, a similar approach is followed. A work \cite{mes:20} ameliorates this hybrid model by enriching it with regularisation parameters and several CNN layers.

We summarise these related works on ABSA in Table~\ref{table:ABSA-related-works}. We do not include the studies conducted only on aspect extraction here, because we have not carried out an aspect detection study comprehensively in an original way. However, as mentioned, our model that identifies the polarities of the given aspects is novel. Only some of these works are covered in this table to give the reader an overview. They mostly perform ABSA. They rely on different approaches, such as classical machine learning methods, recursive and recurrent neural network models, RST, ensemble, or hybrid approaches. These all evaluate their methods on corpora in English. However, these results are not directly comparable, since datasets can be of different genres. CML in the table stands for classical machine learning approaches.

\begin{table}[!thbp]
\small
	\caption[Summary of the related works on ABSA.]{Summary of the related works on ABSA.}
	\label{table:ABSA-related-works}

\end{table}
\begin{center}
\small
	\addtocounter{table}{-1}
	\begin{longtable}{|>{\raggedright\arraybackslash}m{3.2cm}|>{\centering\arraybackslash}m{4.08cm}|>{\centering\arraybackslash}m{3.2cm}|>{\centering\arraybackslash}m{3.15cm}|}

                  \hline

		\endfirsthead
		\multicolumn{4}{c}{Table \ref{table:ABSA-related-works}. Summary of the related works on ABSA. (cont.)\vspace{1em}} \\\hline
		\endhead
		\endfoot
		\endlastfoot
		
		\textbf{Study}& \textbf{Approaches} & \textbf{Dataset} & \textbf{Performance (\%)}\\\hline
\cite{kir:14}& \textit{CML: } $n$-gram, public lexicons + SVM& Laptop & 70.48\\\hline
\cite{wag:14}& \textit{CML: }$n$-gram, public lexicons + SVM& Laptop & 70.50\\\hline
\cite{hoo:16}& \textit{RST:} Rhetorical structures + sentiment lexicons & Restaurant & 60.00\\\hline
\cite{tab:08}& \textit{RST:} Rhetorical structures + sentiment lexicons + topic classifier & Movie & 76.00\\\hline

\cite{ngu:15} & \textit{Recursive} neural networks combining dependency and constituency parsers& Restaurant & 66.20\\
                  \hline

\cite{wan:16}& \textit{Recursive }neural networks + CRF & Restaurant & 84.11\\\hline
                   
    \pagebreak
		\textbf{Study}& \textbf{Approaches} & \textbf{Dataset} & \textbf{Performance (\%)}\\\hline

\cite{maj:18} (\textit{baseline})& \textit{Recurrent: }Inter-aspect relations with LSTM& Restaurant & 80.00\\\hline

\cite{baz:17}& \textit{Recurrent: }2-layer LSTM and Siamese bi-LSTM& Twitter & 86.30\\\hline
\cite{yan:19}& \textit{Ensemble:} LSTM + CNN& SST & 85.80\\\hline
\cite{van:17}& \textit{Ensemble:} CNN + recursive neural networks& SST& 90.39\\\hline

\cite{mes:19}& \textit{Hybrid: } Lexicalised domain ontology + neural attention& Restaurant & 86.31\\\hline
\cite{mes:20}& \textit{Hybrid: } Lexicalised domain ontology + regularised neural attention& Restaurant & 87.10\\\hline

	\end{longtable}
\end{center}
\vspace{-2.0cm}
\vskip -19cm 

\section{Word and Document Embeddings for Sentiment Analysis}

In the literature, it is reported that employing dense word vectors is a better choice than utilising sparse embeddings in most of the NLP tasks. Priorly, latent semantic analysis (LSA) was proven to be a robust word representation model. In \cite{tur:10}, this approach is employed to model word vectors that takes into account co-occurrence statistics. These are reported to perform better than the sparse vectors models by a significant margin for many classification tasks \cite{cor:95}. However, more recently developed word embeddings, such as word2vec and GloVe, can model the word representations better. These vectors are constructed through shallow neural networks or co-occurrence statistics. Although a large body of studies, which take into account semantic and sentimental aspects of words, exist for English, this topic is immature for the Turkish language.

In addition, latent Dirichlet allocation (LDA) is employed in a study \cite{ble:03} to generate the mixture of latent topics. However, these modelled scenarios focus on extracting latent topics and the generation of word meanings is ignored. Therefore, more robust embeddings, which model the semantic, syntactic, and sentiment information of words, have more recently been developed as given below.

Several studies \cite{li2:10,lin:09,boy:10} leverage the sentimental characteristics of words when modelling word embeddings. In \cite{maa:11}, a framework relies on both semantic and sentimental aspects in generating word representations. That study employs the star ratings of reviews for the sentiment information. A study \cite{ham:16} induces a sentiment lexicon by leveraging domain-specific co-occurrence statistics. They outperform the word2vec when they prefer the singular value decomposition (SVD) technique over the word2vec model.

A study \cite{ert:17} on generating word embeddings for the sentiment classification task employs hand-crafted features. After they create word vectors, they produce document embeddings and feed these into the SVM classifier to predict the polarities of the whole document (review). We outperform this baseline study by generating more effective word vectors by using both unsupervised and supervised feature sets. In \cite{ye:18}, the researchers encode sentiment information into word vectors employing a hybrid approach, which combines a CNN model and external sentiment lexicons.

Most studies in the literature employ distant supervision techniques to generate sentiment-aware vectors for English. In a study \cite{fel:17}, emojis are leveraged as a sentiment and emotion source, and a bi-LSTM model is relied on to create word vectors. Another study \cite{tan:14} creates sentiment-aware embeddings by using the contextual and sentiment information of words. In that study, emoticons help capture the sentimental aspect of a nearby word. Our embedding models do not use shallow or deep neural networks. We instead employ contextual, lexical, supervised, and unsupervised techniques. However, we outperform neural networks by a significant margin. Our approach is portable to other languages and can be adapted to other domains as well \cite{ayd:19}.

\section{Context Windows for Word Embeddings and Other NLP tasks}
\label{section:context-windows}

Distributional word vector models can represent words by taking into account several factors such as semantics, syntax, and even sentiments. They can encode information more robustly and densely as compared to sparse embeddings \cite{gol:16}. These vectors are fed as input into machine learning approaches in many natural language processing tasks to better capture the grammatical and semantic structures of the texts \cite{col:11, lin:15, zho:15, lam:16}. The two of the most widely-used word vector models are GloVe and word2vec, which employ co-occurrence statistics. The word2vec approach relies on a shallow neural network and negative sampling techniques. On the other hand, the GloVe model generates matrices based on the corpus statistics and matrix factorisation methods.

There are only a few studies focussing on the definition of context windows used in word vector models. Performance for word representation models is affected more by the differences in hyper-parameters rather than the specific algorithm used \cite{lev:15}. An approach developed in a study \cite{lev:14} is very relevant and similar to ours in that they also generate subclauses when modelling word representations. However, the dependency rule set they use when creating these subclauses is different to ours. In \cite{mac:18}, the effects of word embeddings generated by dependency parsers are investigated with respect to functional and domain similarities among other factors. Count-based approaches are generally outperformed by distributional semantic approaches \cite{ben:03}. In \cite{lis:17}, hyper-parameters of context windows (e.g. left or right windows, cross-sentential contexts) are tested and evaluated. A  grid search is performed to find the optimal context window size for the speech recognition task in \cite{rav:18}. In \cite{ayd:20}, sub-reviews from reviews are extracted using a dependency parser for the sentiment classification task. We utilise the same approach to defining subclauses in sentences. However, different from that work, we use the subclauses when training word embedding models. The SVD technique is reported to outperform the word2vec approach \cite{ham:16, ayd:19} in modelling word vectors. We use both GloVe and SVD models by defining context windows as subclauses, which has not previously been applied in exactly the same manner to the best of our knowledge.


\chapter{METHODOLOGIES AND EXPERIMENTATIONS}
\label{chapter:methods-and-experiments}

In this thesis, we have developed many approaches, which are adapted to several aspects of the sentiment classification task. Our methods are also applicable for different languages with minor changes. Therefore, we describe our approaches and discuss the corresponding experimentations and results in separate subsections. 

Before we introduce our approaches, we give detailed explanations on our preprocessing and feature extraction methods employed differently for the Turkish and English languages. We then elaborate on these aspects specific to the following, corresponding approaches when needed. Before conducting the sentiment classification tasks for the two languages, we have to perform two main different preprocessing methods. On one hand, since Turkish is an agglutinative language, we preprocess tokens using different libraries specific to Turkish, taking into account its linguistic patterns. A specific preprocessing approach we combined with a supervised metric, which we call \textit{``partial surface forms''} and will be explained later, is original for sentiment analysis in an agglutinative language. On the other hand, since English is not a morphologically-rich language, the corresponding processes take much less effort. Besides these different preprocessing techniques for these two languages, the several proposed approaches are language-independent and can be adapted to various other languages. Most of our preprocessing techniques and approaches are cross-lingual provided that similar grammatical libraries (morphological analyser, lemmatiser, dependency parser, etc.) exist for a target language. 

Thereafter, we explain the five main proposed approaches in separate sections, since these each have contributions in various aspects. Several of these are broken down into sub-approaches, since they are broad in focus. We discuss the several datasets used for each of these frameworks. A little repetition occurs throughout the explanations of these different approaches for the corpora. There are two reasons why we do not create a separate datasets section as we do for the preprocessing methods. One of them is that we mostly employ different corpora in two different languages and of various genres. Some of these sets are specific to only one aspect of the sentiment classification task, such as ABSA. The other reason is that we again do not want to explain the datasets far away from the described frameworks and aspire to make the narrative more effective and consistent. Here, we do not separate minor and major contributions made by us, since some of these are dependent on each other. The reason is that that would fragment the narration to a great extent and make the whole presentation more incomprehensible.
 
We also elaborate on the hyper-parameters in the respective subsections. In the corresponding experimentations and results subsections, we present detailed analyses of our evaluations. We point out our minor or major contributions. We discuss the pros and cons of our described frameworks and we conjecture how they can be adapted to other corpora and languages, and even to other NLP tasks.

\section{Preprocessing and Feature Extraction Methods}

\label{sec:pre}
Before applying our approaches, we perform several basic preprocessing and feature extraction schemes for datasets in Turkish and English. These are important since, in the sentiment analysis problem, the overall opinions of reviews can be predicted based on sentiment words, which are extracted through these methods. Among these are techniques based on unsupervised or supervised characteristics of raw texts. For example, the term frequency and tf-idf features are popular engineering techniques, which indicate how important a token is with respect to a document. On the other hand, intensifiers and downtoners can increase or decrease the effect of a word they modify. Also, schemes like delta tf-idf exploit the label information of data. In addition to these widely-used metrics, we also develop a novel feature engineering technique, which generates a polarity lexicon for Turkish morphemes only. We explain all of these and others in detail below.

\subsection{Basic Preprocessing Techniques}
The feature set that we employed in our framework for the sentiment classification of reviews is as follows:
\begin{itemize}
 \item \emph{Term frequency:} A token's frequency is indicative about the importance of that token. As in many types of classification tasks, term frequencies are widely utilised in the sentiment analysis problem. A word's frequency is indicative about the importance of this word. We leveraged the tf-idf technique as a variation of term frequency.

 \item \emph{POS tag:} Several part-of-speech tags can be more informative than the other tags for this classification task. For example, adjectives are in general considered to be the key sentiment-bearing words. Nonetheless, all POS tags may be taken into account as well. We made use of both of these scenarios when generating the feature set. Various morphological analysis and disambiguation tools could be made use of for different languages to identify the POS tags of the document tokens.

 \item \emph{Sentiment words and phrases:} As said, adjectives are useful in expressing sentiments. However, verbs (e.g. \textit{``love''}), adverbs (e.g. \textit{``hilariously''}), and out-of-vocabulary (OOV) words (e.g. \textit{``9/10''}), may also function as features. Additionally, we can also employ phrases to express opinions (e.g. \textit{``cost someone an arm and a leg”}). Therefore, we used them all to assess and evaluate how these affect the performance of our framework.

\item \emph{Sentiment shifting:} Negators can lead the sentiment to shift. For instance, in \textit{``I did not like it much.''}, a negative polarity is expressed, although the review has a positive polarity word, such as \textit{``like''}. However, employing the negation word (\textit{``not''}) does not always necessitate the sentiment to change, as in \textit{``I adored not only her elegance, but also her intelligence.''} In such a case, the opinion word \textit{``adore''} is still assigned a positive connotation. The Turkish negators (be it words or suffixes) are employed for opinion shifting.
\enlargethispage{-\baselineskip}
\item \emph{Opinion rules:} Several rules can be applied when defining a set of sentiment features. For example, when \textit{``less''} or \textit{``more''} is used in the text, the opinion word following it can be assigned a different score. Another example is that, if an entity consumes a resource in large quantities, as in \textit{``This washer uses up a lot of water.''}, a negative score would be assigned to the document. On the other hand, if it rather contributes to the production, a positive score would be assigned. We applied a set of intensifiers that cause a relative increase or decrease in the scores of the words.

\end{itemize}

Using all the words except those whose frequency is below a threshold value improved the performance for several cases. However, for a metric (i.e. 3-feats), all words with a non-neutral score contributes to the success rates, no matter what their raw frequencies are. In the feature selection and extraction stages, we performed the preprocessing operations as follows:

\begin{itemize}
\item When spelling a text in the Turkish language, people can type English characters for the corresponding Turkish characters from time to time, such as u (\textit{``guzel''}) for ü (\textit{``güzel''} (beautiful)). We rely on the Zemberek tool \cite{aki:07} for the correction of such characters.
\item Emoticons, such as \textit{``:)''} and \textit{``:(''}, are not filtered out in this study, because these are indicative of polarity. If the constituent characters of these emoticons are repeated in a row, we normalise them in such a way that we do not lose this extra information (e.g. \textit{``:((((''} is normalised as \textit{``:((''}). So as to be able to capture all of these kinds of emoticons, we use a specific set of regular expressions.
\item We filter out all punctuation marks except \textit{``?''}, \textit{``(!)''}, and \textit{``!''}, since these do not affect the overall sentiment of a document. If a commentator uses the exclamation mark(s) repetitively in a row (e.g. \textit{``?!''}), a negative polarity is likely to be intended to be expressed. For example, in \textit{``And you say that she is graceful??!!''}, the commentator emphasises a negative aspect about an entity, which is a woman in this case.
\enlargethispage{-\baselineskip}
\item As in the repetition of punctuation marks, we think that the words, in which a repetition of characters (e.g. \textit{``müthişşşş''} (greatttt)) occurs, are utilised for an emphasis and we boost their polarity scores. Additionally, we increase the scores of uppercased words. Nonetheless, if all the words of a document are in uppercase form, we hypothesise that there is no emphasis for specific tokens and there is no extra information to exploit for a specific case of the opinion mining task.
\item After conducting the above normalisation processes, we additionally use the İTÜ NLP tool \cite{tor:14} to detect other unnormalised tokens which the mentioned methods cannot find. On the other hand, for the English datasets, we tokenise and lemmatise all words using the spaCy library \cite{hon:15}. As mentioned, since English is not an agglutinative language, we do not need to perform a comprehensive morphological analysis.
\item We then use the morphological parsing \cite{sak:08} and disambiguation tools \cite{sak:07}, and obtain the morphological analysis of every token. In the techniques which we employ in this thesis, we take account of three scenarios for the words: (1) root form, (2) surface form, and (3) partial surface form (Section \ref{sec:partial_surface}). Since Turkish is a morphologically-rich and agglutinative language, it is possible that the removing morphemes can impact the performance negatively. We observe the effects of these morphemes in the sentiment classification problem by employing these three different metrics.
\end{itemize} 

In addition to the above preprocessing techniques, we perform intensification, negation handling, and stop word removal when we generate the feature set. The negator feature set for Turkish is defined as follows.

\begin{itemize}
 
\item The word  \textit{``değil''} (not) changes the sentiment/semantic orientation of the word it immediately follows. For example, the polarity of the word \textit{``hoş''} (nice) in the review \textit{``Hoş değil bu şarkı.''} shifts from positive to negative.
\item Several verbal suffixes in Turkish (e.g. \textit{``ma/me''}) can switch the sentimental and semantic orientation of the verbs. For instance, \textit{``Beğendim.''} (I liked.) has a positive sentiment, whereas \textit{``Beğenmedim.''} (I did not like.) is of the opposite orientation (i.e. negative).
\item The word \textit{``Yok.''} (There is not.) makes the preceding word ``absent''. For example, in the text \textit{``Umut yok artık.''} (There is no hope any more.), the effect of the word \textit{``umut''} (hope) is neutralised when followed by the word \textit{``yok''}.
\item The suffixes \textit{``sız'', ``siz'', ``suz''}, and \textit{``süz''} are other negators used in the Turkish language. For instance, \textit{``ümitsiz''} (desperate) has a negated meaning, whose polarity switches from positive to negative.

\end{itemize} 

The morphological parser and disambiguation tools we employ finds out if a negation suffix is attached to a word root. If a negation word or suffix is detected, we simply append an underscore at the end of the roots of these negated words in the classical machine learning methods. For instance, if the word is \textit{``beğenmedim''} (I did not like), the corresponding feature is stated as \textit{``beğen\_''}. In the case that negation words or morphemes appear explicitly in a sequence, the effect of the negation is eliminated. For instance, in \textit{``şekersiz değil''} (it is not sugarless), a negation suffix and negator word occur consecutively. Hence, the negative orientation of the statement is lost. This does not therefore necessitate the negation of the token \textit{``şeker''} (sugar) in this case.

When performing stop word removal, we do not remove some specific words which function as stop words in generic/common domain, but which are also helpful in modifying or identifying opinions. For example, the words \textit{``bayağı''} (quite) and \textit{``çok''} (very) boost the strength of the polarities of the following words. When performing these, the intensifiers are removed, however, the polarity scores of the words they modify are increased. If such intensifiers occur consecutively, as in \textit{``çok çok hoş''} (quite quite nice), the corresponding, modified polarity score is increased even more. Several words have also a heavier impact on the intensification of sentiments as compared to some others. For example, the word \textit{``daha''} (more) has a less impact on the following word compared to the intensifier \textit{``en''} (most). Hence, we take into account different coefficient scores for these modifiers that are to be relied on in the classification phase.
\enlargethispage{-\baselineskip}

The sentiment scores of the words has a range of $(-\infty, \infty)$. When a negator modifies a word, its polarity is multiplied by -1, causing the polarity to shift. When negators appear repetitively in a row, the sentiment score remains the same. We group the intensifiers into several categories. When employing the intensifier \textit{``more’'} and the similar intensifiers (\textit{``very’'}, \textit{``quite''}, etc.), we multiply the sentiment score of the modified word by a score of 1.2. Contrarily, if the modifier \textit{``less''} and similar  downtoners (\textit{``so so''}, \textit{``a little''}, etc.) explicitly appear in a text, the multiplication coefficient is chosen as 0.8. If intensifiers occur consecutively in the text, the multiplicative factor is defined as $1.2^{x}$ for the polarity boosters and $0.8^{x}$ for the downtoners, whereby $x$ is the number of intensifiers or dowtoners which come one after another. In the case of the booster \textit{``most''}, the word's polarity score is multiplied by 1.5. On the other hand, for the downtoner \textit{``least''}, the multiplication factor is set as 0.5. We found out the optimal weight parameters by trying them out empirically on two datasets and observing their impact on the performance. These schemes for both Turkish and English are applied separately in this thesis.

In this thesis, we developed sentiment classification approaches which could be evaluated on corpora of different genres. One of these has reviews using standard spelling (i.e. movie corpus) and the other has a different jargon (i.e. Twitter corpus). Words appearing less than a threshold number are removed to combat the noise problem. For the movie dataset, this threshold value is chosen as 20 and, for the Twitter dataset, the corresponding value is set as five. However, for a scheme generated with a supervised technique (e.g. 3-feats), as will be described later, words with a low frequency can have a high polarity score. Therefore, when we did not eliminate any words for this metric, we boosted the performance. For us to cover the Twitter case, we update the normalisation process further. Uniform resource locators (URL) are filtered out, for these do not have an impact on the sentiments of documents (tweets). Nonetheless, we do not remove hashtags from the tweets, because they are reported to have a contribution to the opinions expressed \cite{dav:10}. This was also supported by our preliminary experiments. We give an example in Table~\ref{preproc}, which shows the effects of preprocessing techniques on a tweet in Turkish. In this example, the roots of the words are relied on. When evaluating our approaches on English datasets, we employ a specific tokeniser built for the Twitter jargon specific to this language only \cite{bir:09}. 

\begin{table}[t]

  \caption[A sample tweet in Turkish and the impacts of each preprocessing technique throughout the pipeline.]{A sample tweet in Turkish and the impacts of each preprocessing technique throughout the pipeline.}
\bigskip
   \begin{center}
    \begin{tabular}{|p{3.8cm}|c|c|}
    \hline
   \textbf{Technique} &\textbf{Text}\\
   \hline
    \multirow{2}{*}{Raw Text}&Cok gusel hareketler degil mi bunlar yaa! :))))\\
    & [very nice move+plr not ? these interjection! :))))]\\
    \hline
    Tokenisation &Cok gusel hareketler degil mi bunlar yaa ! :))))\\
    \hline
    Deasciification&\c{C}ok gusel hareketler de\u{g}il mi bunlar yaa ! :))))\\
    \hline
    \multirow{2}{*}{\pbox{3.38cm}{\relax\ifvmode\fi Further normalisation steps}} & çok güzel hareketler değil mi bunlar ya ! :))\\
    \hline
   \multirow{9}{*}{\pbox{3.5cm}{\relax\ifvmode\fi Morphological parsing and disambiguation}} &\\
   & çok çok[Det] \\
   & güzel güzel[Adj] \\
   & hareketler hareket[Noun]+lAr[A3pl]+[Pnon]+[Nom] \\
   & değil değil[Conj] \\
   & mi mi[Ques]+[Pres]+[A3sg] \\
   & bunlar bu[Pron]+[Demons]+lAr[A3pl]+[Pnon]+[Nom] \\
   & ya ya[Conj] \\
   & ! ![Punc] \\
   & :)) @smiley[:)][Unknown] \\
    \hline
   \pbox{3.88cm}{Negation and intensification handling} &  çok\_güzel hareket\_ mi bu ya ! :))\\
    \hline
    \end{tabular}
  \end{center}
  \label{preproc}
\end{table}

Apart from making use of only unigrams (tokens) as the feature set, we employ patterns and multi-word expressions (MWE) as well that are formed of word $n$-grams, where $n > 1$. We rely on the Turkish official dictionary (Türk Dil Kurumu (TDK) - Turkish Language Institution) to identify multi-word expressions (e.g. idioms). A multi-word term can contribute to the polarity orientation which may not be determined by the constituent words therein. For example, the multi-word expression \textit{``nalları dikmek”} (to kick the bucket, literally \textit{``to raise the horseshoes up in the air”}) is of negative polarity, while its constituent words have neutral orientations. On the other hand, patterns are defined as bigrams, in which a predefined set of rules based on the POS tags, as defined in \cite{tur:02}, is used and applied for two words occurring consecutively. For instance, the pattern rule ``adverb+adjective” describes the scenario, whereby an adverb is followed by an adjective, which is a bigram feature as a whole. As we will explain later, whereas employing MWEs in addition to the unigram features has an impact on the classification success rates, this is not the case for the patterns scheme.

\subsection{Additional Preprocessing Step: Partial Surface Forms}

\label{sec:partial_surface}

This part applies only to the sentiment classification task for the Turkish language. The morphologically-rich, agglutinative structure of the Turkish languages enables us to obtain additional features (i.e. morphemes) when conducting NLP tasks. In the literature, researchers mostly take into account either the the surface forms or root forms of the words appearing in the texts \cite{yil:14}. However, in sentiment analysis, some suffixes are more associated with sentiment information compared to the remaining others. For example, the morpheme \textit{``-cağız''} as in \textit{``zavallıcağız''} (poor man/woman) expresses an opinion of pitying someone. As another example, conditional morphemes in Turkish, such as \textit{``-se/-sa''} as in \textit{``Keşke sevse.''} (I wish he or she loved.), express wishes or sentiments in an indirect way. Therefore, in this study, we employ these suffixes when generating features in addition to the use of the surface and root forms of the words. For the Turkish corpora we relied on in this thesis, the morpheme POS set consists of 114 different tags defined overall.

In this preprocessing method, a two-staged scenario is modelled. First, we compute the sentiment scores of all the text words that are in their surface form, using the delta tf-idf technique (see Eq. \eqref{eq:deltaaug}). After that, we parse and then disambiguate the words utilising the morphological analysis tools \cite{sak:08, sak:07} to extract their suffixes. We assume that a morpheme in a word has the same sentiment score as that word, the polarity score of all morphemes is calculated by averaging the scores of the words, to which they are attached. For example, if the suffix \textit{``-se/-sa''} explicitly appears only in two words in the corpus, whose surface forms are \textit{``çalışsa''} (I wish he or she worked) and \textit{``okusa''} (I wish he or she had an education), the polarity score of the morpheme \textit{``-se/-sa''} is calculated by averaging the delta tf-idf scores of the two mentioned words. Then, we take into account a percentage of the morphemes that has the highest confidence scores. When choosing a specific set of morphemes, we address the following factors:

\begin{enumerate}
\item \textit{Negator morphemes:} Irrespective of their sentiment scores, we do not remove the negator suffixes. They have a significant effect in determining the polarities of words by shifting the sentiments on a word-basis.

\item \textit{Number and type of discriminative morphemes:} When we build the discriminative suffix set, we set the number of the top positive and negative morphemes as the same. When we do not follow an equal polarity distribution for top negative and positive morpheme sets, we observe that the results become biased towards the sentiment of the majority class. When conducting experimentations, we evaluate the effects of using various percentages of top discriminative suffixes, which range from 0\% to 100\% in increments of 10. When this number is 100\%, words are considered to be in their surface forms (i.e. all morphemes are processed). When this is set at 0\%, their root forms are used (i.e. all suffixes but the negation morphemes are filtered out). We thereby perform a comparative morpheme-based analysis.
\item \textit{Root forms:} Only the morphemes which have the least discriminative polarity scores are filtered out in this preprocessing stage. That is, we do not eliminate any word roots no matter what their delta td-idf scores are. We perform a morphological analysis here and the root forms of words are not related to this process.
\item \textit{Processing more than one corpus:} In the training phase, the polarity scores of suffixes are calculated using the training set of the corpus. Apart from this setting, we analyse the impact of using other corpora as well on the performance. In this case, when determining the sentiment scores of morphemes, we rely on the training set of one corpus (e.g. movie dataset) and the whole set of another corpus (e.g. tweet dataset), when inducing a morpheme polarity lexicon for the movie dataset. A morpheme's polarity is calculated by averaging the sentiment scores of that word with respect to the movie and Twitter datasets. The same process can be generalised and applied for three or more datasets as well. As we will empirically show and discuss in the experimentation sections, we have found out that exploiting additional knowledge by relying on other datasets can lead to a good generalisation and this helps combatting overfitting. Root forms of words can have different sentiments across different corpora. However, the sentiment orientations of morphemes are independent of corpora. For example, the suffix \textit{``-cağız''} expresses the same polarity across all the texts written in Turkish, although this does not hold for the word \textit{``unpredictable''} as said.
\end{enumerate}

In the second stage, after the morpheme polarity lexicon is generated, the suffixes which are  defined as weakly discriminative morphemes are stripped off the surface forms and the remaining parts only are processed. In other words, the words are defined as their root forms in addition to top discriminative morphemes attached to them. For example, when the suffix \textit{``-se/-sa''} has a high polarity score, but if the score of the possessive suffix \textit{``-m''} is below a  predefined threshold, then the word \textit{``yapsam''} (if I did/made) is redefined as \textit{``yapsa''} by filtering our the suffix \textit{``-m''}. Hence, the words \textit{``yapsam''} and \textit{``yapsa''} are updated and changed as the same partial form. After the partial surface forms are obtained, we employed them in the proposed methods as in the same way as the root forms and the surface forms. We will empirically show in the corresponding experimentation sections that employing partial surface form outperforms the use of the root and surface form schemes. We attribute it to the reason that partial forms are additionally more informative about the overall sentiment, where several morphemes are not neutral-like. However, we should state that this metric is applicable only if a labelled dataset exists in the related agglutinative language.

\section{Major Contribution \#1: Unsupervised, Semi-Supervised and Supervised Approaches, and Their Combination}

\begin{figure}[!h]

 \vspace{0.5cm} 
\includegraphics[width=\linewidth]{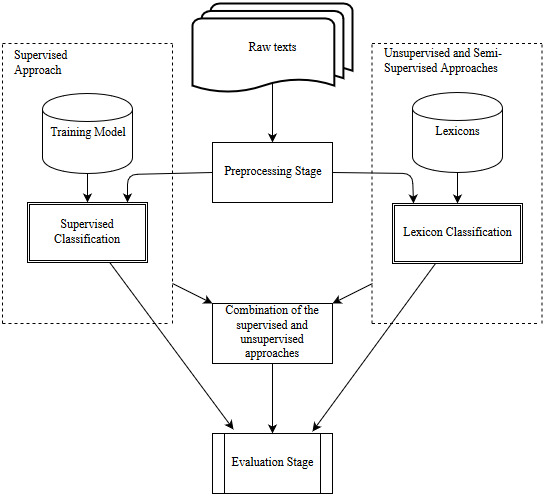} 
\caption{The flowchart of the proposed model.}
 \label{fig:flowchart}
\end{figure}

In this section, we first describe the supervised, semi-supervised, and unsupervised approaches we employed. None of them alone have a major contribution. We only tweak their parameters which can be considered a minor contribution to some degree. However, the way we combine these methods in a novel way can be thought of as a major contribution, especially when thereby achieving significant and state-of-the-art results. The architecture of the proposed approach is visually summarised in Figure~\ref{fig:flowchart}. 

\subsection{Unsupervised and Semi-supervised Approaches}

In the literature, most studies conducted for the sentiment analysis problem in Turkish employ supervised methods \cite{kul:16}. Such works perform many feature extraction and selection methods. Although supervised approaches are diverse and can vary based on their definition, the only unsupervised model used in identifying the sentiments of the Turkish text is the one which refers to the translation of polarity lexicons into their Turkish versions in addition to conducting some further analyses \cite{vur:13, tur:14}. However, translating sentiment lexicons and lexical databases has three drawbacks which we list below:

\begin{enumerate}
\item It requires an intense effort by human annotators. It is also prone to errors.

\item A word that expresses a positive opinion in a language can have the opposite sentiment for another language. For instance, in some parts and countries of Asia, people can make use of the word \textit{``smile''} in the text to cover up embarrassment, shame, or humiliation that is apparently a negative polarity. In Viet Nam, this word is used as a substitute for the expression \textit{``I'm sorry''} or other patterns of behavioural expressions \cite{fon:09}. The word \textit{``smile''} may therefore not be considered to have the same polarity orientation and score in Vietnamese as in English or French, which constitutes as a subjective topic.

\item Besides the cultural aspect, even people who speak the very same language may translate words into their forms in the target language differently. Thus, a word can be labelled different sentiments with respect to different translators.

\end{enumerate}

Because of these drawbacks, in lieu of translating an available sentiment lexicon into its Turkish form, we developed approaches that calculate the polarity scores of the words in an automatic manner without resorting to annotating a bulky set of words manually by humans. We rely on two algorithms, which are domain-independent and domain-specific methods, for this goal. The first model computes a generic polarity score for each word without exploiting any domain-specific knowledge. We construe this approach as an unsupervised technique, because it leverages only a manually-generated small set of seed words with generic polarities. We also develop a domain-specific approach by observing the impact of different domains on the sentiments of corpus words. This is a semi-supervised method, since we manually define the seed words by employing the domain and polarity information. When conducting experimentations, we have observed that the domain-specific methods have performed better than the domain-independent method as we expected.

\subsubsection{Domain-Independent}

\label{sec:dom-ind}

In the domain-independent model, we build a polarity lexicon by calculating a sentiment score (SC) for each word. In this respect, we use the pointwise mutual information (PMI) metric \cite{tur:02}. Given a word \textit{word}, the polarity score is computed in accordance with the formula given below:

\begin{normalsize}
\begin{equation}
\begin{aligned}
\label{eq:1}
SC(word) = \log_2(\frac{hits(word\ NEAR\ `harika\textrm')}{ hits(`harika\textrm')} \times \frac{ hits(`berbat\textrm')}{hits(word\ NEAR\ `berbat\textrm')})
\end{aligned}
\end{equation}
\end{normalsize}

The words \textit{``harika''} (great) and \textit{``berbat''} (awful) are antonyms determined manually. $NEAR$ is an operator that corresponds to the co-occurrence of the words $w_1$ and $w_2$ in the pattern \textit{``$w_1$ NEAR $w_2$”}. That is, this models the statistical scenario that these two words co-occur in sliding context windows. We employ three different operators and patterns in this model, as will be described and discussed in the following paragraphs. \textit{hits(query)} denotes the number of hits returned by a search engine with regard to the given query.

Eq. \eqref{eq:1} is defined by performing division of the PMI formula for the pair \textit{word} and \textit{``harika''} (great) by the PMI formula for the pair \textit{word} and \textit{``berbat''} (awful), and then filtering out the effect of the \textit{hits(word)} term in the denominator in the first part and the \textit{hits(word)} term in the numerator in the second part. In this way, the polarity scores of given words are identified by computing the ratio of the statistics of its co-occurrence with a positive word (e.g. \textit{``harika''} (great)) and its co-occurrence frequency with the predefined negative word (e.g. \textit{``berbat''} (awful)), and then performing normalisation with regard to the total frequencies of the words of this antonym pair. We also conduct smoothing operation by adding a small number (0.001) to the \textit{hits} function to handle the problem of zero-occurrences. The higher the score of Eq. \eqref{eq:1} is, the higher positive sentiment orientation a word has (the more probably it expresses a positive sentiment). If the opposite holds, this is most likely of negative polarity.

Unlike in other relevant studies in Turkish, we compute the hit frequency values through a search engine (Yandex) \cite{yand}, in lieu of relying on manually-curated datasets. We assess and analyse the performance of the search mechanism by employing one operator and additionally two collocational patterns. The operator \textit{``$w_1$ NEAR(k) $w_2$”}, as stated above, returns the number of co-occurrences of the given tokens in the same window of the length $k$. The collocational patterns are \textit{``$w_1$ ve $w_2$” (``$w_1$ and $w_2$”)}, such as \textit{``zeki ve güzel''} (``intelligent and beautiful'') and \textit{``hem $w_1$ hem $w_2$”} (``both $w_1$ and $w_2$”), such as ``hem cin fikirli hem çalışkan'' (``both astute and hard-working''). We take into account these types of conjunctions as operators, for they are used in connecting words, phrases, or clauses with similar sentiment orientations or semantic meanings. For instance, the phrase \textit{``hard-working and beautiful''} is more sensible than \textit{``lazy and beautiful''}. That is, the conjunction \textit{``and''} in general links two words of the same or similar sentiments.

We have found out that the operator $NEAR$ produces more hits and leads to more accurate and robust polarity values than the patterns of collocations. We used several values for the context window length and observed that the optimal window length is $k = 12$ for the operator $NEAR$. In other words, when we look up the six words on both the left and the right of a target word, we achieve the best results. The optimal value was detected by assessing the final performance of the polarity classification model taking account of word polarity scores produced with respect to these various context window lengths. When we increased the value $k$ more, words and phrases of opposite polarities may tend to occur more frequently and the context is assumed to be too broad. By contrast, when we set the context window size as a smaller value, we miss the co-occurrences of the informative collocations of words, which makes the context too narrow and insufficient.

We have built a seed word set consisting of ten antonym pairs that are non-stop words and that are among the most frequently appearing tokens, as we show in the upper part of Table~\ref{posNeg}. The reason that ten antonym pairs as seed words are chosen in this thesis is that the same applies in most studies in the literature as well \cite{deh:16}.  The size preferred could be considered to be small and insufficient. Nonetheless, these antonym pairs are powerful indicators of widely-used and highly representative sentiments. These can capture the implicit and explicit sentiments of the text words effectively and accurately.

\begin{table}[!h]
\vspace{-5mm}
\centering
  \caption{Different seed words of opposite polarities chosen manually for the domain-independent and the domain-specific methods.}
\bigskip
  \begin{minipage}{\textwidth}
    \begin{tabular}{|p{2.55cm}|p{4.9cm}|p{\mylength}|}
    \hline
    \textbf{Technique} & \textbf{Positive}& \textbf{Negative}\\
    \hline
     \multirow{10}{*}{\pbox{1.8cm}{\relax\ifvmode\centering\fi Domain \\ independent}}  & sevgi (love) & nefret (hate) \\ & harika (great) & berbat (awful)\\ & tatlı (sweet) & ac{\i} (painful)\\ & olumlu (positive) & olumsuz (negative)\\ & :) (happy emoticon) & :( (sad emoticon)\\ & güzel (beautiful) & çirkin (ugly)\\ & doğru (correct) & yanlış\ (wrong)\\ & zevkli (enjoyable) & sıkıcı (boring)\\ & iyi (good) &  kötü (bad)\\ & sevimli (lovely) & sevimsiz (unlovely)\\
&&\\
 \hline
     \multirow{10}{*}{\pbox{1.8cm}{\relax\ifvmode\centering\fi Domain \\ specific}}  & sürükleyici (gripping) & yorucu (wearisome) \\ & \c{s}a\c{s}{\i}rt{\i}c{\i} (unpredictable) & öngörülebilir (predictable)\\ & sıradışı (unusual) & kalıplaşmış (clich\'e) \\ & başyapıt (masterpiece) & vasat (mediocre)\\ & büyüleyici (fascinating) & iğrenç (awful)\\ & güzel (beautiful) & çirkin (ugly)\\ & doğru (correct) & yanlış\ (wrong)\\ & zevkli (enjoyable) & sıkıcı (boring)\\ & iyi (good) &  kötü (bad)\\ & sevimli (lovely) & sevimsiz (unlovely)\\
   \hline
    \end{tabular}
  \end{minipage}
  \label{posNeg}
\end{table}

\begin{figure}

\includegraphics[width=\linewidth]{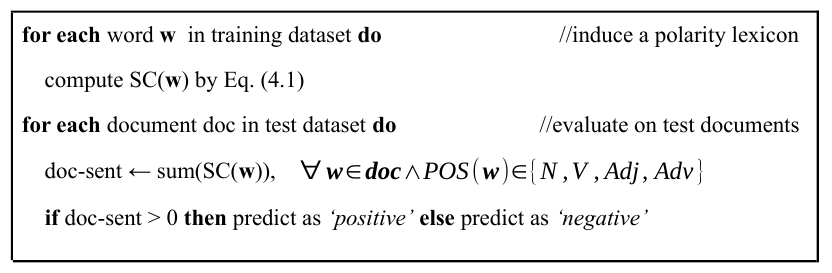} 
\caption{Algorithm for unsupervised approach.}
 \label{fig:algo1}

\end{figure}

We calculate the sentiment values of given words by Eq. \eqref{eq:1} for each antonym words and lastly average these. The reason that we rely on ten antonym pairs and their average, in lieu of a single antonym pair, is to combat the bias that may occur towards a single pair. In other words, we do not want to base our metric on a very specific context. The sentiment score of a document/review is produced by averaging all the polarity scores of words occurring in that text. When the averaged sentiment value exceeds zero, the predicted sentiment is positive. If the opposite holds, the sentiment is predicted as a negative. First, only the words with the part-of-speech tags, which are noun, adjective, verb, and adverb, are taken account of in the reviews. In the preliminary experimentations, we observe that other tokens, such as conjunctions, are not generally indicative of the overall polarity of the document for some approaches. However, some tokens, such as \textit{``10/10''} with the POS tag \textit{``unknown''}, can also contribute to the performance for several modules and datasets (e.g. the 3-feats technique as will be discussed later). This unsupervised method is summarised in Figure~\ref{fig:algo1}.

\subsubsection{Domain-Specific}

\label{sec:dom-spec}

The approach described in the above section predicts the polarity score of tokens without exploiting the domain information of a review. Nevertheless, the sentiment orientation and score of a token can differ based on corpora with different genres. For example, as said, the word \textit{``unpredictable''} has a positive polarity for the movie domain (\textit{``unpredictable plot''}), whereas it has a negative polarity for the reviews made in the car domain (\textit{``unpredictable steering''}). In order to model this domain-specific scenario based on these factors, we adopt the methods described in a study \shortcite{ham:16} for the Turkish language by implementing minor changes in our version. We evaluated this approach on the Twitter and movie datasets. However, this may be adapted to other domains as well after conducting modifications. In this section, we describe and explain the approach for the scenario modelled for the film corpus.

As for the domain-independent method, we define a seed word set consisting of positive and negative antonym pairs. In this case, words are sentimentally specific to the movie corpus. We list these antonym sentiments in the lower part of Table~\ref{posNeg}. As shown, there is an overlap between the seed words defined for the domain-independent and domain-specific scenarios. This is indicative of the fact that some words (e.g. \textit{``doğru''} (correct) and \textit{``yanlış''} (wrong/incorrect)) can be considered to be too specific to and generic/broad for a domain at the same time.

Using those manually chosen seed words, a propagation method is utilised to generate a polarity lexicon for a specific domain/dataset. The underlying mechanism of this approach is that we can hypothesise two words are similar to each other, if they co-occur in the same context windows directly or indirectly. Accordingly, a graph is built such that vertices represent corpus words. On the other hand, edges between nodes represent how frequently the linked words co-occur in sliding windows. If the link's weight is heavier, we assume that those two words are closer to each other in the vector space model (VSM).

So as to compute the weights of these edges, a matrix \textbf{M} whose entries correspond to the positive PMI (PPMI) values is built. These PPMI values (i.e. $\textbf{M}_{i,j}$) are computed as in the following equation, where $w_i$ and $w_j$ are the connected words in the graph.

\begin{equation}
\label{eq:2} 
      \textbf{M}_{i,j}  = max(\log(\frac{p(w_{i}, w_{j})}{p(w_{i}) \times p(w_{j})}), 0)\\
\end{equation} 

In this equation, $p(w)$ refers to the frequency of a word in the sliding window contexts across the whole corpus, whereas $p(w_i,w_j)$  is the probability that those two words co-occur in the sliding windows of a fixed length. We have taken into consideration  several hyper-parameters which are the context window sizes ranging from 10 through 30 in increments of 5. We achieve the highest performance when choosing this value as 15. The max equation in the formula is employed, since co-occurrence statistics are not reasonable when the value is below 0. When building this matrix, a vector per row word $w$ is created. We then compute the edge/link weights of this graph. We calculate the edge weight $\textbf{E}_{i,j}$ between two words by using the cosine similarity metric between these two embeddings.

\begin{figure}[h] 
\begin{center}
\includegraphics[width=\textwidth]{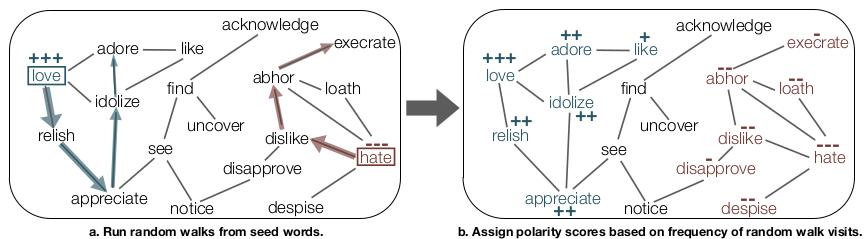} 
\caption{The visual summary of the semi-supervised approach \cite{ham:16}. We tweaked the parameters of this approach and improved the success rates when evaluating the model on Turkish and English datasets.}
 \label{fig:semi-superv}
\end{center}
\end{figure}

While generating this graph, a random walk approach to propagating the effects of sentiments across the corpus words is followed. In this method, if some tokens co-occur with a predetermined seed polarity word frequently and are close to this seed token, these tokens are assigned the same or similar polarities scores. This is visually summarised in Figure~\ref{fig:semi-superv}. The equations are formed of the following notations. $\textbf{v}$ is the vocabulary set of corpus words. The below algorithms identify the positive and negative sentiments independently. $\textbf{P}_N$ and $\textbf{P}_P$ correspond to negative and positive polarity scores, respectively. Before starting to iterate, the ${\textbf{P}_P}^{(0)}$ and ${\textbf{P}_N}^{(0)}$ embeddings are initialised. In this respect, every entry of these two vectors is assigned the score of $\frac{1}{|\textbf{v}|}$. We then update the sentiment score embeddings at each iteration as shown in the below equation. (In this formulation, we only explain the case for the positive class. The analogous process applies for the negative vectors as well.)

\setlength{\abovedisplayskip}{-13.29pt}
\begin{align}
\label{eq:4}
    {\textbf{P}_P}^{(k+1)} = (1 - g)  \textbf{E}  {\textbf{P}_P}^{(k)} + g \textbf{s}
\end{align}

In this formula, $k$ refers to the number of the iteration. \textbf{s} denotes the vector utilised for seed words, whereby the predetermined seed word entries have the weight of $\frac{1}{|\textbf{s}|}$, whereas the other entries have a score of 0. As such, the formula is formed of two constituents. The $\textbf{E}$ matrix holds the co-occurrence statistics information. The vector $\textbf{s}$ is utilised that favours positive seed tokens. The constant \textit{\textbf{g}} is determine the effect of local contextual information. If this has a low score, local information is favoured. Otherwise, global information is given more importance such that seed polarity words have a larger impact on the final sentiments of all words.

While performing the propagation process across the graph, the sentiment values of tokens keep converging to a value at each iteration. This shows that tokens that co-occur with seed polarity words directly or indirectly too frequently have the same or similar polarities. On the other hand, the scores of those words far away from the seed words in the graph are affected and updated to a lesser degree, which probably means that these are neutral-like tokens. Therefore, the absolute polarity values of such words are lower. To combat this bias, we update and increase the local contextual consistency value, where we steadily decrease the \textit{\textbf{g}} value, differently from the baseline approach \cite{ham:16}. The main intuition behind this scenario is that the set of seed polarity words we determined in advance may not be the optimal ones and can be misleading for some corpus words. We perform the iteration until we meet the convergence or meet a maximum number of iterations (i.e. $N$).

Later, we generate the positive sentiment score vector (i.e. $\textbf{P}_P$) as shown below:

\begin{equation}
\label{eq:sumFact}
    \textbf{P}_P = \sum_{k=1}^{N} \frac{{\textbf{P}_P}^{(k)}}{k!}
\end{equation} 

Here, we rely on the vectors built throughout the iterations, and then generate a series and lastly sum these vectors column-wise. In this sequence, at each step $k$, we perform division of these vectors by the factorial value $k$. The reason why we chose the factorial function in lieu of the exponential decay or some other functions is that we wanted to give priority to those words similar to the seed polarity tokens in the VSM. We observed a slight increase of 1\% in the performance by employing Eq. \eqref{eq:sumFact} as compared to the baseline approach \cite{ham:16}. We have thereby shown the effects of tweaking the parameters in the formula. As said, those tokens far away from seed polarity words are assigned more neutral-like sentiments in the end.

Lastly, we generate the polarity score vector which is notated by $\textbf{P}$. The entries of this sentiment vector, each of which is referred to as $w$, are computed by the formulation in Eq. \eqref{eq:pol}. When the score of $\textbf{P}(w)$ exceeds 0, the word is considered a positive token. If the opposite holds, it is considered to be of negative polarity. Different coefficients are employed in Eq. \eqref{eq:pol} than the baseline study \cite{ham:16}. As mentioned, we have therefore observed a small increase in the performance.

\begin{equation}
\label{eq:pol}
    \textbf{P}(w) = \log(\frac{\textbf{P}_P(w)}{\textbf{P}_N(w)})
\end{equation} 

\begin{figure}[!h]

\includegraphics[width=\linewidth]{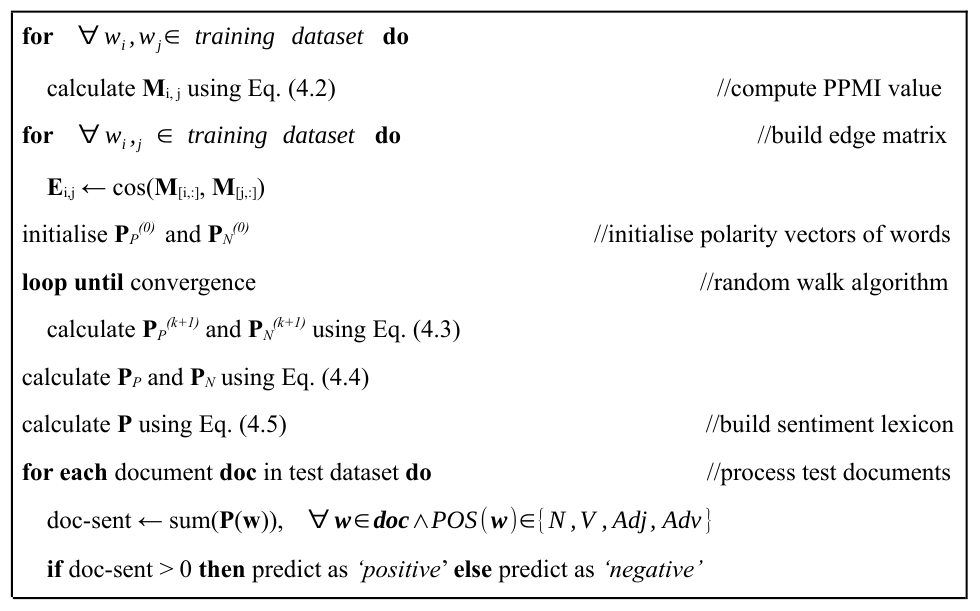} 
\caption{Algorithm for semi-supervised approach.}
 \label{fig:algo2}

\end{figure}

After we identify the polarity scores of each corpus word, we sum all the polarity values in a document. If the summed score of a document is above 0, the overall sentiment is predicted as positive; otherwise, it is predicted as negative. For instance, in Table~\ref{preproc}, polarity scores are found to be \textit{``çok\_güzel'' [+3] and ``:))''} [+2]. The overall summed score is therefore +5. That is, this review is predicted to be positive. We summarise this semi-supervised method in Figure~\ref{fig:algo2}.

\subsection{Supervised Approaches}

We categorise the supervised approaches we utilised in this thesis under two scenarios. In the first case, we make use of conventional machine learning models, such as SVM. We  have initially created several feature engineering techniques and fed these into classifiers. In the latter case, we rely on deep learning models. In this way, we do not need to perform  comprehensive preprocessing, feature selection and extraction operations, since these are mostly handled automatically by these architectures.

\subsubsection{Conventional Machine Learning Models and Feature Sets}

For the classical machine learning architectures, the choice of the ``right'' feature sets rather than classifier models can lead to more successful results. Several feature weighting metrics are accordingly relied on for this approach. The first two schemes are the delta idf and delta tf-idf techniques that are given in Eqs. \eqref{eq:deltaraw} and \eqref{eq:deltaaug}, respectively \cite{mar:09}:

\begin{align}
                            \text{delta idf\textsubscript{w}}  &=\log\frac{\frac{N_{P,w}}{N_P}+0.001}{\frac{N_{N,w}}{N_N}+0.001} \label{eq:deltaraw} \\
                            \text{delta tf-idf\textsubscript{w, d}}  &=(0.5 + 0.5 \times \frac{f_{w,d}}{max_{\{w' \in d\}}f_{w',d}}) \times \text{delta idf\textsubscript{w}} \label{eq:deltaaug} 
\end{align}

 Since these two exploit the sentiment information present in the training dataset, they generally lead to state-of-the-art performances. In \cite{mar:09}, it is stated that this metric performs much better compared to the tf-idf scheme, which does not use the sentiment information. In Eq. \eqref{eq:deltaraw}, delta idf\textsubscript{w} denotes the polarity value of the word $w$ in the dataset. $N_P$ and $N_N$ are the number of positive reviews and negative reviews, respectively. $N_{P,w}$ and $N_{N,w}$ are the counts of documents in the positive and negative dataset, where the word $w$ explicitly occurs, respectively. Numerator and denominator are both normalised to handle the class imbalance problem. For smoothing, we add a value of 0.001 to both numerator and denominator. The delta idf\textsubscript{w} metric is utilised to denote the weight of word $w$, ignoring the frequency statistics on a single review-basis.

Eq. \eqref{eq:deltaaug} is based on a variation of the tf-idf scheme. $f_{w,d}$ denotes the frequency of the word $w$ in the document $d$. We normalise these values such that we divide these frequencies by the most frequently appearing token in the $d$, which is formulated as by $max_{\{w' \in d\}}f_{w',d}$. We thereby prevent a bias towards lengthier documents. delta tf-idf\textsubscript{w, d} takes into consideration both the raw frequency of a word in a document and the supervised score.

The third weighting metric utilised is the classical tf-idf scheme. In this scenario, we do not exploit the sentiment information unlike the above-mentioned ones. We calculate tf-idf scores for the words occurring in both positive and negative datasets. The other metric models the scenario, in which minimum, mean, and maximum word polarity scores calculated by Eq. \eqref{eq:deltaaug} are additionally concatenated on a review-basis, in contrast to those other scores that are built on a word-basis. As will be shown in the results, these three scores in general lead to the best success rates. We also combine the three polarities scheme with the tf-idf metric by performing column-wise concatenation. In other words, unsupervised and supervised techniques are combined in this last scheme.

After building all those features, we used classifier models to predict the polarity labels of documents. The conventional machine learning models employed for this approach are SVM, decision tree (J-48), NB, and k-nearest neighbour (kNN) algorithms. In the literature, generative models are generally stated to be outperformed by discriminative models  \cite{ng:02}. Our experimental results also demonstrate that this holds true, as will be discussed in the results section.

Apart from the machine learning models, we utilise a simple and basic supervised method as well that we name \textit{``log scoring''} (LS). In this method, we only sum the delta tf-idf scores of all the document words. If the overall summed score exceeds the threshold 0, the sentiment is predicted as positive; otherwise, as negative. We have observed that such a basic technique achieves almost as high success rates as the SVM classifier produces. The LS method even performs better than the kNN model.

\subsubsection{Neural Network Approaches}

In addition to conventional machine learning models, we have also utilised two deep learning architectures, which are the CNN and LSTM models. In these approaches, only tokenisation is performed as a preprocessing technique. We do not generate explicit feature sets either for these two networks. As mentioned, these deep learning models can generate features automatically on their own.

\enlargethispage{-\baselineskip}

LSTM is an RNN model consisting of LSTM units \cite{hoc:97}. Every cell in this network ``remembers'' values over arbitrary time intervals and accounts for memory in the architecture. The three gates in these cells are in fact conventional artificial neurons. An activation function is applied taking the weighted sum of the outputs in these gates as input. These architectures are superior over classical RNNs in the sense that these can remember the previous states more effectively and robustly in a long-term dependency manner.

In the LSTM model we relied on, we conduct sentiment classification task on a document-basis. At the most-bottom layer, we feed word embeddings into the architecture. Word embeddings have been trained with the skip-gram setting on a large corpus that consists of 951M words \cite{yil:16}. A shallow neural model is used to generate word vectors as in \cite{mik:13}. This algorithm employs a log-linear classifier for exploiting the statistical information. Contextual windows are accordingly defined and relied on to create dense embeddings. Those words that do not co-occur in the same context windows directly or indirectly are modelled to have less similar representations. The size of those vectors is chosen as 300. The number of LSTM units is selected as 196. So as to prevent overfitting, we determined the dropout rate to be 0.4. The dropout process is performed only in the training phase, not for the development and testing stages. A softmax classifier is employed over the top layer of the architecture. Here, the categorical cross-entropy problem is handled. The maximal number of epochs is set at 200. However, if the model converges or the performance for the validation set starts to decrease after a specific epoch, we choose the optimal model parameters according to the development phase.

Convolutional neural networks have recently gained popularity mostly in the areas of the image and video recognition tasks, and also in NLP tasks \cite{gol:17}. As a conventional feature selection method, the use of $n$-grams with a fixed length may lead to a loss in the information. For example, in two trigrams, if the first two words are the same in both these trigrams, but the last word is different, these two are considered to be completely different, although this should not be the case. Nonetheless, the CNN architectures can handle this issue to some degree. These models capture the most informative local information and use this in the classification stage.

We relied on a publicly available CNN code repository \cite{bri:git-cnn}. We slightly tweaked the parameters of this repository to adapt it to the Turkish languages and to feed off-the-shelf word embeddings into the system. We use the character encoding as UTF-8 and filtered out several tokenisation rules that are specific to English only. We have thereby incorporated several tokenisation and other preprocessing techniques for the Turkish language. So as to feed static word vectors into the model, we have updated the repository with a few lines of code.

The scores obtained in the initial layers are convolved into scalar values by employing several filters with different lengths. First, word embedding values are reduced to single numbers, when these are scanned through sliding windows. We perform max-pooling such that only the most important and maximum values are taken into account. Other parameters, such as dropout regularisation and L2 values among others are employed to combat the overfitting issue and are fine-tuned by the use of a development set. Finally, we perform opinion classification over the last layer, on a fully connected softmax classifier module.

\begin{figure}

\includegraphics[width=\linewidth]{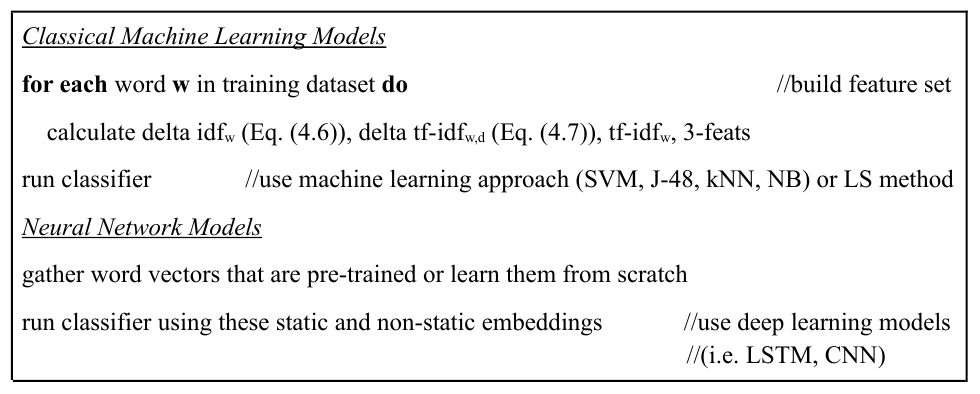} 
\caption{Algorithm for supervised approaches.}
 \label{fig:algo3}

\end{figure}

Non-static word vectors that are learnt from scratch during the training phase and static (pre-trained) word vectors \cite{gun:17} are both employed and tested as input independently. We utilise those word vectors as in the LSTM classifier. We preferred the skip-gram model over the continuous bag-of-words metric for the word2vec approach, since this generates more robust embeddings when a large corpus is available. The length of these pre-trained embeddings is 300. Dense embedding models can combat the data sparsity problem and are more informative as compared to the PMI approach that is employed in the unsupervised and semi-supervised metrics. In other words, exploiting the co-occurrence information about two words cannot be as effective as in employing a neural network model, when fed with a large corpus as input. We summarise all the supervised approaches in Figure~\ref{fig:algo3}.

\subsection{Combining Unsupervised and Supervised Feature Metrics}

Another approach we rely on is the combination of the unsupervised, semi-supervised, and supervised methods, which is a novel and effective feature engineering method developed by us. In this method, if the unsupervised and supervised sentiment values of a token are of different polarity signs (i.e. one positive, the other negative), we consider that word to be sentimentally ambiguous and take into consideration only a fraction of the supervised score. That is, we multiply the supervised score by a specific coefficient. Otherwise, we average these unsupervised and supervised scores, as shown below: 

\begin{small}
\setlength{\abovedisplayskip}{-12pt}
  \begin{equation}
  \label{eq:comb}
    combSC_{w} = 
    \begin{cases}
      c_{s} \times supervised\_score(w), & \text{if}\ opposite\\&\ \ \ signs \\
      c_{u} \times unsupervised\_score(w) + c_{s} \times supervised\_score(w), & \text{otherwise}
    \end{cases}
  \end{equation}
\end{small}

\begin{figure}

\includegraphics[width=\linewidth]{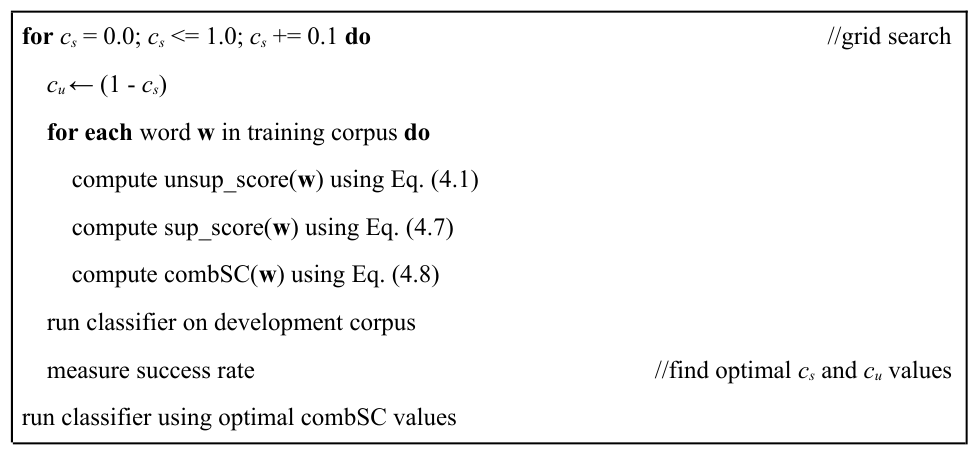} 
\caption{Algorithm for the combination of unsupervised and supervised approaches.}
 \label{fig:algo4}

\end{figure}

In this formulation, the unsupervised score and the supervised score correspond to the scores calculated in, respectively, Eq. \eqref{eq:1} and Eq. \eqref{eq:deltaaug}. For the supervised case, we prefer the delta tf-idf metric (Eq. \eqref{eq:deltaaug}) over other weighting schemes, since it leads to better performances. $c_u$ and $c_s$ are the coefficients of the unsupervised and supervised scores, respectively, where $c_u + c_s = 1$. So as to find the most optimal coefficient values, we first split the whole corpus into training, validation, and test sets. After that, grid search is performed on the development set using nested 10-fold cross-validation, where coefficients can range from 0.1 through 0.9 in increments of 0.1. Another scenario would have been that a supervised regression approach could be followed. Nonetheless, our simple approach was effective, robust, and reliable as well, so we employed our technique. We found out that the optimal values are $c_u = 0.3$ and $c_s = 0.7$ for both datasets. This indicates that, as can be expected, the supervised score is more reliable and informative about a word's polarity, thus it contributes more to the overall score. Nonetheless, the unsupervised weight also has an effect and ignoring it causes the performance to decrease. When the signs of the unsupervised and supervised polarity scores are opposite, we employ only the supervised polarity score multiplied by $c_s$ and the unsupervised component is ignored. The intuition behind is that the supervised score is more reliable; however, due to ambiguity, we lessen its impact. After the feature scores are computed in a combined manner, we feed them into classical machine learning algorithms. As will be discussed later, employing this basic combined approach improves the performance in some cases, especially when we do not perform intensification and negation handling. We give the summary of this approach in Figure~\ref{fig:algo4}.

\subsection{Experimental Evaluation}

We evaluated the proposed approach on text data in two domains for the binary sentiment classification problem in Turkish at document-level. In order to test whether our approaches are portable to other languages, we evaluated the methods on three English corpora as well. In this section, we first give the details of the corpora and the possible values used for the hyper-parameters. Then, we give a detailed explanation on the experiments and make comments on the results obtained.

\subsubsection{Datasets}

\label{sec:datasets-}

In this approach, we first used two datasets of different genres. The first one is a movie dataset consisting of movie reviews in Turkish collected from a popular website \cite{beyaz}. This is the same corpora in the work that we consider as a baseline for comparison \cite{tur:14}. The size of the dataset is 20,244 reviews, where the average number of words in reviews is 39. The reviews have star scores ranging from 0.5 to 5 in increments of 0.5 points. If the overall sentiment score of a review is equal to or lower than 2.5, then we considered it to be negative, whereas if the score is equal to or higher than 4 it is considered as positive. In overall, there are 13,224 positive reviews and 7,020 negative reviews in this movie dataset.

The second corpus is a Twitter dataset formed of tweets in Turkish. The tweets in this corpus are much shorter and noisier compared to the reviews in the movie dataset described above. The training set is composed of 3,000 tweets and was taken from the website of the Kemik NLP group of Yıldız Technical University. Tweets in this set are about two pioneering Turkish mobile network operators, which are Turkcell and Avea. The tweets in the development and test sets which cover various topics were collected and curated by two undergraduate students and were annotated as positive, negative, or neutral. We filtered out the neutral reviews from the dataset since we perform binary sentiment classification only for this approach. We merged and combined all the sets, shuffled them, and applied 10-fold cross-validation on them. We thereby take into account several topics and domains in the same corpus and test how it impacts the performance. In total, there are 1,716 tweets, of which 743 are positive and 973 are negative. The annotators measured the Cohen's Kappa inter-annotator agreement score to be 0.82.

We also utilised three other datasets in English to test the portability of our approaches across languages. One of them is again a movie corpus collected from the web \cite{pan:05}. In this dataset, there are 5,331 positive reviews and 5,331 negative reviews in this dataset. The other is a Twitter dataset containing nearly 1.6 million tweets annotated by using a distant supervised method \cite{go:09}. In the method leveraged, if a tweet includes a positive emoticon, it is labelled as positive and, if it includes a negative emoticon, it is labelled as negative. Otherwise, it is labelled as neutral. That is, these tweets have positive, neutral, and negative labels. We have chosen 7,020 positive tweets and 7,020 negative tweets randomly. As a third dataset, we used the dataset collected for the widely-known SemEval 2017 competition \cite{ros:17} to compare our performance against the contestants. Since we perform binary sentiment analysis, we chose the Subtask B dataset for ``tweet classification according to a two-point scale''. This corpus is formed of tweets including topic information along with sentiment. The training dataset consists of 20,508 tweets, whereas the test dataset is composed of 6,185 tweets. We removed neutral and ``off topic'' reviews from the training set and processed the remaining 18,962 tweets.

Since there are separate training and test datasets for the SemEval task, we have not performed cross-validation for this corpus. We split 20\% of the training dataset as validation set to find the optimal values for the combination of supervised and unsupervised approaches. In the other corpora, in the approach combining the supervised and unsupervised methods described previously, we used nested ten-fold cross-validation. In all the other experiments, we employed non-nested ten-fold cross-validation. We partition the data into three parts: as 80\% for training set, 10\% for development set, and 10\% for test set. The validation portion is used to determine whether the convergence criterion is met, to prevent overfitting, and to find the optimal hyper-parameter values.

\subsubsection{Hyper-parameters}
 
We used the scikit-learn framework \cite{ped:11} for the classical machine learning algorithms (J-48, SVM, kNN, and NB). In SVM, we relied on linear kernel since sentiment analysis is generally considered to be a linearly classifiable problem. For the kNN method, we chose the nearest neighbour number $k$ as three and used the cosine similarity metric. We chose $k$ as an odd number to prevent ties while classifying a review. We label a document as positive or negative, depending on the majority of the sentiments of its nearest neighbours.

In the LSTM framework, we applied only tokenisation and lemmatisation in the preprocessing stage, since the model carries out most of the feature selection and extraction tasks inherently and automatically. In the case of CNN, we chose the number of filters as 128 and set the filter sizes as 3, 4, and 5. In this network, there are four layers, which are the word vector layer, convolutional layer, max-pooling layer, and softmax layer. We set the dropout rate at 0.5 and chose the L2 regularisation value as 0.005. We did not perform negation handling and multi-word expression extractions, since this model is assumed to achieve this effect via the sliding windows. We have trained the network until the convergence criterion is met or for at most 200 epochs, as in the LSTM case. If the loss on the validation set stopped decreasing, we consider the early stopping criterion is met and we cut the training phase at that point.

\subsubsection{Results}

We first show the performances (F1-score) of the five feature weighting schemes in Table~\ref{ML}. Each metric was applied to both datasets using the traditional machine learning methods. We should state that in this experiment, we do not measure the success of a metric on a word-basis. That is, we are not interested in how accurately a metric (say delta idf\textsubscript{w}) determines the polarity of a word. Instead, we measure the performance on a review-basis by employing the polarity scores of the words in the review determined by the metric, and then predicting the review as positive or negative.

In the experiments, we tested the methods using the surface forms of the words, the root forms of the words, and the forms obtained by concatenating the roots with the morphemes having the highest confidence scores (partial surface forms). We found out that using the partial surface forms of words as features produces the best results. We have also observed that, unlike the unsupervised and semi-supervised approaches, using all the tokens gives better success rates than only using words with several specific POS tags (e.g. noun, adjective, verb, adverb). The main reason behind it is that some tokens that are not morphologically correct words (e.g. the token \textit{``10/10''} in a movie review) may carry sentimental information. The figures in Table~\ref{ML} correspond to partial surface form and all tokens cases.

\begin{table}
 \centering
  \caption{Contribution of each weighting scheme to the performances (\%) of the supervised approaches for the two datasets.}
\bigskip
    \begin{tabular}{|l|c|c|c|c|c|c|}
    \hline
    \multicolumn{2}{|l|}{\textbf{Weighting scheme} \& \textbf{Dataset}}& \textbf{J-48}&
     \textbf{SVM}&
     \textbf{kNN}&
     \textbf{NB}&
     \textbf{LS}\\
    \hline
   \multirow{2}{*}{delta idf}& Movie & 89.71&  89.89 & 73.32  &   89.18&  88.32\\
    						& Twitter & 78.75&   78.56&  65.12&  77.00&  75.84\\
    \hline
    \multirow{2}{*}{delta tf-idf}& Movie & 89.92 &   90.01&  74.25&  89.53&  88.64\\ 
    						& Twitter & 78.65&   79.29&  66.26&   78.40& 76.05\\
    \hline
    \multirow{2}{*}{tf-idf}& Movie & 88.82&   89.45&  72.89&   88.87&  -\\
    					& Twitter & 76.38&   77.56&  64.28&   77.01&  -\\
    \hline
    \multirow{2}{*}{3-feats}& Movie & 89.87&   \textbf{90.98}&  74.73&   89.53&  -\\
    					& Twitter & 78.74& \textbf{79.54}&  66.43&   78.48&  -\\
    \hline
    \multirow{2}{*}{tf-idf + 3-feats}& Movie & 89.80&   90.01&  74.00&  89.43&  -\\
    							& Twitter & 78.01&   78.12& 64.54& 77.07& -\\
    \hline
    \end{tabular}
  \label{ML}
\end{table}

We find out that the best success rates were achieved using the 3-feats model for both datasets, which is a simple model that relies on only three features. This is an interesting finding, which can be attributed to the specific nature of the sentiment classification problem. In this problem, the average sentimental meaning of a document accompanied with the maximum and minimum word values may successfully signal the variance of sentiment of the review. That is, it takes into account the most extreme values occurring in a review. If the maximum, minimum, and mean polarity scores of a review differ from each other to a great extent, we would hypothesise that it has ``borderline'' traits. This is akin to how extreme values are captured in the pooling operation of CNNs architecture. For instance, the first reviewer may make a comment like: \textit{``Graphics were hilarious; the plot was OK. However, the actress was beyond terrible.''} On the other hand, the second commentator's review could be \textit{``All the graphics, plot, and actors were OK.''} Although these two reviews might have the same mean polarity score, the variance occurring in the first review is much greater. That is, it can be considered an outlier review, which affects the performance in a different manner.

Table~\ref{MV} shows the results of the unsupervised (Section~\ref{sec:dom-ind}) and semi-supervised (Section~\ref{sec:dom-spec}) approaches, compares them with the supervised case, and also shows the effect of various combination strategies on the success rates. For the supervised classifiers alone, we include the results of the 3-feats method obtained with SVM, which prove to be the most successful ones. All the experimentations were conducted with both the combination of unigrams with MWEs and patterns. We observe that the use of patterns results in worse performance. The success rates for the unsupervised and semi-supervised methods are much lower than those of the supervised learning approaches as can be expected. We tested several combinations of the four supervised methods using majority voting scheme. We achieve the best results with the ensemble combination of J-48, SVM, and NB. The results obtained under this ensemble of classifiers outperformed SVM by a small margin. This result indicates that in some cases the instances misclassified by a classifier are compensated by the other classifiers, and a more robust framework is built. The final result we obtained is the combination of the unsupervised and supervised methods. This combination also using the majority voting scheme yields the highest performances. This signals the necessity and contribution of the unsupervised metric in sentiment classification. The result obtained for the movie corpus (91.17\%) outperforms the baseline study in Turkish \cite{tur:14} by a significant margin, in which an accuracy of 89.5\% is reported to be obtained on the same dataset. Our success rate is also the highest for the binary sentiment classification task in Turkish on the movie dataset, to the best of our knowledge.

\begin{table}[!h]

  \caption{Summary of performances (\%) of the unsupervised, semi-supervised and supervised methods (using the 3-feats technique), and their combinations for the two datasets.}
\bigskip
\begin{center}
    \begin{tabular}{|l|c|c|c|c|}
    \hline
   \textbf{Method} & \multicolumn{2}{c}{\textbf{Unigram + MWE}} & \multicolumn{2}{|c|}{\textbf{Pattern}}\\\cline{2-5}
    & \textbf{Movie} & \textbf{Twitter} & \textbf{Movie }& \textbf{Twitter}\\
   \hline
  
   Unsupervised&70.12&67.82&61.42&59.88\\
   Semi-supervised&72.28&68.59&65.86&61.97\\
   \hline
   J-48&89.87&78.74&84.21&73.28\\
   SVM&90.98&79.54&84.67&73.44\\
   kNN&74.73&66.43&70.16&63.12\\
   NB&89.53&78.48&84.12&73.76\\
    \hline
   Supervised Majority Voting&       \multirow{2}{*}{\pbox{0.8cm}{\relax\ifvmode\centering\fi 91.03}}& \multirow{2}{*}{\pbox{0.8cm}{\relax\ifvmode\centering\fi 80.48}}& \multirow{2}{*}{\pbox{0.8cm}{\relax\ifvmode\centering\fi 85.10}}& \multirow{2}{*}{\pbox{0.8cm}{\relax\ifvmode\centering\fi 74.09}}\\
  (J-48, SVM, NB)&&&&\\
 
   Unsupervised + Supervised Majority&\multirow{2}{*}{\pbox{0.8cm}{\relax\ifvmode\centering\fi \textbf{91.17}}}&
\multirow{2}{*}{\pbox{0.8cm}{\relax\ifvmode\centering\fi \textbf{80.59}}}&
\multirow{2}{*}{\pbox{0.8cm}{\relax\ifvmode\centering\fi 85.36}}&
\multirow{2}{*}{\pbox{0.8cm}{\relax\ifvmode\centering\fi 74.55}}\\
     Voting (J-48, SVM, NB)&&&&\\
    \hline
    \end{tabular}
  \label{MV}
\end{center}
\end{table}

Although the neural network architectures we used are supervised models, we show them separately from the classical supervised models, since they are not subjected to different feature weighting schemes as in Table~\ref{ML}. Table~\ref{Neural} shows the performances of the deep learning architectures. Employing static word vectors (word2vec) results in worse performance compared to the non-static word vectors, which are learnt from scratch. The reason is that in the non-static case we learn the embeddings during the training phase by exploiting the sentiment information. On the other hand, the static vectors are pre-trained, and we do not use class labels in the learning phase of word representations. As in all other experiments, the use of word embeddings and neural models led to lower success rates for the Twitter dataset compared to the movie dataset. This can be attributed to the Twitter's distinctive and noisy jargon, which makes it more difficult to normalise the tweets successfully, and to generate meaningful and comprehensive representations of tweets. There are many OOV words that could not be normalised, thus they lack their corresponding word embeddings. Since there are relatively less OOV words in the movie corpus, we could feed their available embeddings into the network and obtain better performance as compared to the Twitter dataset.

\begin{table}[thbp]
 
  \caption{Performances (\%) for two neural network models using different word embedding types for the two datasets.}
\bigskip
\begin{center}
    \begin{tabular}{|l|c|c|c|}
    \hline
    \textbf{Embedding type}&\textbf{Dataset}&\textbf{LSTM}&\textbf{CNN}\\
   \hline
       \multirow{2}{*}{\pbox{2.8cm}{\relax\ifvmode\fi Static}} &Movie&87.98&89.69\\
	&Twitter&75.84&78.58\\\hline
       \multirow{2}{*}{\pbox{2.8cm}{\relax\ifvmode\fi Non-static}}  &Movie&88.04&\textbf{90.25}\\
	&Twitter&75.86&\textbf{78.74}\\
    \hline
    \end{tabular}
 \end{center}
  \label{Neural}
\end{table}

An interesting point that we observed in the CNN experiments is that when we perform intensification and negation handling, and take account of multi-word expressions, the success rates drop by about 2\%. Since the filtering mechanism within the framework carries out these tasks inherently, our intervention proves to be not only unnecessary, but also harmful. For example, when we perform negation handling (e.g. \textit{``etkileyici değil''} (not impressive) is changed as \textit{``etkileyici\_''} (impressive\_)), SVM generates better results, whereas CNN does not. Thus, some feature engineering techniques may have opposite effects on different machine learning algorithms. 

In the literature, neural network models used in this work are in general reported to outperform classical/traditional learning methods especially when fed with bulky datasets. However, the success rates we obtained using the LSTM and CNN models are not as high as those produced by the supervised methods used, especially with the 3-feats metric and the majority voting scheme (i.e. the ensemble classifier). That is, we found out that conventional machine learning methods could perform better than the two popular neural network models by relying on different and effective feature engineering techniques. For several scenarios, the neural network models could not compensate for the advanced features additionally implemented for the classical machine learning schemes. For example, in the SVM approach, we made use of multi-word expressions, whereas in the neural networks word embeddings for those expressions and phrases were absent. Another case is that we can intensify the polarity strengths of words and feed these modified input vectors into the classical machine learning methods. However, we cannot intensify or modify the values of word embeddings that are generated in \shortcite{yil:16}. However, when we do not employ those feature selection methods, neural networks outperform the classical machine learning methods. We think that this is on account of the advanced nature of the feature extraction processes used in the conventional methods. 

It can be argued that the performance of the neural models will improve if the OOV words are handled properly. There are several ways to decrease the number of OOV words and generate embeddings for such words as discussed in the literature \cite{che:18,gar:18}. For example, the embedding of an OOV word can be taken as the average of the embeddings of words appearing in the same context window(s) \cite{hor:17}. The success rates of the neural models may increase if we include these words in the models. However, since most of the OOV words occur rarely and we remove the words that occur below a frequency threshold for several cases, this may result only in a slight increase in the performance. We leave a detailed analysis of this issue for future study, possibly for post-doctoral researches.

\enlargethispage{-\baselineskip}

We also analyse the contribution of the preprocessing operations on the success rates. We show the effect of each operation in Table~\ref{modules}. We use the delta tf-idf\textsubscript{w, d} metric and SVM. Normalisation refers to the operations of deasciification, punctuation removal, and other tasks provided by the İTÜ normalisation tool. Noise elimination is the removal of tokens occurring less than seven times. In the MWE step, multi-word terms are included in addition to the unigram features. The steps denoted by emoticons, negation handling, and intensification correspond to including also emoticons in the feature set, processing negation words and suffixes, and taking intensifiers into account, respectively. In the partial surface forms step, we use partial forms formed of 90\% and 50\% of the top morphemes for the movie and Twitter datasets, respectively. As shown in the table, it increases the success rates with a significant margin for the movie dataset and with a small margin for the Twitter dataset. The detailed analysis of partial forms will also be given in the next section. Finally, the \textit{``all POS''} step refers to the case where all the text words are processed rather than only the four categories, namely noun, adjective, verb, and adverb, as in the previous steps. We see that each step contributes to the success rate. In the Twitter case, normalisation and removing the POS restriction also boost the performance because of the idiosyncratic nature of the medium and the large number of OOV words. In summary, normalisation, using partial surface forms, and all the tokens regardless of their POS tags are found to be the most effective methods to increase the success rates.

\begin{table}[!h]
\centering
  \caption{Contributions of preprocessing modules to the performance (\%) of the SVM method using the delta tf-idf metric.}
\bigskip
    \begin{tabular}{|m{8.3cm}|>{\centering\arraybackslash}m{1.7cm}|>{\centering\arraybackslash}m{1.7cm}|}
    \hline
    \textbf{Module}&\textbf{Movie Dataset}&\textbf{Twitter Dataset}\\
   \hline
   No normalisation & 83.09 & 59.75\\
    \hline
   Normalisation & 85.88 & 68.21\\
    \hline
   Normalisation + noise elimination & 86.81 & 69.57\\
    \hline
   Normalisation + noise elimination + MWE & 86.98 & 69.71\\
    \hline
   Normalisation + noise elimination + MWE  + emoticons & 87.12 & 70.16\\
    \hline
   Normalisation + noise elimination + MWE  + emoticons + negation handling & 87.41 & 72.67\\
    \hline
   Normalisation + noise elimination + MWE  + emoticons + negation handling + intensification & 87.47 & 75.02\\
    \hline
    Normalisation + noise elimination + MWE  + emoticons + negation handling + intensification + partial surface forms & 89.91 & 76.12\\
    \hline
    Normalisation + noise elimination + MWE  + emoticons + negation handling + intensification + partial surface forms + All POS & \textbf{90.98} & \textbf{79.54}\\
    \hline
    \end{tabular}
  \label{modules}

\end{table}

As a final experiment, we also evaluated the proposed approaches on three datasets in English in order to test the portability of the approaches across languages. To give an overview, we show the most important findings which are produced by a subset of the methods and the feature engineering techniques. We do not apply morphological operations on the words (negation suffixes, partial surface forms, etc.), since English has a simple inflectional, derivational, and morphological structure. The results are given in Table~\ref{eng}. For the SemEval 2017 dataset, in addition to the F1-scores shown in the table, we also computed the ``average recall'' values as stated in the relevant paper in order to reasonably compare our results with those of other participants. In the ``unsupervised + supervised majority voting'' scheme, we obtained a score of 77.9\% and ranked 15th among 24 teams in total, including us. For this scheme, we found the optimal coefficients to be 0.6 and 0.4 for the supervised and unsupervised components, respectively, by leveraging the validation set. When we apply our other approaches and schemes, we rank worse overall. As in the case of Turkish, combining the unsupervised and supervised methods gives rise to the best success rates. When we compare the classical machine learning models and the neural models, we see that they have similar performances. We do not observe a large difference between the two paradigms as in Turkish, probably due to the simpler nature of the feature extraction process in this case (English).

\begin{table}[!h]
\centering
  \caption{Summary of performances (\%) of the unsupervised, semi-supervised and supervised methods, and their combinations for the three datasets in English.}
\bigskip
    \begin{tabular}{|>{\raggedright\arraybackslash}p{7.8cm} | c | c | c|}
    \hline
    \textbf{Approach}& \textbf{Movie} & \textbf{Twitter} & \textbf{SemEval}\\
   \hline
   
   Unsupervised&66.67&64.47&61.13\\
   Semi-supervised&68.18&64.53&63.45\\
   \hline
   SVM (tf-idf)&70.99&72.02&72.02\\
   SVM (3-feats)&73.97&74.72&73.48\\
   Supervised majority voting &73.99&74.02&74.42\\
   Unsupervised + supervised majority voting&\textbf{74.78}&\textbf{74.98}&\textbf{75.86}\\
\hline
   CNN (Non-static)&74.11&74.53&73.73\\
   LSTM (Non-static)&73.92&74.52&74.84\\
    \hline
    \end{tabular}
  \label{eng}
\end{table}

The most similar work to ours carried out for sentiment analysis in Turkish is the work proposed in \cite{tur:14}. They rely on unigram and bigram features with tf-idf weights, and achieve an accuracy of 89.5\% on the same movie dataset. We observed in our work that combining the supervised algorithms with unsupervised approaches leads to better overall performance. We used paired t-test and the approximate randomisation technique to measure the significance of the difference between the two works. The result obtained with the combined method (91.17\%) significantly outperforms the success rate obtained in \cite{tur:14} on the same dataset at $p=0.05$. In \shortcite{kay:12}, several supervised machine learning methods are also used and evaluated on the domain of political news. They analysed the effect of different types of features, such as unigrams and bigrams, adjectives, and a predefined list of polarity words, with term frequency and Boolean weights. They obtained an accuracy of 77\%. To the best of our knowledge, employing a semi-supervised domain-specific approach, combining unsupervised and supervised approaches, utilising the partial surface forms method, and making use of both static and non-static word embeddings in neural architectures for Turkish are the novel aspects in this proposed approach.

\subsection{Conclusion and Future Work}

In this approach, we have developed a framework for two-class sentiment classification problem in Turkish. The reviews and tweets were first subjected to comprehensive preprocessing operations tailored for the language and then features were extracted based on the normalised texts. We proposed methods that follow three main approaches. In the unsupervised and semi-supervised approaches, the polarity score of words were determined using a set of antonym seed words and then the sentiment of a document is obtained as the sum of the scores of the words it contains. In the supervised approach, we used several feature weighting metrics including novel metrics. The supervised methods were analysed in two groups, which are the traditional learning algorithms and the deep learning models. Lastly, the third approach formed ensembles of classifiers and also combined the unsupervised and supervised approaches in a novel way.

We have evaluated all these methods on two Turkish datasets with different characteristics. We conclude that combining unsupervised/semi-supervised and supervised methods yields the best results in both datasets. This indicates that incorporating knowledge obtained in an unsupervised manner into the classification process seems to be necessary to obtain more successful results. To the best of our knowledge, this is the first thesis work in Turkish that extracts sentiment scores of words using search engines, that employs domain-specific sentiment analysis, and also that combines unsupervised and supervised schemes. We have also evaluated our methods on three corpora in English and achieved the best results with this combination scheme as well. This proves the portability of this approach to other languages. The unsupervised and semi-supervised approaches can be adapted to other domains, and languages besides English and Turkish easily by choosing a relevant set of antonym pairs.

The experiments have shown that the feature engineering techniques used in determining the features and using suitable weighting schemes are the most important aspects in the supervised case. We observed that some preprocessing operations and specific feature selection methods may hamper the performance for some of the learning algorithms, such as negation handling and intensification in the CNN framework. Another finding is related to the agglutinative nature of the language in the sense that, in addition to the root forms of the words, making use of the morphemes within the surface forms gives rise to better results after performing normalisation. We have also shown that conventional machine learning algorithms using novel and effective feature engineering techniques can outperform the deep learning models, which are the hottest topic in the field of machine learning as of 2020.

We will extend our approach in the future by (1) applying it to other domains, such as hotel or restaurant reviews, (2) implementing different feature engineering techniques, and (3) performing error analysis for the ``unsupervised + supervised method'' to see the differences in the decisions of the classifiers. We also plan to (4) utilise sentiment lexicons in English, translate them into Turkish, and combine these polarity scores in lieu of unsupervised scores with supervised scores as discussed previously. A combination of supervised and unsupervised approaches can also be employed in neural network models and word embeddings in a similar manner.

\section{Major Contribution \#2: Morphological Analysis for Sentiment Analysis in Turkish}

In this approach, we examined the effect of a fine-grained morphological analysis for sentiment analysis in Turkish. Although the previous approach is applied on datasets in two languages, this approach is applicable only for Turkish and other agglutinative languages. In this case, partial surface forms as described beforehand are employed as features. After generating these feature sets, where some morphemes attached to their root forms are sentimentally more discriminative, we fed document vectors into classical machine learning methods to predict the polarities on a review-basis.

\subsection{Results}

\begin{figure}

\setlength{\abovedisplayskip}{14.5pt}
\includegraphics[width=\linewidth]{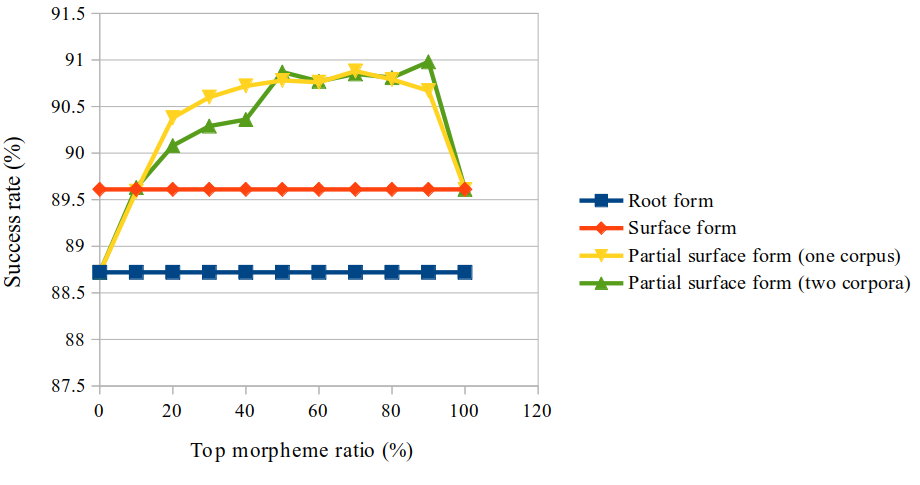} 
\caption{Effects of fine-grained morphological analysis with respect to different top (most discriminative) morpheme percentages on the Turkish movie corpus.}
 \label{morphoMovie}
\end{figure}

\begin{figure}[!htpb]

 \vspace{0.5cm} 
\includegraphics[width=\linewidth]{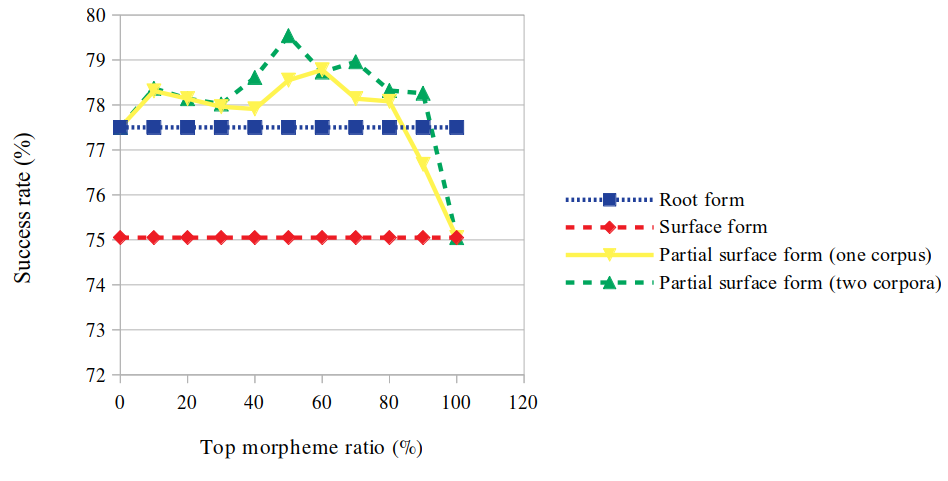} 
\caption{Effects of fine-grained morphological analysis with respect to different top (most discriminative) morpheme percentages on the Turkish movie corpus.}
 \label{morphoTwit}
\end{figure}

We performed an elaborative analysis about the partial surface forms technique we used for the Turkish texts. We examined the impacts of employing various ratios of suffixes and inducing morpheme polarity lexicons based on more than one corpus, and performed a comparative analysis with respect to surface and root forms of words. We empirically show the results in Figure~\ref{morphoMovie} (movie corpus) and Figure~\ref{morphoTwit} (the corpus of tweets), which are the same corpora discussed in the previous subsection. The performances have been achieved by employing the 3-feats technique and the SVM model. The horizontal axis in these figures represents the ratio of suffixes with the maximal confidence scores that are covered by the partial surface forms. The ratio 0\% corresponds to the root form, whereas 100\% corresponds to the surface form. The partial surface forms have been experimented in two different settings: We either employ one dataset (only the training set of the whole corpus) or process two datasets (again, only the training set of the target corpus and both the training and test sets of the other corpus). These figures show that when relying on only a subset of all the suffixes by taking account of their discriminative strength, our approach could perform better than the models using the root form and the surface form for almost all the top (morpheme) percentage values. We relate this good performance of partial surface forms to various factors. Some morphemes that have neutral-like polarities had better be filtered out, since these do not carry sentiment information. By processing only sentimentally the most discriminative morphemes by exploiting the polarity label information, we generate more robust sentiment representations for tokens occurring in an agglutinative language. When using several datasets at the same time in inducing morpheme polarity lexicons, we found out that this is especially useful for the Twitter dataset (Figure~\ref{morphoTwit}). We hypothesise that the reason is that the additional movie corpus used as a supplementary dataset is a less noisy and a bulkier corpus carrying more information.

Table~\ref{morp} summarises the impacts of differing morphological settings on the performance. In the partial surface form scenario, we include only the best success rates given in Figures \ref{morphoMovie} and \ref{morphoTwit}, which correspond to 90\% of the suffixes for the movie corpus and 50\% for the tweet corpus. The best performances are achieved by employing the partial surface forms metric. Thus, we have observed that making use of a set of discriminative morphemes is the most effective method for sentiment analysis in Turkish. Our review classification results are significant at $p=0.05$, when we compare the partial surface forms technique to root and surface forms.

\begin{table}[thbp]
\centering
  \caption{Summary of the impacts of fine-grained morphological analyses on the success rates (\%) for the two Turkish corpora with respect to the 3-feats scheme and the SVM method.}
\bigskip
    \begin{tabular}{|l|c|c|}
    \hline
   \textbf{Morphological setting} & \textbf{Movie} & \textbf{Twitter}\\
   \hline
   Root forms&88.72&77.50\\
    \hline
   Surface forms&89.61&75.05\\
    \hline
   Partial surface forms&\textbf{90.98}&\textbf{79.54}\\
    \hline
    \end{tabular}
  \label{morp}
\end{table}

\subsection{Conclusion}

Since Turkish is a morphologically-rich language, taking into account morphological features as well boosted the performance. Some morphemes are observed to be neutral-like, whereas some others are likely to express sentiments. That is, surface forms and root forms can both show weaknesses compared to the incorporation of the most sentimentally effective morphemes. The supervised algorithm we generated may be adopted for other morphologically-rich and agglutinative languages without producing any hand-crafted sentiment features specific to morphemes, and successful results could be arrived at as long as labelled data are available for one corpus (or more). In the case that no training data exist for a language, the previously mentioned unsupervised and semi-supervised methods can be used to generate polarity lexicons in the first place, and then the morpheme sentiment lexicon can be induced. However, as can be expected, we would not achieve as high accuracies for these cases.

As future work, we plan to (1) determine the sentiment scores of morphemes by implementing a neural network model which can more effectively capture this information and to (2) perform an error analysis.

\section{Major Contribution \#3: Combining Recurrent and Recursive Neural Networks for Aspect-Based Sentiment Analysis Using Inter-Aspect Relations}

In the literature, most studies conducted on sentiment analysis use either recursive or recurrent neural network frameworks alone. Recurrent networks can model the temporal effect and the sentiment information can accordingly be propagated throughout a text. By contrast, recursive neural networks can capture syntactic structures of the texts and the polarity information can be exploited in the training phase. Only several related works merge both these models in an ensemble neural network for polarity detection. In this approach, an original neural network model is proposed such that both these neural models are combined for ABSA. Relying on constituency and dependency parsers, we initially break down each review into sub-reviews (subclauses) that bear the sentiment information related to only the corresponding aspect terms. After we generate and train the recursive neural models based on the parse trees of the sub-reviews, we use their output in the recurrent model. This ensemble approach is evaluated on two corpora of different categories in English. Results we arrived at are state-of-the-art and we outperform a study (an RNN-only framework) which we chose as a baseline by a significant margin for the two corpora.

A single review in general bears a sentiment about an entity, such as a service, product, or political act. However, reviewers can also comment on different aspects of the same entity in a single review. For example, in \textit{``I found the ambiance of the restaurant great overall; however, the main dish was served a bit cold and lately.''}, opinions expressed towards the aspects \textit{``ambiance''} and \textit{``dish''} are positive and negative, respectively. Therefore, we need to perform a fine-grained analysis rather assign a single overall polarity score to a review.

\enlargethispage{-\baselineskip}

If multiple aspects are commented about in a single review, the polarity of these each is likely to affect the following or preceding aspects as well. For instance, in \textit{``I liked the taste of pizza more than that of the chips.''}, it is observed that the negative sentiment of the aspect \textit{``chip''} is expressed in an indirect way by the first aspect \textit{``pizza''}. That is, the polarities of aspects are likely to affect one another in the same review. Additionally, several conjunctions (e.g. \textit{``and''}, \textit{``also''}, \textit{``however''}, and \textit{``but''}) lead to aspects sharing their polarities with other aspects or influencing the polarities of other aspects in the same review. In the sentence \textit{``The quality of the display of this laptop is so sensational, so is the price thereof.''}, a correlation exists between the polarities of the aspects because of the conjunction \textit{``so''}. A work \cite{maj:18} develops a recurrent neural network approach by exploiting such inter-aspect relations, when modelling such scenarios.

Such RNN models utilise the sequence information in a series of objects. The effect of the sentiments is thereby propagated across the same text, be it in a forward or reverse setup. Nonetheless, the grammatical information is not covered by such models. For example, after a sentence is broken down and parsed into its constituent phrases, the words in the same subtrees/subgroups are more likely to be semantically, syntactically, and sentimentally more akin to each other than those words in other subtrees. Hence, relying on recursive neural networks is useful in that the words in the same subtrees can be of the same or similar sentiment scores. If we merge this structural, sentimental information into other neural network models, such as RNNs exploiting the temporal information, we can arrive at more comprehensive sentiment analysis frameworks. As such, sub-models in this ensemble models can each compensate for what the other lacks.

In this approach, a novel framework is proposed for ABSA \cite{pon:14}. To exploit the sentiment information of aspects, we merge recursive and recurrent neural network models. In the recurrent sub-model, we rely on an off-the-shelf framework \cite{maj:18} which employs gated recurrent units. In the recursive sub-model, we describe original approaches to extracting sub-reviews, each corresponding to aspect term groups, from whole reviews. All sub-reviews are captured such that every one of them is modified by one polarity at most. Each sub-review is considered a separate review and these are trained by recursive neural networks. We extract the root vectors from these sub-reviews in a distant-supervised manner, whereby these root embeddings represent the aspects therein.  We then feed these vectors as input into the recurrent neural networks and the effect of sentiments is propagated along with other information.

We test and evaluate the original approach using two corpora, which are restaurant and laptop datasets provided in the Task 4 of SemEval-2014. Our approach performs better than the baseline study \cite{maj:18} significantly. Employing the recurrent network only cannot capture polarity information as effectively. The recursive model helps create the \textit{``optimal''} sentiment root embeddings of the sub-reviews, which are later merged with the related aspect component vectors in the recurrent model. When incorporating a recursive model trained with a distant-supervised approach into a recurrent model and, as such, forming a novel ensemble framework, we boosted the performance, where an increase of 1.6\% on average is observed for the two datasets.

Our research objective in this approach is to obtain better performances for ABSA. To achieve this goal, as mentioned in a preceding chapter, we address the following research questions. Could we enhance recurrent neural networks by incorporating distant sentiment information? Can we merge recursive and recurrent neural models in an ensemble form? If so, what is the effect? Why do such ensemble approaches perform better than those modelling only one of its sub-models? Can we break down reviews into sub-reviews/subclauses using syntactic parsers such that each sub-review bears only one related sentiment expressed towards the aspect or aspects therein? Is employing dependency parsers a better choice over the use of constituency parsers in ABSA? If so, what is the reason?

\subsection{Methodology}

We conduct ternary aspect-based sentiment classification in this approach, where the aspects are either negative, positive, or neutral. First, we perform basic tokenisation and other preprocessing operations, such as lowercasing, on texts employing the spaCy library \cite{hon:15}. Then, we generate sub-reviews from whole reviews corresponding to each aspect by using a constituency and a dependency parser. Each of these sub-reviews may contain one aspect or more. In the end, we merge recurrent and recursive neural networks into an ensemble form as mentioned. In the remainder of this section, we first give a brief overview on the baseline recurrent model. Then the proposed recursive neural network model and the ensemble model based on these two sub-models are described.

\subsubsection{Submodel 1: Recurrent Model}

\label{sec:recurrent}

In this sub-model, the framework \cite{maj:git-iarm} developed by the study \cite{maj:18} which we regard as the baseline method is used and enhanced by us. This model employs inter-aspect relations in such a way that aspects can have an impact on the polarities of the previous or succeeding aspects. We keep the hyper-parameters of this model the same to perform a consistent, comparative analysis. We arrive at better results for two corpora of different genres when integrating this recurrent network with a recursive network in an original way.

\begin{figure}[h] 
\begin{center}
\includegraphics[width=\textwidth]{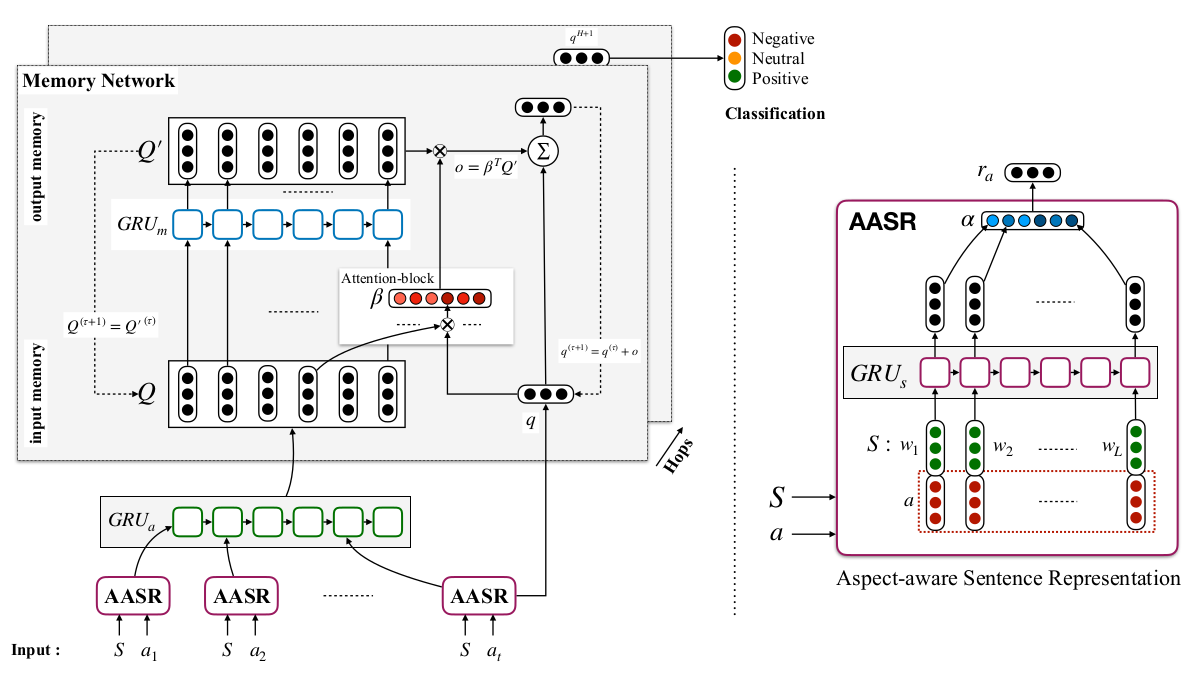} 
\caption{The architecture of the baseline study \cite{maj:18}. AASR denotes \textit{Aspect-Aware Sentence Representation.}}
 \label{baseline}
\end{center}
\end{figure}

We explain the baseline study briefly as follows. The GloVe vectors are fed as input into the system. In aspect-aware sentiment representation (AASR) modules, the contextual information is propagated throughout the word sequence. Additionally, an attentive mechanism is relied on to identify and boost the effects of the polarities that are expressed towards the modified aspects. This is repeated for each aspect term group so that its polarity has an impact on those of the others. A softmax classifier is used to identify the sentiments of the given aspects over the top layer. A mechanism named ``multiple hops'' is also used in their study. This concept is defined as follows: The hidden outputs of the text are repeatedly fed as input into the system several times. Hereby, a fine-grained representation of aspect terms is modelled. This approach is visually summarised in Figure~\ref{baseline}.

\subsubsection{Submodel 2: Recursive Model}

Recursive neural network architectures are made use of to train models by modelling the grammatical structures of the sentences. For example, in the review \textit{``I adored this flick!''}, the verb \textit{``adore''} bears a positive polarity orientation and this modifies an object noun, which is \textit{``flick''}. The sentimental and semantic information are thereby captured from the text more saliently and successfully as compared to other models (feed-forward neural networks, \textit{n}-grams, bag-of-words, etc.). In the remainder of this section, the constituency and dependency parser sub-models we developed in this sub-approach are described. We combine these models with the recurrent network model as will be explained and discussed in Section~\ref{sec:ensemble}.

\textit{\underline{Constituency Parser Sub-model}}

Constituency parsers generate parse trees by decomposing sentences into chunks. For the sentiment analysis problem, breaking down the texts into such phrases can help organise polarity information in a well-defined, structured manner. For instance, if a negator (e.g. \textit{``not''}) explicitly appears in a chunk, this would lead to a sentimental change in an opposite direction. A similar mechanism holds true if a phrase is followed by a contrastive conjunction, such as \textit{``however''}. Modelling such structures along with the polarity information can help us achieve a better performance \cite{soc:13}.

\begin{figure}[h] 
\begin{center}
\includegraphics[width=\textwidth]{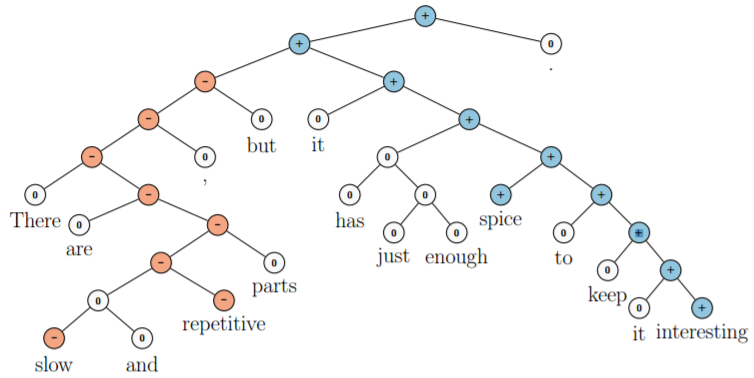} 
\caption{A sample review parsed into its chunks/phrases along with their polarity categories, which range from very negative to very positive (-{}-, -, 0, +, ++).}
 \label{recNeuTenNet}
\end{center}
\end{figure}

According to this sub-model, we rely on the Stanford CoreNLP framework \cite{man:14} to parse the sentences into constituency phrases/chunks along with polarity labels. We give an example that visualises the recursive structure of a review in Figure~\ref{recNeuTenNet}. In this example, every node of the parse tree is modified by a polarity category, which may be very negative (-{}-), negative (-), neutral (0), positive (+), or very positive (++). These polarities are identified by the above-mentioned tool in a distant-supervised way. That is, this approach does not exploit the label information present in the training dataset. As said earlier, this sub-model can effectively capture the negation and scope relations in such parse trees.

Before using the parse trees in our recursive model, we first identify aspect groups/chunks. In other words, per aspect term, we create a subtree covering the aspect term. We then train these sub-reviews/subtrees in the recursive model separately. The reason behind this approach is that the same review may have various aspects with opposite sentiments. Therefore, when we break down reviews into sub-reviews with respect to aspect terms, this helps us gather the relevant sentiments of these aspects. The fine-grained analysis can thereby boost the performance for ABSA.

Our original approach to creating a subtree per aspect term is described as follows. We first start scanning a tree from the leaves (terminals) that represent aspect term words which are labelled in the training data and then go in upward direction in the tree. If a node whose polarity is non-neutral is encountered, we hypothesise that this sentiment word modifies the corresponding leaf aspect term and we cut the parse tree at that point. As such, we take this node and all of its child nodes as the aspect subtree for the corresponding aspect term. Accordingly, this subtree represents the aspect term that is covered in it. It is therefore probable that the same sub-reviews/subtrees can encompass different aspect terms, as in \textit{``I loved the ambiance and service overall.''} Additionally, several rules are defined for negators. For instance, when a negating word appears at a higher level than the node where we cut the tree, we keep expanding the subtree in an upward direction until we cover the negator node and all of its children into the same subtree. Then, we cut the parse tree at that point. This is usually the case that the negating words lie at higher node levels compared to the words modified by them. Lastly, after generating these sub-reviews, we train them disjointly as if they are different reviews by the conventional recursive neural model \cite{soc:13}.

We arrived at better results as compared to the baseline when taking sub-reviews into consideration and training them separately for our ensemble framework. Nevertheless, this sub-model has a drawback. As mentioned, three gold polarity labels are defined for aspects provided in the SemEval-2014 datasets, which are used in the evaluation stage. However, if a node of positive or negative polarity is encountered, we do not expand the tree per aspect term in the bottom-up manner any more. In other words, we do not take account of the case whereby an aspect term can be neutral. Yet another deficiency about the constituency sub-model is that we can fail in generating robust or meaningful subtrees from time to time. We hypothesise that the reason may be that constituency parsers have an incompetence: They cannot function as effectively as dependency parsers for the opinion mining task. Dependency parsers define relationships (e.g. modification) much more clearly in contrast to constituency parsers. In dependency parser outputs, sub-reviews are in general linked to each other with predefined relationships. Thus, we are more capable to accurately extract these relevant subclauses from the whole sentences. Such a capability is not provided by constituency parsers.

After having generated the sub-reviews using the constituency parser tool, we rely on an open source recursive neural network framework \cite{che2:18} when training these trees beforehand. To handle the overfitting problem, we make use of a development dataset. The maximum epoch number is chosen as 30, since when exceeding this value, in general overfitting occurs. We treat the vector size as a hyper-parameter and define the value set as \{30, 50, 100\}. The embeddings at the nodes encode the sentimental information in the parse tree and are updated in the training phase. After the training epochs are completed, we employ the root vector of each sub-review per aspect in the ensemble model.

\textit{\underline{Dependency Parser Sub-model}}

\label{sec:dep}

As said, when creating sensible aspect subtrees to be used in the ensemble framework, constituency parser can be unsuccessful. In the literature, it is also reported that these parsers cannot usually perform as well as dependency parsers in the sentiment analysis domain \cite{don:15}. Therefore, dependency parsers are also made use of in this approach to generating sentiment embeddings for aspect terms, which are later fed into our ensemble approach.

\begin{figure}[h] 
\begin{center}
\includegraphics[width=\textwidth]{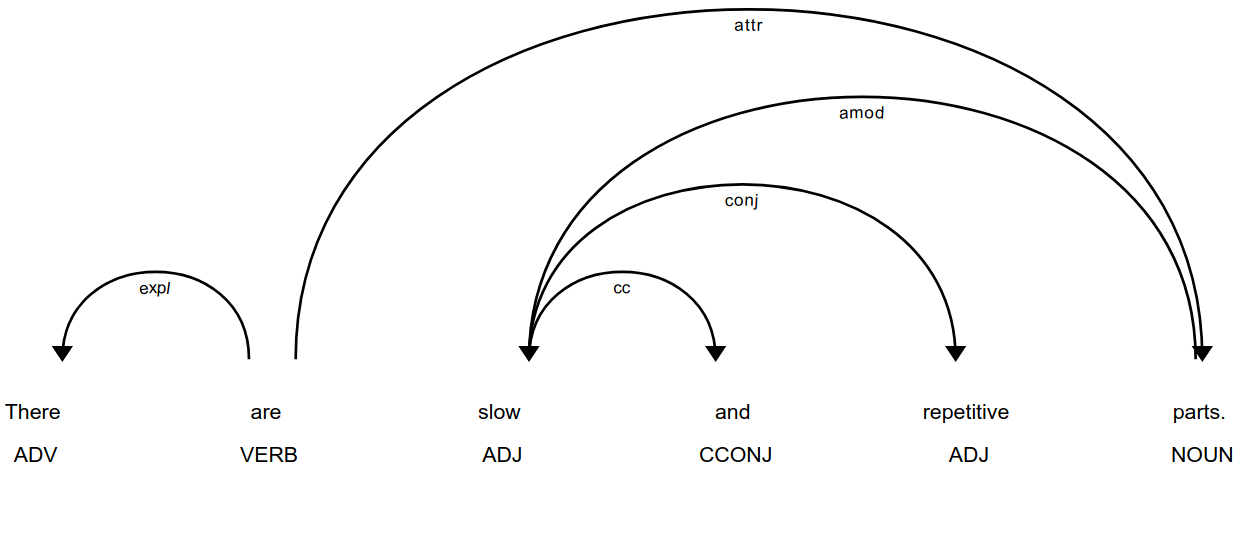} 
\caption{Example of a sentence decomposed into its relationships.}
 \label{depenEx}
\end{center}
\end{figure}

Words in a text are connected to each other based on the binary relationships between them by dependency parsers. Vertices in the tree are tokens in the sentence(s) and edges are labelled by relationships. Parents, which are the source of an edge, modify their child nodes. For instance, in \textit{``There are slow and repetitive parts.''}, the words \textit{``slow''} and \textit{``repetitive''} modify the word \textit{``parts''}. This example is visualised in Figure~\ref{depenEx}. In this example, the relationship \textit{``amod''} stands for adjectival modifier. In dependency parse trees, for a word can be linked to another word with a modifier relationship, these parsers can help capture the sentiment information as well more effectively and accurately than constituency parsers do. The spaCy library is employed to generate dependency relationships from a text.

As is the case in the constituency parser model, first, we extract sub-reviews per aspect term from whole reviews relying on the dependency parser. To achieve this, a few rules are defined to decompose a whole dependency tree generated for a review into sub-review trees. If the word's  part-of-speech tag is a verb (be it in gerund, infinitive, or any other form) and this is linked to conjunction or clausal component relationship, this marks the existence of a sub-review. In other words, all the child nodes of a verb that are linked through relationships other than those two are incorporated into the corresponding sub-review in a recursive manner. For example, in \textit{``There are slow and repetitive parts, but it has just enough spice to keep it interesting.''}, the word \textit{``are''} is the main verb of this sentence. The conjunction relationship links it to the secondary verb \textit{``has''}. The dependency tree is therefore cut at this point. We cover all the children of the verb \textit{``are''} except \textit{``has''} (and its respective children) in a recursive manner and build a sub-review. We perform the same process for the word \textit{``has''} with regard to its children. Hence, the resulting sub-reviews are \textit{``There are slow and repetitive parts.''} and \textit{``But it has just enough spice to keep it interesting.''}

After generating sub-reviews, we remove subtrees (subclauses) which have no aspects and are therefore considered unnecessary. We then filter out the preceding and following redundant conjunctions, such as \textit{``and''}, if any. We also discard the punctuation marks (e.g. \textit{``!''}) at the beginning and end of the sub-reviews in the first place. Then, we append the punctuation mark appearing at the end of the whole review (e.g. \textit{``!''}) to the end of each sub-review instead to arrive at consistency. In case there are not any clausal components or conjunctions in a review, we hypothesise that there are not any possible subclauses/sub-reviews and the review has only one single sentiment. However, a single sub-review (subtree) can consist of one aspect or more than one aspect. We accordingly obtain more robust sub-reviews conforming to these rules compared to the outputs of the constituency parser.  

\begin{figure}[h] 
\begin{center}
\vspace{0.10cm}
\includegraphics[width=\textwidth]{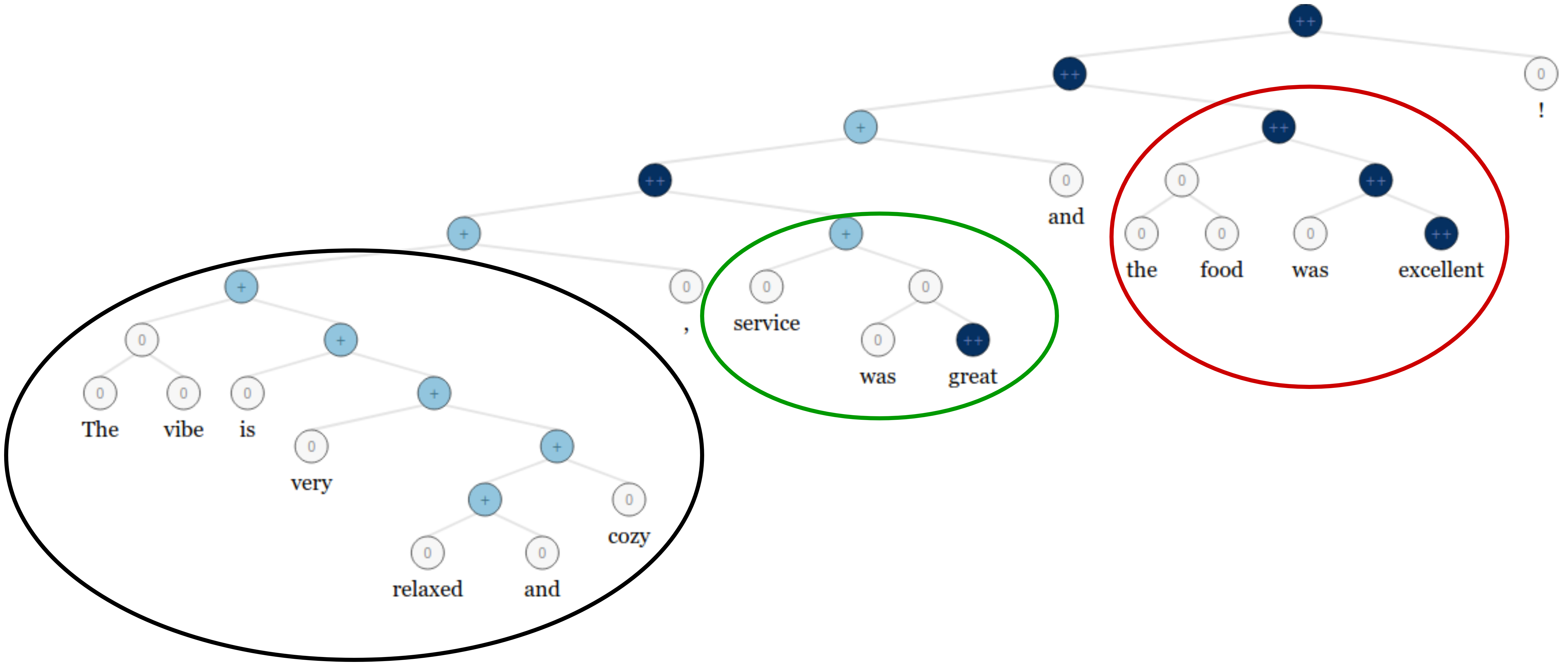} 
\vspace{0.10cm}
\caption{The text \textit{``The vibe is very relaxed and cozy, service was great and the food was excellent!''} decomposed into its sub-reviews. Each encircled part in the figure is a sub-review.}
 \label{subrevs}
\vspace{0.51cm}
\end{center}
\end{figure}

We give a sample text that visualises sub-reviews created by the constituency and dependency parser models in Table~\ref{sub-reviews}. The reader can see that all the sub-reviews built are relevant only to the aspects therein for the dependency parser model. Additionally, the subtree covers aspect-specific sentiments expressed only towards their aspects. Nevertheless, in the constituency parser model, when decomposing a review into sub-reviews, outputs shown in the table are not as consistent. For instance, the model predicts the words \textit{``French''} and \textit{``gourmet''} as positive by a polarity score of 3 according to the Stanford NLP library. The bottom-up approach defined for the constituency parser therefore stops scanning the tree, when a word of a positive or negative polarity is come across by the algorithm. That is, we cut the parse tree at that point. We visually show the last sentence (review) broken down into its sub-sentences (sub-reviews) in Figure~\ref{subrevs}. Each of the three encircled parts in the figure~is a sub-review.

\begin{table}[!h]

\centering
\caption{Example sentences and the sub-sentences for each aspect term built employing the constituency and dependency parsers. The gold aspect terms are underlined.}
\bigskip
\begin{tabular}{|>{\raggedright\arraybackslash}p{4.3cm}|>{\raggedright\arraybackslash}p{3.5cm}|>{\raggedright\arraybackslash}p{3.5cm}|}
\hline
\multirow{2}{=}{\textbf{Review}} & \multicolumn{2}{c|}{\textbf{Sub-reviews generated}}\\\cline{2-3}
&\multicolumn{1}{>{\centering\arraybackslash}m{3.5cm}|}{\textbf{Dependency}}&\multicolumn{1}{>{\centering\arraybackslash}m{3.5cm}|}{\textbf{Constituency (Bottom-up)}}\\
\hline
\multirow{3}{=}{It may be a bit \underline{packed} on weekends, but the \underline{vibe} is good and it is the best \underline{French food} you will find in the area.}


& \textbf{1.} It may be a bit \underline{packed} on weekends.&\textbf{1.} \underline{Packed} on weekends.\\
&\textbf{2.} The \underline{vibe} is good.&\textbf{2.} The \underline{vibe} is good.\\
&\textbf{3.} It is the best \underline{French food} you will find in the area.&\textbf{3.} \underline{French food}.\\
\hline
\multirow{3}{=}{ I love and I know \underline{gourmet food} by excellence!} & \textbf{1.} I love and I know \underline{gourmet food} by excellence!&\textbf{1.} \underline{Gourmet food}.\\
\hline
\multirow{3}{=}{The \underline{vibe} is very relaxed and cozy, \underline{service} was great and the \underline{food} was excellent!} &\textbf{1.} The \underline{vibe} is very relaxed and cozy!&\textbf{1.} The \underline{vibe} is very relaxed and cozy!\\
&\textbf{2.} \underline{Service} was great!&\textbf{2.} \underline{Service} was great!\\
&\textbf{3.} The \underline{food} was excellent!&\textbf{3.} The \underline{food} was excellent!\\
\hline
\end{tabular}
\label{sub-reviews}
\end{table}

\enlargethispage{-\baselineskip}

After we build sub-reviews from reviews making use of the dependency parser, we train the recursive sentiment model. Accordingly, each parse tree representing a sub-review is trained in a recursive way, independently of one another. These are then fed as input into the ensemble framework. We employ a publicly available code repository for this sub-model \cite{kor:17}. We slightly changed the parameters of this framework to adapt it to our corpora. In this sub-model, as in the constituency parser model, each node is modified by a polarity label. A sentiment lexicon \cite{ham:16} in which the words are labelled as -1 (negative), 0 (neutral), or 1 (positive) is leveraged to identify the polarities of words. If this lexicon does not include a token, we assign it the score of 0. This polarity lexicon is used in identifying the sentiment scores of each node in the dependency parse tree. Despite the fact that the labels can belong to one of the five sentiments for the constituency parser model, conducting a coarser analysis in the dependency parser model helped us achieve higher success rates in the final ensemble framework. Exploiting these sentiment labels, we recursively train these vectors at the nodes of the parse tree.

\enlargethispage{-\baselineskip}

Our models have been evaluated for various vector lengths of 30, 50, and 100. After the training phase for the recursive neural network is completed, the root vectors for each aspect group are fed as input into the ensemble framework. As we empirically explain and show in Section~\ref{sec:results}, we have separately assessed the performances of vectors of different lengths. As will be shown Section~\ref{sec:results}, the dependency parser model performs significantly better compared to both the baseline model alone and the ensemble approach with the constituency parser module.

Apart from this basic approach, two additional, different settings are also tested such that gold polarity labels of aspects corresponding to the nodes in the parse tree are taken into consideration. For the first variant, the leaves in the tree representing aspect terms are labelled as the same gold polarity values of aspects that are present in the training data instead of using the information in the sentiment lexicon. These sentiment scores can be either -1, 0, or 1. These are conducted for the training stage. On the other hand, in the testing phase, we perform majority voting scheme to identify the polarity of an aspect term in the test data by looking up the polarity values in the training dataset of this aspect. In the second scenario, the polarity of the root node in the tree is assigned the gold label of the aspect it covers. The reason behind that is that the root node is the most effective component when propagating sentiment through the tree. The above-mentioned majority voting process is also performed for this variant as well. Nonetheless, although we used the gold sentiment labels present in the training data, we observed a decrease in the accuracies. We think that it stems from the fact that combining sentiment information coming from different sources disarranges the consistent structure of the tree. For instance, a word's sentiment score can be -1 according to the polarity lexicon. On the other hand, this score could be defined as 0 in the training data as a gold label. Thus, this would lead to an ambiguity. Therefore, when we used the sentiment scores relying on only one source (i.e. sentiment lexicon), we arrived at better results. That is, this approach produces a more consistent sentiment structure for the text and the model would remain more robust. We will discuss the related results in Section~\ref{sec:results}.

\subsubsection{Ensemble Framework Combining Recursive and Recurrent Models}

\label{sec:ensemble}

\begin{figure}[h!] 
\begin{center}
\includegraphics[width=\textwidth]{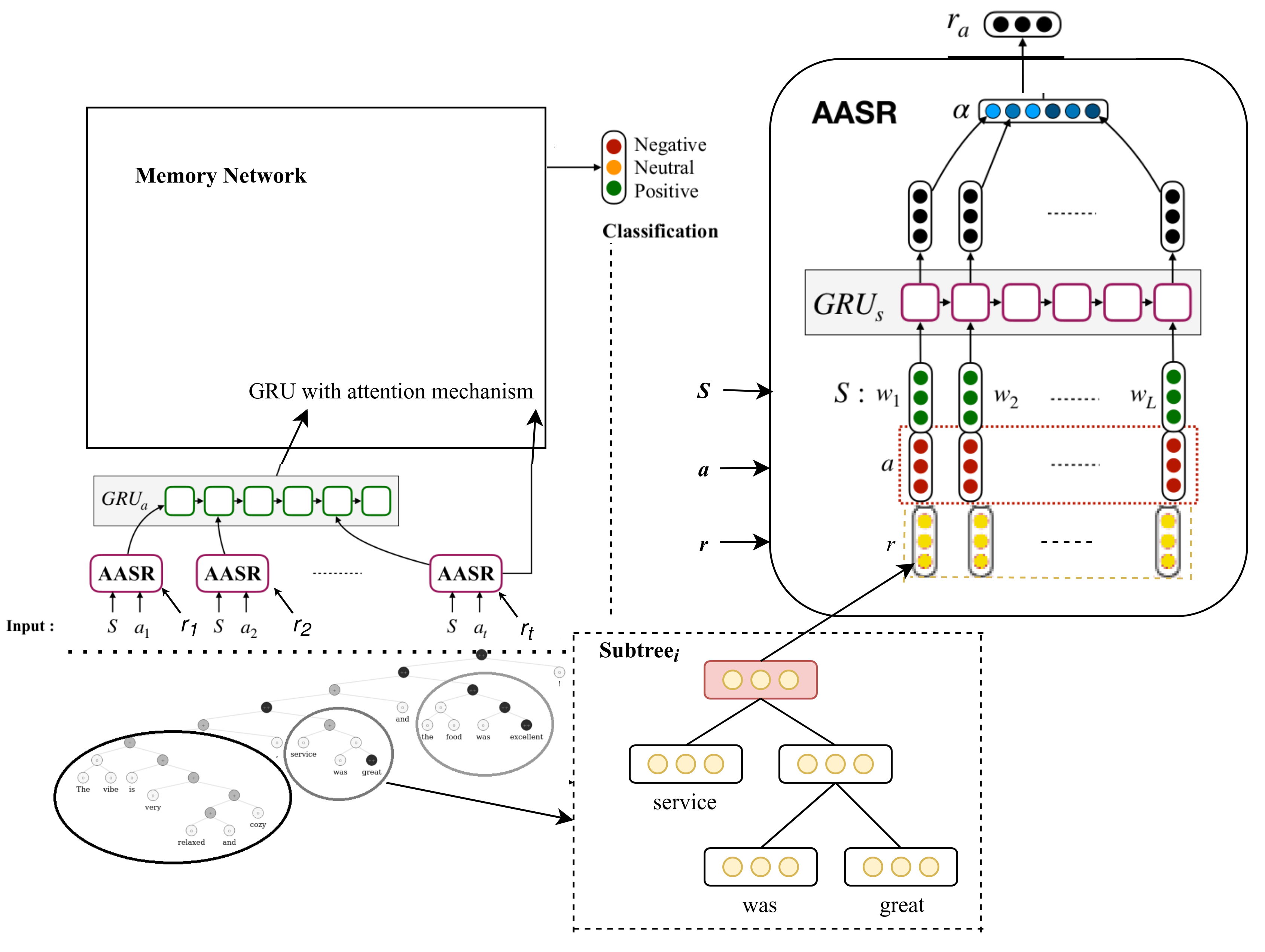} 
\caption{Visual summary of our proposed ensemble framework.}
 \label{ensemble}
\end{center}
\end{figure}

The above-described recursive and recurrent models are integrated into an ensemble form in a hierarchical manner. In this scenario, the root embeddings of the parse trees trained in the recursive network are employed as input in the recurrent network. The baseline study \cite{maj:18} for this approach only makes use of GRUs, a recurrent model, with an attention mechanism, as explained in Section~\ref{sec:recurrent}. Our recurrent sub-model is the same as this. The second sub-model we relied on is a recursive neural network using a constituency or dependency parser. We also extract sub-sentences from whole sentences by leveraging these tree parsers in an original manner. As mentioned, the recurrent sub-model captures the temporal information being propagated throughout the text, whereas the recursive sub-model extracts the grammatical and sentiment information from the reviews in a fine-grained scope. That is, those two sub-models can be considered to complete each other in areas where they cannot perform as robustly and we obtain more powerful representations when combining these two. We show the ensemble framework in Figure~\ref{ensemble}. 

As mentioned, in this approach, in addition to the recurrent baseline model, we have also employed the corresponding root embeddings of the parse tree that covers the affected aspect and incorporated it into the ensemble model. The module named memory network is the same as that in the baseline study, which is only a recurrent model. The sample given on the bottom left demonstrating the sub-review generation process is the same as in Figure~\ref{subrevs}. $Subtree_i$ corresponds to the second sub-review. In this case, since three sub-sentences are generated from a single sentence, three AASR modules are produced. In this approach, we first break down reviews into sub-reviews, each of which holds the relevant sentiment about the affected aspect. Then we train those sub-reviews separately in parse trees. After that, we feed the root embeddings of these trees into the recurrent model. In AASR modules, aspect term embeddings and root embeddings of the related aspect subtrees are concatenated along with all sentence vectors. The GloVe embeddings of sentence words are utilised. If an aspect consists of more than one word, we take their average to represent the aspect group as a whole.

The main intuition behind our using root vectors is that these embeddings capture the overall sentiment, semantic, and syntactic information of the sub-review tree more comprehensively compared to the nodes below its level. The propagation process of sentiments throughout the parse trees is mostly affected by this root node. When we have also taken these sub-review parse trees and root embeddings into consideration, we observed that the recurrent model is enriched by incorporating additional grammatical and sentiment information in a novel way as well.

\subsection{Experimental Evaluation}

In the remainder of this section, first, we describe the corpora we utilised for ABSA. We then explain the hyper-parameters we have used. Lastly, we give a detailed explanation on and discuss the experimental settings and the results we obtained.

\subsubsection{Datasets}

We evaluated our methods on two datasets which are provided in the Task 4 of the SemEval-14 competition. These two corpora are restaurant and laptop reviews. In both these datasets, aspects are given in advance. That is, the competitors only try to find out what the polarities of these aspect terms are. Therefore, we have not developed a specific approach to extracting aspect terms from the datasets for this task. Each aspect is assigned a sentiment gold label, which can be negative, neutral, or positive. These are visualised in Table~\ref{datasets}. The statistics state the distribution of sentiments across the datasets.

\begin{table}[!h]
\centering
\caption{Details of the datasets we used in this approach.}
\bigskip
\begin{tabular}{|l|l|c|c|c|}
\hline
\multicolumn{2}{|l|}{\textbf{Dataset}} & \textbf{Positive} & \textbf{Neutral} & \textbf{Negative}\\
\hline 
\multirow{2}{*}{{\relax\ifvmode\fi Restaurant corpus}} & 
Training & 2,159 & 632 & 800\\
\cline{2-5}
& Test & 730 & 196 & 195\\
\hline
\multirow{2}{*}{{\relax\ifvmode\fi Laptop corpus}} & 
Training & 980 & 454 & 858\\
\cline{2-5}
& Test & 340 & 171 & 128\\
\hline
\end{tabular}
\vspace{4mm}
\label{datasets}
\end{table}

We have not performed cross-validation, since the datasets are split as the training and test sets in advance. We split 10\% of the training set as the development set to handle the overfitting issue in our neural network models and to detect the optimal values among hyper-parameters.

\subsubsection{Hyper-parameters}

Three publicly available code repository are utilised for this approach. We show the hyper-parameter sets in Table~\ref{recursiveHyperparams}. Two of these are defined for the recursive neural network models, in which dependency and constituency parse trees are trained disjointly and independently of each other. Hidden states at each node are trained in these parse trees. The other setting is the same recurrent model as in the baseline. Hidden vector sizes are set at 30, 50, and 100, and we tried to figure out which one of them produces the optimal results. We used a validation set to find these values maximising the performance. When we choose the size of hidden vectors as 100, we observe the overfitting to occur for all cases. We think the reason is that that makes the model too complex. At the bottom-most layer, we feed the GloVe vectors of the size 300 corresponding to tokens in the text. On the other hand, the vectors at intermediate nodes can be of various lengths. For the constituency parser, we learn the leaf embeddings from scratch, because the code repository we relied on does not allow us to use pre-trained vectors. The hyper-parameters utilised in the baseline approach are listed in Table~\ref{recurrentHyperparams}.

\begin{table}[!h]
\centering
\caption{Hyper-parameters made use of in the recursive and recurrent neural networks.
 \textit{``Optimal size''} corresponds to the optimal size of embeddings at intermediary nodes in trees.}
\bigskip
\setlength{\tabcolsep}{3pt}
\begin{tabular}
{|>{\raggedright\arraybackslash}p{79pt}%
|>{\raggedright\arraybackslash}p{60pt}%
|>{\centering\arraybackslash}p{87pt}%
|>{\centering\arraybackslash}p{87pt}%
|>{\centering\arraybackslash}p{74pt}%
|}

\hline
\multicolumn{2}{|l|}{\textbf{Hyper-parameter}} & \textbf{Constituency parser model}&\textbf{Dependency parser model}&\textbf{Recurrent model}\\
\hline
\multicolumn{2}{|l|}{Learning rate}&0.001&0.05&0.001\\
\hline
\multicolumn{2}{|l|}{\pbox{52pt}{Maximum number of epochs}}&30&30&30\\
\hline 
\multicolumn{2}{|l|}{Batch size}&20&25&30\\
\hline
\multirow{2}{*}{\pbox{2.0cm}{\relax\ifvmode\raggedright\fi Optimal size}}&Restaurant& 50&50&-\\
\cline{2-5}
&Laptop&30&30&-\\
\hline

\end{tabular}
\label{recursiveHyperparams}
\vspace{0.68cm}
\end{table}

\begin{table}[!h]
\centering
\caption{Hyper-parameters made use of in the baseline model.}
\bigskip
\begin{tabular}{|p{120pt}|>{\centering\arraybackslash}p{70pt}|>{\centering\arraybackslash}p{70pt}|}
\hline
\textbf{Hyper-parameter} &\textbf{Laptop}& \textbf{Restaurant}\\
\hline
Hop Count&10&3\\
\hline
GRU Hidden State Size&400&300\\
\hline
GRU Output Size&400&350\\
\hline

\end{tabular}
\label{recurrentHyperparams}
\vspace{0.68cm}
\end{table}

\subsubsection{Results}
\label{sec:results}

We list the accuracies obtained for the two corpora in Table~\ref{ens-results}. In the first row, the results for the baseline approach are given. The two rows below it are the accuracies produced for different recursive neural network model settings, which are later integrated to the ensemble framework. First, the results for the constituency parser are given. Then, those for the dependency parser are listed. The three approaches described in Section~\ref{sec:dep} are labelled  \textit{``root”}, \textit{``leaves”}, and \textit{``gold aspect”}. In the \textit{``root”} metric, the root vectors of the trained parse trees are fed into the GRUs network. For the \textit{``leaves”} feature, instead of root vectors, leaf aspect embeddings learnt are employed. On the other hand, \textit{``gold aspect''} models the scenario in which the sentiment labels present in the training dataset are exploited as explained in the previous subsections. That is, the label of the root node is assigned the same polarity of the affected aspect in the corresponding sub-review. We also utilised gold labels only for the leaf nodes in the parse trees, that is, not for the root node. As mentioned, for identifying the polarities of the nodes in the test parse trees in advance, we performed a majority voting scheme taking account of the frequency statistics in the training data. However, if an aspect word in the test data does not occur in the training data, we exploit the label information in the sentiment lexicon. However, we do not list the results of this last scheme in the table, since it does not produce state-of-the-art and significant results. If the reader has a hard time comprehending these concepts, he or she can refer to Section~\ref{sec:dep}. The embedding size corresponds to the lengths of vectors at the nodes in the parse trees. Here, we test and assess the impact of embedding sizes only for the recursive neural sub-model. The reason is that we wanted to make a consistent comparison between our model and the baseline approach utilising the same hyper-parameter settings.

\begin{table}[t]
\centering
\caption{Accuracies (\%) obtained for the baseline method and our ensemble approach which integrates the recursive neural network model into the baseline recurrent network. Results for different embedding size and other settings are listed.}
\bigskip
\begin{tabular}{|p{4.5cm}|p{1.5cm}|p{1.9cm}|c|c|}
\hline
&&&\multicolumn{2}{c|}{\textbf{Accuracy (\%)}}\\
\cline{4-5}
\textbf{Model}&\textbf{Feature}&\textbf{Recursive NN vector size}& \textbf{Restaurant}& \textbf{Laptop}\\
\hline
\multicolumn{3}{|l|}{Baseline (IARM) \cite{maj:18}}&80.00&73.80\\
\hline
\multirow{6}{*}{\pbox{4.5cm}{\relax\ifvmode\raggedright\fi Constituency Parser + Recurrent NN}} &\multirow{3}{*}{\pbox{1.5cm}{\relax\ifvmode\raggedright\fi Leaves}}&30&77.98&72.23\\
&&50&79.20&72.10\\
&&100&78.10&72.00\\
\cline{2-5}
&\multirow{3}{*}{\pbox{4.5cm}{\relax\ifvmode\raggedright\fi Root}}&30&79.48&\textbf{74.20}\\
&&50&\textbf{80.27}&73.73\\
&&100&78.10&72.97\\
\hline
\multirow{6}{*}{\pbox{4.5cm}{\relax\ifvmode\raggedright\fi Dependency Parser + Recurrent NN}} &\multirow{3}{*}{\pbox{1.5cm}{\relax\ifvmode\raggedright\fi Gold aspect label}}&30&78.30&72.90\\
&&50&78.57&72.85\\
&&100&77.90&72.48\\
\cline{2-5}
&\multirow{3}{*}{\pbox{4.5cm}{\relax\ifvmode\raggedright\fi Root}}&30&79.90&\textbf{76.15}\\
&&50&\textbf{80.90}&75.75\\
&&100&79.82&73.70\\
\hline
\end{tabular}
\label{ens-results}
\end{table}

We empirically show that we outperform the baseline approach (IARM) for both corpora when the root scheme is utilised. This holds true for both the dependency and constituency parser settings. According to these results, for the laptop dataset, we outperformed all the 26 teams including us that participated in the Task 4, aspect-polarity detection in the SemEval-2014 competition. When it comes to the restaurant reviews, we rank second among all the teams. We attribute our success to the following reason. The top-performing teams rely on the SVM classifiers, into which they feed hand-crafted features among others. However, in our approach, a robust ensemble framework is built combining two different deep neural network models. It is well-stated in the literature that in general conventional machine learning methods are not as effective as deep neural networks and can hardly compete with them \cite{gol:17}. Deep learning networks generate features on their own and rely on more complex and more robust, representative models.

When performing a fine-grained analysis, we observe that dependency parsers produce more successful results as compared to constituency parsers. The main intuition behind it is that the modifier relationship can be directly captured in dependency parser module unlike in constituency parser module. These modifier relations are useful in both generating sub-reviews and identifying sentiment expressions made towards aspects. The root scheme is observed to help arrive at the best results. The reason is that the model is mostly affected by the root nodes in terms of both sentiments and syntax. That is, these nodes have an impact on a wide scope. For the dependency parser module, exploiting the gold aspect labels in addition to sentiment lexicons leads to inconsistency and it causes the performance to decrease. That is using the labels present in both the training data and sentiment lexicon, which can be different for the same words, can lead to contradiction and ambiguity. Lastly, vector lengths of 30 and 50 lead to the best performances for the recursive neural network models, whereas the size 100 gives rise to overfitting.

Additionally, the results for two different scenarios, which are a single aspect (SA) or multiple aspects (MA) modules, are listed and a comparison between our approach and the baseline is made in Table~\ref{singlemultiasp}. Here, only the best accuracies (dependency parser module with the \textit{``root''} setting) for our approach are shown. SA is the case where there exists only one aspect in the review, whereas MA refers to the case where there is more than one aspect in the review. Each review in the dataset has either an SA or MA. That is, every review has at least one aspect. We demonstrate the impact of the use of the MA scenario on the performance, modelling the interplay of sentiments between different aspect terms. We arrived at better performances for the MA module. We show that we outperformed the baseline method for all cases. We performed the Stuart-Maxwell test and observed that the results we obtained are significant $p = 0.05$.

\begin{table}[!h ]

\vspace{0.9cm}\centering
\caption{Accuracies (\%) that are obtained in the baseline method (IARM) and our ensemble framework. SA stands for single aspect scenario, MA for multiple aspect scenario.}

\bigskip
\begin{tabular}{|p{5.0cm}|c|c|c|c|}
\hline
\textbf{Approach}& \multicolumn{2}{c|}{\textbf{Restaurant}} & \multicolumn{2}{c|}{\textbf{Laptop}}\\
\hline
& SA & MA & SA & MA\\
\cline{2-5}
Baseline (IARM)&78.6&80.48&73.4&74.1\\
Our ensemble framework&\textbf{79.5}&\textbf{81.38}&\textbf{75.75}&\textbf{76.45}\\
\hline
\end{tabular}

\label{singlemultiasp}
\end{table}

In summary, our incorporating a recursive neural network model into a recurrent model and forming an ensemble framework boosts the performance for ABSA. We enrich the recurrent model encoding temporal information with a recursive model capturing grammatical and sentiment information. We originally generate sub-reviews from whole reviews and train them in the recursive model. When the mentioned two neural networks are merged, a more robust representation is modelled, thereby leading to better accuracies.

\subsection{Conclusion and Future Work}

We have developed an ensemble approach to performing ternary ABSA. Our framework is composed of a recurrent and recursive sub-model. We first extract sub-reviews from reviews in a novel way. Each of these sub-reviews encompasses only one sentiment expressed towards the aspects covered therein. A single sub-sentence parse tree may include one aspect or more. We include only the relevant sentiment, ignoring other opinions expressed towards other aspects in the same review. Constituency and dependency parser modules are utilised when training our recursive neural models. In the end, the root vectors of these parse trees are fed as input into a GRU network, which is a recurrent model.

It is observed that combining recursive and recurrent neural networks provides us with an enriched and a more robust model. In the recurrent model, we capture the temporal information, whereas in the recursive model, we can extract inherent grammatical and sentimental features from the texts. That is, both these compensate for what the other module lacks. Our experimental results show that we significantly outperformed the baseline approach for two datasets. When using the dependency parser, we achieve a better performance as compared to the use of constituency parser. The intuition behind it is that dependency parsers can directly capture modifier relationships. These modifiers indicate which sentiment is expressed towards which aspect or noun. We  conjecture that our ensemble classifier model can be applied to other NLP classification tasks or computational linguistics fields.

As future work as post-doctoral research, we plan to extend this approach by (1) incorporating a CNN framework to enrich the system with the contextual representation of aspect terms, (2) enhancing the constituency parser module to effectively capture the neutral opinions as well, (3) relying on other sentiment lexicons induced by semi-supervised approaches as mentioned or use other supervised techniques to better model the polarities of words in the trees, (4) training both the recurrent and recursive neural network models at the same time to arrive at more successful results, and (5) using and adapting our ensemble approach to other languages. In addition to these, we also plan to (6) compare our approach to works based on RST and hybrid solutions that incorporate concept-based reasoning into ontology-based reasoning using deep learning models and (7) evaluate our methods on other corpora, such as the SemEval-2015 corpora, where reviews can consist of more than one sentence.

We also plan to (6) evaluate our methods on other datasets with different categories (e.g. the SemEval-2015 corpora) where reviews may consist of more than one sentence and (7) compare our performances to those of RST studies and the hybrid solutions which combine concept- and ontology-based reasoning with deep learning models.

\section{Major Contribution \#4: Generating~Word~and~Document~Embeddings
for~Sentiment~Analysis}

Polarities of words can differ based on the corpus. Generating common sentiment lexicons for a language and making use of them cannot lead successful results to be obtained for corpora of different genres. In this approach, we combine the lexical, contextual, and supervised characteristics of the text words. In this regard, the semantic definitions of words appearing in the dictionary are leveraged. The tokens surrounding a target word represent domain-specific information and supervised scores of words capture their sentiments. We observed that combining labelled information of words (i.e. positive or negative) with lexical features generated from dictionaries boosts the performance. We perform an ablative combination of contextual, dictionary-based and supervised feature sets, and create novel embeddings. Additionally, we incorporate hand-crafted features into the word2vec embeddings learnt from scratch. We generate domain-specific sentiment embeddings for two datasets that are the movie and the Twitter corpora in Turkish. When we create document embeddings and feed them as input into the SVM method, our approaches outperform the baseline studies for Turkish by a significant margin. These original models are evaluated on two English datasets of different genres as well. These also perform better than the word2vec approach. This empirically shows that our methods are cross-domain and portable to other languages \cite{ayd:19}.

Neural network models in general perform better than the classical machine learning algorithms for most classification and regression tasks, including opinion mining ~\cite{gol:16}, when fed with bulky datasets as input. In these models, dense word embeddings are employed to combat the data sparsity issue. These provide us with more \textit{``meaningful”} and robust representations. These vectors indicate how similar and close the words are to one another in the VSM.

As mentioned, in the literature, most of the studies rely on vectors, such as word2vec~\cite{mik:13}, in which only the syntactic and semantic representations of the words are taken into account. Ignoring the sentimental aspects of words can cause the tokens with opposite sentiments to be closer to each other in the space model, provided that their semantic and syntactic features are similar. 

In the Turkish language, there exist only a few studies that use sentiment information when inducing word and document vectors. In contrast to the studies conducted for English and other commonly-used languages, we employ the official Turkish dictionary in this thesis, and incorporate both supervised and unsupervised features into the model at the same time so as to produce a unified score per dimension of each word embeddings.

The main contribution of this approach is generating novel and effective word embeddings which capture semantic, syntactic and sentimental characteristics of tokens, and exploit all this information when creating word vectors. We rely on the word2vec embeddings trained on our corpora as the baseline approach as well. Apart from utilising those vectors, we additionally build a hand-crafted feature set on a review-basis and generate document embeddings. We test and evaluate these vectors on several corpora of different genres. We will empirically show that we perform better than the approaches which do not leverage the sentiment knowledge. We outperform the other works carried out on the sentiment classification task in Turkish media as well. We evaluated our original vector models on two different datasets in English as well in the end. We also outperformed the baseline models for the English language. 

\subsection{Methods Developed}

We create various word embeddings that can represent the contextual, lexical, semantic, and sentimental characteristics of words. Additionally, as a baseline, we also induce word2vec vectors of the corpus words by training this vector model on these corpora. After producing those dense vectors, we also incorporate hand-crafted features into review (document) embedding representations and conduct binary classification, as we will explain and discuss later.

\subsubsection{Corpus-Based Approach}
Contexts are informative in the sense that similar words in general occur in the same contextual windows. For instance, the word \textit{``smart''} is more probably to co-occur with the word \textit{``hard-working''} than with the word \textit{``lazy''}. The similarity may be represented semantically, syntactically, and sentimentally. In this sub-approach, we capture all these characteristics and create vectors for the corpus tokens.

First, we build a matrix whose entries correspond to the number of co-occurrences of the row and column words with regard to sliding context windows. In this matrix, diagonal entries are the numbers of sliding context windows that the corresponding row word occurs in all the reviews in the dataset. Then, we perform normalisation by dividing scores in each row by the maximal score in the same row.

\enlargethispage{-\baselineskip}

Second, we use the principal component analysis (PCA) method to decrease the number of dimensions. It extracts latent meanings from data and relies on high-order co-occurrence statistics after performing noise removal. We set the reduced column (attribute) number of the matrix as 200. We thereafter calculate the cosine similarity score between each row pair $w_i$ and $w_j$ as shown in Eq. \eqref{eq:cos} to detect the similarity of two different word (row) embeddings.

\begin{align}
	cos(w_{i}, w_{j}) = \frac{w_{i}^\intercal  w_{j}}{||w_i||  ||w_j||} \label{eq:cos} 
\end{align}

Third, all the scores in the matrix are subtracted from 1 so that we have a dissimilarity matrix. We then use this matrix as an input in the fuzzy c-means clustering algorithm. The number of clusters is chosen to be 200, as this is stated to be one of the standard values for word embeddings in the literature. After having performed clustering, the dimension \emph{i} for a corresponding token is indicative of the degree to which this token belongs to the cluster \emph{i}. The main intuition behind this method is that if two tokens are close to each other in the VSM, these tend to appear in the same or similar clusters with similar probabilities. Lastly, each word in the dataset is represented by a 200-dimensional embedding.

In addition to this method, we rely on the SVD method taking the co-occurrence matrices as input, where we compute the matrix $\boldsymbol{M}^{PPMI} = \boldsymbol{U\Sigma} V^{T}$. Here, the $\boldsymbol{U}$ is an $m \times m$ real or complex unitary matrix, $\boldsymbol{\Sigma}$ is an $m \times n$ rectangular diagonal matrix with non-negative real numbers on the diagonal, and $\boldsymbol{V^{T}}$ is an $n \times n$ real or complex unitary matrix. Positive pointwise mutual information (PPMI) score between each word is computed and the truncated SVD is performed. We take into consideration only the \textit{\textbf{U}} matrix per word. We have again set the singular value number at 200. That is, every corpus word is modelled by a 200-dimensional embedding, as shown below:

\setlength{\abovedisplayskip}{-19pt}
 \begin{align}
	\boldsymbol{w}_i = (\boldsymbol{U})_i \label{eq:w-i} 
\end{align} 

\subsubsection{Dictionary-Based Approach}

In Turkish, there are not well-established sentiment lexicons as in English. In this sub-approach, the official Turkish dictionary (TDK) is employed to compute word polarities \cite{ayd:13, ayd-2:13, ayd1:14, ayd2:14}. Although this dictionary does not include polarities specific to words, when we combine it with domain-specific sentiment scores extracted from the labelled dataset, we achieve state-of-the-art performances.

First, we build a matrix whose row entries correspond to words in the dataset and column entries are the words appearing in their dictionary definitions. Boolean approach is preferred in this model. For example, for the row word \textit{``cat''}, the entries corresponding to the column words explicitly appearing in its dictionary definition are each assigned a score of 1. The matrix entries corresponding to those column words not occurring in the dictionary definition of \textit{``cat''} have a score of 0 in the same row.

When we perform clustering on the matrix, we observe that words with similar meanings are grouped in the same or nearby clusters. Nevertheless, sentiment information cannot be effectively captured in this model. For example, the words \textit{``happy''} and \textit{``unhappy''} are grouped in the same cluster, for they have common words in their dictionary definitions, such as \textit{``feeling''}. On the contrary, words of different polarities should actually be located far away from each other in the VSM.

We therefore rely on an approach to move such words as far away from one another as possible in this space model, even if there are several common words in their lexical definitions. Each score in a row is multiplied by the raw sentiment score of the word in the same row. Hence, we obtain more robust clusters. Only exploiting the training set, the sentiment score for each word is computed as shown in Eq. \eqref{eq:w-t}. In fact, this formula is the same as Eq. \eqref{eq:deltaraw} except that only the smoothing value chosen for this case is different.

\begin{align}
                            w_{t}  &=\log\frac{\frac{N_{t}}{N}+0.01}{\frac{N'_{t}}{N'}+0.01} \label{eq:w-t}
\end{align} 

Here, $ w_{t}$ is the polarity score of the word $t$, $N_{t}$ is the total number of reviews/tweets in which $t$ appears in the corpus containing positive reviews. $N$ is the vocabulary size of the positive dataset. $N'_{t}$ and $N'$ denote the corresponding values for the corpus of negative sentiment. We normalise the values to handle the imbalance problem by adding a small number to both numerator and denominator for smoothing.

On the other hand, apart from the multiplication of the raw supervised sentiment values with unsupervised scores, we additionally multiply all row scores with only +1 if the corresponding row word expresses a positive sentiment or with -1 if this expresses a negative sentiment. It is observed that we thereby achieve better performance. 

\begin{figure}[!h] 
\begin{center}
\includegraphics[width=6.8cm]{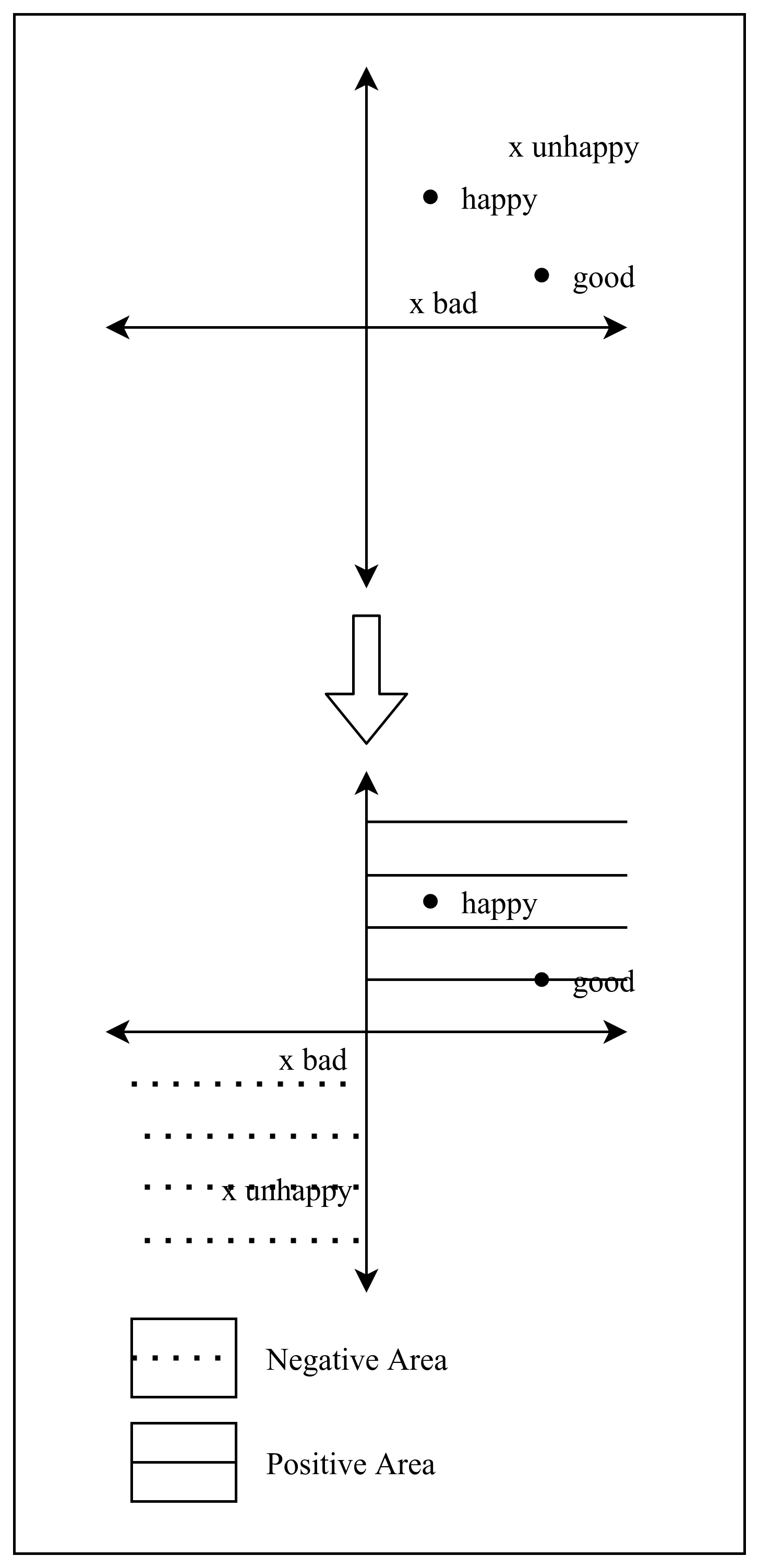} 
\caption{The impact of leveraging the labels of tokens as well in the dictionary-based approach. This figure visualises how words with the opposite polarities get far away from each other in the VSM.}
 \label{supervDict}
\end{center}
\end{figure}

\enlargethispage{-\baselineskip}
The impact of this multiplicative process is visualised in Figure~\ref{supervDict}, visualising the positions of word embeddings in the VSM. These \textit{``x''} words are of negative polarity, whereas \textit{``o''} words are of positive polarity. In the upper sub-plot, the words with different sentiments are close to each other, due to their having common dictionary definition words. Only the information in the dictionary definitions is relied on there, ignoring the supervised scores. Nevertheless, when employing the supervised score (+1 or -1), tokens with different sentiments (e.g. \textit {``happy''} and \textit{``unhappy''}) move far away from one another as they are translated across different coordinate regions. Now, words with a positive sentiment are in quadrant 1 and words with a negative sentiment are in quadrant 3. Thus, words with similar sentiment scores get closer in the VSM and could be more accurately grouped in the same or close clusters in the VSM. Apart from clustering, we used the SVD algorithm as well to conduct dimensionality reduction for the fully unsupervised dictionary approach and made use of the newly created matrix by performing ablative combinations with other sub-approaches. The number of dimensions is again set at 200 with regard to the \textit{\textbf{U}} matrix. These are elaborated in the later sections. When evaluating this sub-approach on the datasets in English, we relied on the SentiWordNet lexicon~\cite{bac:10}. When preferring the SVD reduction approach over the use of clustering method, we achieve better performances.

\subsubsection{Supervised Contextual 4-scores}
The last module we used is a basic scheme which extracts four supervised scores per token in the corpora. These values are generated as follows. For each target token in the dataset, we scan through all of its context windows. In addition to the sentiment value of a target word (which we name ``the self-score''), maximum, minimum, and averaged scores are used out of all the sentiment scores of words appearing in the same context windows as the target token. We calculated the word polarity scores by using Eq. \eqref{eq:w-t}. These scores are obtained by relying on the training set.

In this way, those four scores per word are taken into account, which is more informative compared to the use of a single self-polarity score. However, this metric is entirely supervised in contrast to the previous two sub-modules.

\subsubsection{Combination of Word Vectors}

Apart from employing these three sub-modules disjointly, all the matrices generated in the previous approaches are also combined. In other words, we perform concatenation of the reduced forms (SVD - U) of dictionary-based, corpus-based, and the 4-score vectors of each word column-wise. Accordingly, each token is modelled by a 404-dimensional embedding, for dictionary-based corpus-based embedding components have each 200 dimensions, whereas the 4-score embedding component is formed of four values, whereby the sum is 404.

The intuition behind this ensemble approach is that some sub-methods compensate for what the others lack. For instance, the corpus-based approach can capture the domain-specific, syntactic, and semantic characteristics. On the other hand, the 4-scores module models label information and the dictionary-based approach helps capture the general lexical and semantic information. When we combined these three approaches, we have produced more representative and robust word models.

 \subsubsection{Creating Document Embeddings}

After generating various vector types as mentioned, we generate document embeddings. In this respect, documents can be reviews or tweets. Per document, we average all the embeddings of those words appearing therein. Additionally, we define a set of hand-crafted features, which are maximal, minimal, and mean polarity scores of words appearing in a document. Those sentiment values are calculated as shown in \eqref{eq:w-t}. That is, these values are generated on a review-basis, not on a word-basis. For instance, if a document is composed of six words, this would have six polarity scores and only three of these polarity values are employed as explained. In the end, we concatenate the averaged document vectors to these three sentiment scores.

\begin{figure}[h]
 \begin{center}
\includegraphics[width=11.66cm]{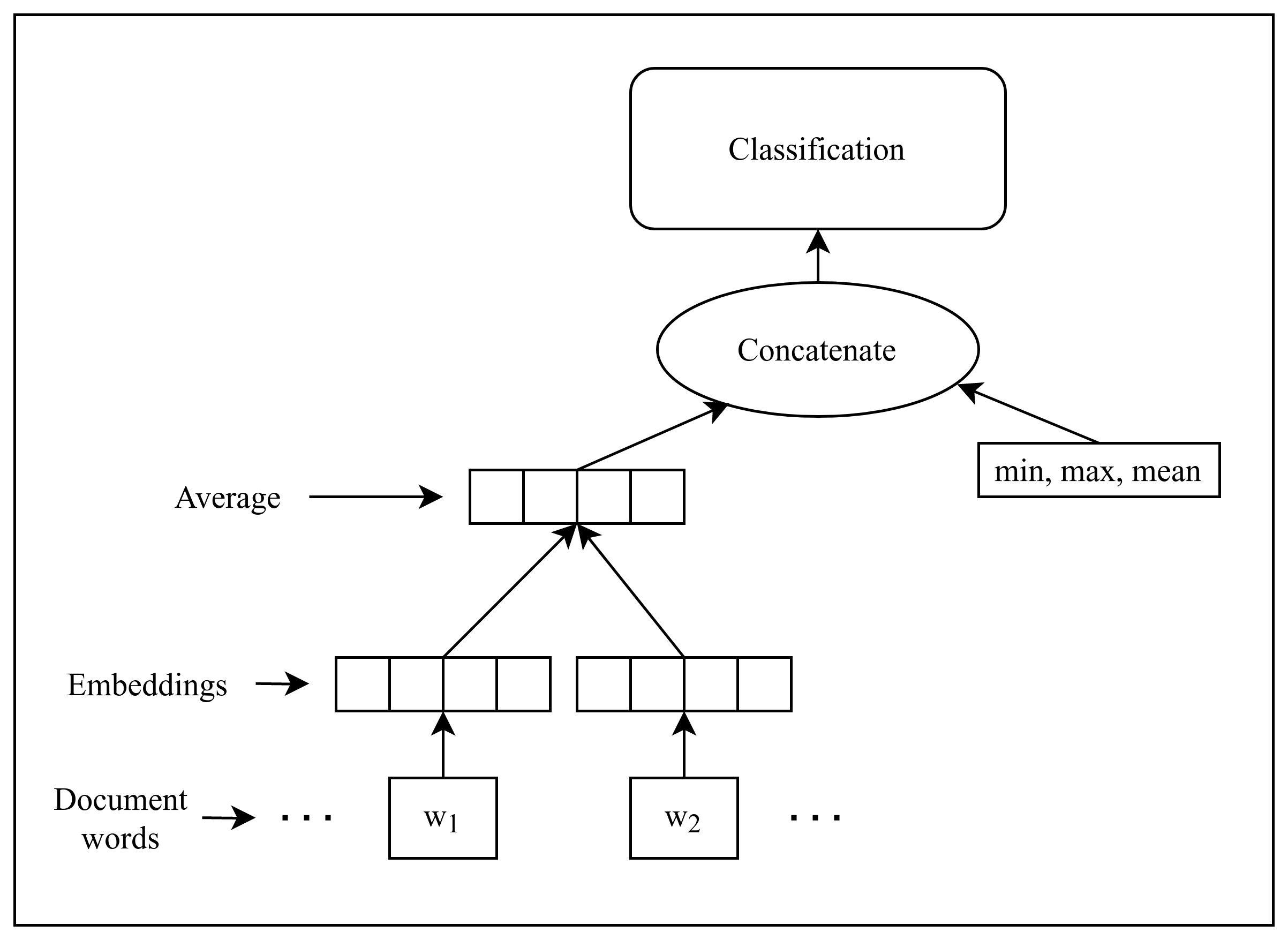} 
\caption{The flowchart of our system.}
\label{classify}
\end{center}
\end{figure}

Thus, each document is modelled by the averaged word embedding and three additional hand-crafted sentiment scores. We thereafter use these vectors in an SVM classifier to predict the polarity of a review. We give the flowchart of the proposed approach in Figure~\ref{classify}. When we combine the embeddings generated by word-based features with the overall three sentiment scores created on a review-basis, better-performing and state-of-the-art results are obtained by us.

\subsection{Corpora}

We relied on two corpora of different genres in both Turkish and English when conducting experimentations and evaluating our approaches.

As the first Turkish dataset, we employed the same movie corpus consisting of 20,244 reviews that are collected from a popular website, as mentioned in Section~\ref{sec:datasets-}. We have randomly chosen 7,020 negative and 7,020 positive reviews, formed a balanced dataset, and processed only them. The second corpus for Turkish is the same Twitter dataset consisting of 1,716 tweets, which we explained in the preceding subsections.

We also relied on two other datasets which are in English to test the portability of our approaches to other languages. Among these, the first is a balanced movie dataset \cite{pan:05} downloaded from a website \cite{bri:git-cnn}. In total, there are 5,331 negative reviews and 5,331 positive reviews in this corpus, as discussed previously. The second dataset is a Twitter corpus that consists of approximately 1.6 million tweets which are labelled through a distant supervised algorithm~\cite{go:09}. These tweets are assigned negative, positive, and neutral polarities. We chose 7,020 positive tweets and 7,020 negative tweets in a random manner to create a balanced corpus for the binary classification task.

\subsection{Experiments}
\subsubsection{Preprocessing}
We performed similar preprocessing techniques to those given and discussed in Section~\ref{sec:pre} for this approach as well. We explain these as follows. The Turkish speaking people can sometimes spell English characters for the corresponding Turkish characters (e.g. \textit{``i''} for \textit{``\i''}, \textit{``c''} for \textit{``\c{c}''}) when typing messages in an electronic format. So as to perform normalisation on these tokens, the Zemberek tool \cite{aki:07} is employed. Punctuation marks except only \textit{``!''} and \textit{``?''} are filtered out, since these do not have an effect on the overall sentiment of a review/tweet. In this approach, emoticons (e.g. \textit{``:(((''}), and idioms (e.g.\textit{``kafay{\i} yemek''} (lose one's mind)), have not been removed, since two or more words can bear a sentiment together when forming an idiom, no matter what polarity its constituent words have. Due to the morphologically-rich structure of the Turkish language, the morphological parser and disambiguation tools \cite{sak:07,sak:08} are used. We additionally performed stop word removal and negation handling. When handling negator words/morphemes, an underscore is added to the end of a word in case negation occurs. For instance, \textit{``sıkıcı değil''} (not boring) is redefined as \textit{``sıkıcı\_''} (sıkıcı\_) in the feature selection stage before generating supervised scores.

\subsubsection{Hyper-parameters}

We employed the LibSVM utility in the WEKA tool \cite{hal:09}. We relied on the linear kernel for the binary classification task. We generated word2vec vectors by training them on all the four datasets making use of the Gensim library~\cite{rad:10} with the skip-gram method as the baseline. The length of these embeddings is set as 200 for a consistent comparison. As mentioned, other vectors, generated by relying on the clustering and the SVD modules, are of length 200 as well. In the c-means clustering sub-method, the maximal number of iterations is set at 25. If the model converges before reaching that number, we stop the training earlier.

\subsubsection{Results}

As mentioned, we evaluated all our approaches on four corpora that are the movie and Twitter datasets in Turkish and English. We have learnt all vectors from scratch for these datasets. As the performance criterion, the accuracy metric is preferred, since all the corpora are completely or almost balanced. 10-fold cross-validation is performed for all the corpora. We employed the approximate randomisation method to conduct the significance tests. In this approach, we try to predict the labels of reviews. That is, we evaluate and assess our models on a review-basis, not on a word-basis.

The results we obtained are shown in Table~\ref{SW}. The \textit{``3-feats''} metric denotes the hand-crafted feature set, which are the maximum, minimum, and mean polarity scores of documents as previously described. As can be seen, at least one of our approaches performs better the baseline approach (word2vec) for all the English and Turkish datasets of different categories. As can be expected, all of our approaches also perform better when we exploit the sentimental characteristics that are extracted on a review-basis and when we concatenated them to word embeddings. In most cases, the supervised 4-scores metric produces the best-performing results, since it makes use of the annotation knowledge based on the sentiments of words.

\begin{table}[t!]
\caption{\label{SW} Performances (\%) for different features used as input in the SVM model when detecting the polarities of documents. The baseline approach is chosen as the word2vec algorithm.}
\bigskip
\begin{center}
\begin{tabular}{|l|r|r|r|P{1.3cm}|}
\hline 
\multirow{2}{*}{\bf{Word vector type}} & \multicolumn{2}{c}{\bf Turkish (\%)} & \multicolumn{2}{c|}{\bf English (\%)} \\ \cline{2-5}
&\bf{Movie}&\bf{Twitter} &\bf{Movie}&\bf{Twitter}\\
\hline
Corpus-based + SVD (U) & 76.19 & 64.38&66.54 &\textbf{87.17}\\
Dictionary-based + SVD (U) & 60.64 & 51.36&55.29 &60.00\\
Supervised 4-scores & \textbf{89.38} & \textbf{76.00}&\textbf{75.65}&72.62 \\
Concatenation of the above three & 88.12 & 73.23&73.40&73.12 \\
Corpus-based + Clustering & 52.27 & 52.73&51.02 &54.40\\
word2vec & 76.47  & 46.57&57.73&62.60 \\
\hline
Corpus-based + SVD (U) + 3-feats& 88.45 & 72.60&76.85 &\textbf{85.88}\\
Dictionary-based + SVD (U) + 3-feats & 88.64 & 71.91&76.66 &80.40\\
Supervised 4-scores + 3-feats& \textbf{90.38} & \textbf{78.00}&\textbf{77.05}&72.83 \\
Concatenation of the above three + 3-feats & 89.77 & 72.60&77.03&80.20 \\
Corpus-based + Clustering + 3-feats & 87.89 & 71.91&75.02 &74.40\\
word2vec+ 3-feats & 88.88 & 71.23&77.03&75.64 \\\hline
\end{tabular}
\end{center}
\end{table}

As shown in the table, the clustering approach generally leads to the lowest performance.  We observed that the corpus - SVD module always performs better as compared to the clustering method. We think the reason is that, in SVD, we take into consideration the most important and informative singular values. The corpus - SVD method performs better than the word2vec model for several corpora. When the 3-feats metric is not employed, the corpus-based SVD approach produces the best-performing accuracies for the Twitter corpus in English. This has also empirically been shown that the use of basic and simple models can be a better and more successful choice compared to that of complex models. For instance, our statistical corpus-based approach outperforms the word2vec neural network model. We also found out that the accuracy decreases in several cases when we make use of the sentiment information, as in the case for the English tweets.

Since the official Turkish dictionary covers almost all the movie corpus words, the dictionary approach produces state-of-the-art results. On the other hand, the dictionary lacks many words, which appear in short and noisy tweets. Hence, its performance is not the best among all results. However, as can be expected, when we combine the dictionary method with the 3-feats metric, we observe a great improvement.

The accuracies produced for the movie corpora are much higher than those for the Twitter datasets for almost all of our methods. The reason is that the texts in the Twitter datasets are usually much noisier and shorter. We observed that when setting the \textit{p} value as 0.05, our results are statistically significant as compared to the baseline approach for Turkish \cite{ert:17}. Our several approaches produce better success rates than those sentiment analysis models relied on for English \cite{fel:17,tan:14} as well. That is, our results are both significant and state-of-the-art for both these languages. As said, when we additionally leveraged three hand-crafted sentiment features, we arrived at the best results.

A CNN model is also used for this approach. Nonetheless, the SVM method that is a classical machine learning model performed as effectively. Therefore, here, we cover only the best results, ignoring those obtained for the CNN approach.

As a qualitative assessment of word representations, we have listed the most similar words to target words by exploiting the cosine similarity metric. We have performed a limited analysis to find out which words are the most similar with regard to different modules. In Table~\ref{MSW}, we show the most akin words to the given query words. Tokens of the same polarity are observed to be the most similar and closest in the VSM. For instance, the most similar word to \textit{``muhteşem''} (gorgeous) is \textit{``10/10''}, both of which are of positive polarity. As shown in the table, our corpus-based approach is better at capturing domain-specific characteristics and features as compared to the word2vec model, which can generally model syntactic and semantic characteristics, however, not the sentiment information.

\begin{table}[t!]
\caption{Most similar words to given queries with regard to the corpus-based approach and the baseline algorithm (word2vec).}
\bigskip
\begin{center}
\begin{tabular}{|l|c|c|c|}
\hline 
\bf{Query Word} & \bf{Corpus-based} & \bf{word2vec}\\

    \hline
Muhteşem (\textit{Gorgeous})& 10/10 & Harika (\textit{Wonderful})\\
    \hline
Berbat (\textit{Terrible})& Vasat (\textit{Mediocre}) & Kötü (\textit{Bad})\\
    \hline
Fark (\textit{Difference})& İlginç (\textit{Interesting}) & Tespit (\textit{Finding})\\
    \hline
Kötü (\textit{Bad})& Sıkıcı (\textit{Boring}) & İyi (\textit{Good})\\
    \hline
İyi (\textit{Good})& Güzel (\textit{Beautiful}) & Kötü (\textit{Bad})\\
    \hline
Senaryo (\textit{Script})& Kurgu (\textit{Plot}) & Kurgu (\textit{Plot})\\
  
\hline
\end{tabular}
\end{center}
  \label{MSW}
\end{table}

\subsection{Conclusion}

In this approach, we have demonstrated that employing word embeddings which can capture only syntactic and semantic features and characteristics can be enhanced by incorporating sentimental aspects as well into the model. Our approach is cross-domain and portable to other languages. We conjecture that they can be applied to other datasets of different genres and also to other languages apart from Turkish and English, when minor changes are performed.

This approach is one of the few studies performing opinion mining for the Turkish language and exploiting sentiment characteristics of words when creating word embeddings. This approach outperforms all the other methods. Any of the sub-modules we proposed in this approach can be used disjointly and independently of the others. Our approach can be applied to other classification fields, such as concept extraction and topic classification, when supervised labels are adapted to new categories.

We have empirically shown that even methods based on unsupervised models, such as the corpus-based approach, may perform better than supervised models in classification tasks. Combining several approaches, each of which may compensate for what others lack, can lead to the generation of more robust word representations. Our word embeddings are built and assessed by feeding them into conventional machine learning algorithms. However, these produce state-of-the-art results. Although we used a conventional machine learning algorithm (SVM) over a neural network classifier in order to identify the polarities of documents, we have obtained accuracies of over 90\% for the Turkish movie dataset and an accuracy of about 88\% for the English Twitter dataset at best.

In this approach, we have only performed binary sentiment analysis as in the literature. We plan to (1) expand our framework in future by also taking neutral reviews into consideration to test its generalisability to all sentiment categories. We would also (2) employ the Turkish WordNet to enhance the robustness of the word and document vectors.

\section{Major Contribution \#5: Redefining Context Windows as Subclauses for Word Vector Models}

In this approach, we propose an approach, where a word context is defined in terms of a subclause rather than a sliding window. We generate subclauses by using a dependency parser. The rule set we defined for generating subclauses/sub-sentences is novel. As a sentence is broken down into subclauses, these are chosen as the context windows for the target words in the sentence. We thereby take into account linguistic, syntactic, and even sentimental patterns as opposed to sliding windows, which cannot capture these as robustly and effectively. In this way, we employ more semantically-oriented, variable-size contexts as compared to fixed-size relative position-based contexts. To evaluate the effectiveness of the proposed approach, we generate word embeddings by redefining the context windows for the GloVe and SVD models and perform binary classification by using the document vectors. We applied this approach only on English datasets. However, this can be adopted for other languages with minor changes as well, if a dependency parser library exists for the target language.

\subsection{The Proposed Approach}

In this approach, we change the local context window definition for both the GloVe and SVD models, and redefine these contexts as formed of subclauses. In this regard, we parse a sentence into subclauses as in \cite{ayd:20}. As previously mentioned in the ensemble neural network model, subclauses are also identified using the same rule set for this approach. We explain this dependency-based subclause generation process as follows (albeit a bit repetitively). We first generate dependency tree of sentences using the spaCy library. Then we scan the nodes in these trees starting from the root. We iterate recursively every child node of words. If the POS tag of a word is verb and this word is not modified by a clausal component (i.e. ccomp) or a conjunction relationship (i.e. conj), we recursively add its children into the same group marked as a subclause. Otherwise, we cut the dependency tree at these points and start forming different subclauses. After generating these subclauses, if any, we remove the preceding and trailing conjunctions. We also add the punctuation mark of the sentence to the end of each subclause. An example was shown in Figure~\ref{subrevs} that is relevant to this process. In that sample review, the sentence is broken down into three subclauses. Each subclause (subtree) is considered a separate context and is used as the context window for every word within the subtree when generating the embedding of that target word. Even those neighbouring words in the sentence could be considered semantically far away from each other, if they are in different subclauses. That is, those words in the same subclauses are more related to each other and this information is useful when generating word embedding models.

For the GloVe algorithm, we consider the context of a word as formed of the subclause in which the word occurs. For the SVD algorithm, the co-occurrence matrix is built by considering the subclauses separately. In other words, two words that co-occur in the same subclause add one to their co-occurrence count. After the co-occurrence matrix is factorised as $\boldsymbol{U\Sigma} V$, we take the $\boldsymbol{U}$ matrix only, since it is reported to yield the best performance \cite{ham:16}. In this matrix, each row corresponds to a word's representation.

After forming context windows as subclauses for both of those word embedding models and generating word vectors, we evaluate the effectiveness of the embeddings on two document-based classification tasks. We represent a document as the average of the embedding vectors of the words in the document. These document vectors are then fed as input into an SVM classifier.

\subsection{Experimental setup}

In the literature, lexical similarity and analogy tasks are heavily leveraged for assessing the quality of word representations \cite{lin:16}. In this approach, we follow a different track and perform extrinsic evaluation on two tasks to measure the quality of the word embeddings based on subclausal contexts. Table~\ref{hyper} lists the hyper-parameters used for the Glove and SVD models. Context window type can be regarded as a hyper-parameter, where sliding window refers to the baseline approach for both the GloVe and SVD methods, and subclause to the proposed approach. Subclause generation value indicates what relationships in the dependency trees are used to create subclauses. Sliding window size and sliding window position are related to only the baseline approach. We did not take into account the left-side contexts alone, since, in a baseline study \cite{lis:17}, the best-performing practices are reported to be with symmetric and right-side windows. However, this can differ depending on the linguistics properties of the language of corpus. The parameters which are word embedding size and stop word removal are applicable for both the baseline and the proposed approaches in this thesis.

\begin{table}[thbp]
  \caption{Possible values chosen as hyper-parameters for the experiments.}
\bigskip
\begin{minipage}{\textwidth}
\begin{center}

 \begin{tabular}{| p{5.5cm} | c|}
   \hline
   \textbf{Hyper-parameter} & \textbf{Possible values}\\
 \hline
  Context window types&\{subclauses, sliding windows\}\\
  \hline 
Sliding window size&\{5, 10\}\\
\hline
  Sliding window position &\{right, symmetric\}\\ 
\hline
  Subclause generation&\{conj, conj+ccomp\}\\
\hline
  Word embedding size&\{100, 300\}\\
\hline
  Stop word removal&\{yes, no\}\\
   \hline
  
  \end{tabular}
\end{center}

  \end{minipage}
  \label{hyper}
\end{table}

We train the GloVe and SVD models with 100- and 300-dimensional vector settings separately. We use two versions of each model, one with all the words included and the other with stop words removed. The epoch number for the GloVe algorithm is set as 10.

\subsection{Datasets}

We evaluated the performance of our models on two datasets corresponding to different classification tasks after training the word vectors on these corpora. The first dataset is a movie corpus labelled for binary (positive and negative) sentiment classification \cite{pan:05}. This is the same balanced dataset consisting of 10,662 reviews, as explained in Section~\ref{sec:datasets-}. We split the whole dataset as 80\% for training (8,530 reviews and 179,674 tokens) and 20\% for testing (2,132 reviews and 44,391 tokens). In addition to the sentiment analysis problem, we also evaluated this framework on a dataset that belongs to another NLP classification task to test its portability across various domains. This second one is a nearly balanced dataset used for the spam e-mail binary classification problem \cite{spam-data}. The training set consists of 5,729 e-mails (797,892 tokens) and the test set is composed of 1,708 e-mails (567,903 tokens). We use the spaCy library to tokenise words. We use the same library to generate dependency parse trees and to obtain the subclauses for each review/e-mail. We perform lemmatisation and filter out words whose frequency is lower than four. 

\subsection{Results}

In the experiments, we compare the performance of the proposed approach that makes use of subclauses as contexts with that of the baseline approach. Additionally, we perform an ablation study to assess the impact of each hyper-parameter on the performance. In this regard, we observe the success of the proposed approach with respect to different scenarios. All the experiments were performed in Python and C++ on an Ubuntu Linux-based laptop, with the Intel i5 processor and a 4 GB of RAM. The training phase lasted for 32 minutes and 5 minutes on the average for the spam and sentiment tasks, respectively. The execution of the subclause extraction module took about one and a half times as long, since it takes time for the dependency parser to generate all relationships. We show the results in Table~\ref{results} in terms of accuracy. 

\begin{table*}[ht]
\centering
  \caption{Accuracy (\%) across all models and metrics with various hyper-parameters trained on different corpora.}
\bigskip
 \begin{tabular}{|P{2.39cm}|P{1.61cm}|P{0.6cm}|P{2.49cm}P{0.9cm}|P{0.9cm}|P{0.9cm}|P{0.9cm}|P{0.9cm}|}
   \hline
   \multirow{3}{*}{\textbf{Word vector}}&\multirow{3}{*}{\textbf{Domain}}&\multirow{3}{*}{$N$}& \multicolumn{2}{c|}{\multirow{2}{*}{\textbf{Subclauses}}} & \multicolumn{4}{c|}{\textbf{Sliding windows}}\\
\cline{6-9}
 
  &&&&&\multicolumn{2}{c|}{$k=5$}&\multicolumn{2}{c|}{$k=10$}\\
\cline{4-9}
  &&&\multicolumn{1}{c|}{conj + ccomp}&conj&right&sym&right&sym\\
\hline
  \multirow{4}{*}{\pbox{2.11cm}{\relax\ifvmode\raggedright\fi GloVe}}&\multirow{2}{*}{Sentiment}&100&\multicolumn{1}{c|}{59.56}&\textbf{60.22}&58.77&59.05&58.44&59.89\\
 &&300&\multicolumn{1}{c|}{\textbf{62.42}}&61.58&58.02&60.41&58.11&61.39\\
 \cline{3-9}
&\multirow{2}{*}{Spam}&100&\multicolumn{1}{c|}{\textbf{86.99}}&86.42&85.70&80.84&82.19&81.83\\
&&300&\multicolumn{1}{c|}{89.45}&\textbf{90.27}&83.47&86.93&81.89&81.66\\

 \hline
  \multirow{4}{*}{\pbox{2.11cm}{\relax\ifvmode\raggedright\fi SVD - U}}&\multirow{2}{*}{Sentiment}&100&\multicolumn{1}{c|}{\textbf{69.51}}&69.44&65.66&66.55&67.16&67.87\\
 &&300&\multicolumn{1}{c|}{70.45}&\textbf{74.38}&68.99&68.80&69.60&70.35\\
 \cline{3-9}
&\multirow{2}{*}{Spam}&100&\multicolumn{1}{c|}{95.60}&96.12&95.54&93.67&96.25&\textbf{96.48}\\
&&300&\multicolumn{1}{c|}{\textbf{97.24}}&97.18&96.77&95.84&96.89&97.12\\

   \hline

  \end{tabular}
  \label{results}
\end{table*}

In the table, the value $N$ stands for the embedding size. The value $k$ denotes the window length. $sym$ stands for symmetric context windows. $conj$ indicates that subclauses are generated by cutting the sentence dependency tree based on only the conjunction relationship. $conj+ccomp$ denotes the same scenario based on both the conjunction and clausal component relationships. The above experiments correspond to the scenarios, where stop word removal is not performed. As can be seen, our approach (defining subclauses as context windows) outperforms most of the baseline sliding windows modules. We performed McNemar's significance test and found out that our results are significant ($p < 0.10$) except only one case where we could not outperform the baseline (SVD, Spam detection, $N=100$). We conjecture that our approaches can perform significantly well on larger corpora as well, which should be verified by future research. The results of the experiments and evaluations are detailed below for each hyper-parameter.

\subsubsection{Context Window Type}

The proposed sub-approach where context windows are formed of subclauses outperforms the baseline sliding windows method in almost all cases. This shows that subclauses can semantically and syntactically capture the relationships between the words therein more robustly and successfully as compared to fixed-size windows for word vector models. We observed that, in some cases, also taking into account the \textit{``comp''} relationship in addition to the \textit{``conj''} relationship when splitting the trees into subclauses can boost the performance or vice versa. We attribute it to the following factors. When generating narrower, that is, more specific and separate contexts for both these associations, we focus on a different aspect of the sentence. Therefore, those words in these narrower contexts can be more strictly related to each other. On the other hand, the \textit{``comp''} relationship (e.g. when immediately followed by \textit{``which''}) can specify an attribute/event about the preceding words by modifying them in broader contexts. Hence, these can also be considered related to each other and redefining them as a whole subclause can be considered a better approach in a way.
\subsubsection{Window Size and Orientation}

These hyper-parameters are related to only the baseline context window method. In regard to the sizes of sliding windows, the window size of 10 performs better than the size of 5 as can be expected. The reason might be that a wider context of words can extract lexically and semantically more informative analogies and relationships between words. For the window position parameter, we observed that using context from both sides outperforms the case where only the right context is used. This can be attributed to the fact that words around a target word in both relative positions carry more relevant information about the target word than the words on one side only. Making use of all these words can therefore model the vector representation of a target word more successfully.

\subsubsection{Word Embedding Size}

A larger size of a vector can encode more information using distributional semantics. In parallel with this, we observed an increase in the performance when using 300-dimensional embeddings as compared to 100-dimensional vectors.

\subsubsection{Stop Word Removal}

When we filtered out stop words from the corpora, we observed a decrease in the performance. This is the opposite of the results given in the baseline paper \cite{lis:17}, where removing stop words was reported to boost the performance. We attribute the reason to two factors. The first is that we employ a dependency parser and the parser needs to take into account all the words during parsing. Therefore, every token has a functionality in generating a consistent parse and subclauses. The second is that our corpora are much smaller in size compared to those used in the baseline paper. Hence, stop words add to the sizes of the datasets and are likely to contribute to the informative relationships between words. However, we do not include the effect of stop word removal in the table. Instead, we show the results, where stop words are not filtered out, since this is the best-performing scenario.

\subsection{Conclusion}

In this approach, we showed that using subclauses as context windows in word vector models outperforms the sliding window approach. Relying on linguistic patterns and capturing semantic, syntactic, and sentimental relationships between words within subclauses provides us with a more robust word representation. Our approach is cross-domain and cross-model.

As future work, we plan to extend this study to non-English corpora, to the word2vec algorithm, and to other machine learning models as the context windows used in convolutional neural networks and other similar architectures. We also plan to perform error analysis, conduct lexical similarity tasks, use rhetorical structure theory \cite{zel:19} in lieu of parsers, and leverage larger corpora to generate more powerful representations.

\section{Minor Contributions}

\subsection{Minor Contribution \#1: Semi-Supervised Approach with Tweaked Parameters}

As described and discussed in Section~\ref{sec:dom-spec}, we employed the semi-supervised approach for Turkish and English datasets similarly as in \cite{ham:16}. Accordingly, we propagate the polarities across the graph built by relying on co-occurrence statistics. However, as mentioned, we changed the formula such that we take into account factorial values in matrix and vector processes instead of the exponential or a linearly increasing function. Since this has been discussed in a previous section, we do not delve into its details again here. As said, this increased the success rates significantly. Nonetheless, since we have not made a significant contribution besides changing the formula parameters for the semi-supervised approach, we consider it to be a minor contribution.

\subsection{Minor Contribution \#2: Aspect Term Extraction for English}

Another minor contribution made by us is a method that extracts aspects from English corpora employing a supervised technique. We rely on SemEval datasets, in which aspects are manually annotated and are provided in the training set. Approach we developed is as follows. First, we perform feature extraction by defining a set of rules. We extract nouns and noun phrases, and give more importance to these words, because mostly nouns can be a candidate aspect. Among other features are GloVe vectors, POS tag vectors of words, and dependency relationship vectors. That is, for each noun or noun phrase word(s), we take into account these embeddings and concatenate them, and finally feed these vectors as input into an SVM classifier, which predicts a noun/noun phrase as an aspect or a non-aspect word. If this is a noun phrase, we average all the corresponding vectors therein. GloVe vectors are indicative of a word's semantic properties and aspects can generally be considered to have similar dense representations. POS tag vectors are also important, since, as said, words that are noun or noun-like tokens are more likely to be an aspect. Lastly, dependency relationships model the scenario, wherein aspects are more likely to be modified by specific grammatical and syntactic associations. For example, adjectives and verbs in general modify candidate aspects, as in \textit{``I loved its sound quality!''} and \textit{``This smartphone has a good resolution.''} However, these features alone are not very novel besides concatenating them all. On the other hand, we have not achieved significantly high success rates as compared to the baseline method. Therefore, we do not include the experimental results related to this approach in this thesis book. 

As future work, we plan to define a broader set of features and feed them as input into a neural network architecture, which is a more effective model than the SVM classifier for most cases. Also, we would employ dense POS tag vectors in lieu of the Boolean POS tag embeddings, since the latter cannot capture the part-of-speech similarities and properties as effectively. For instance, a gerund verb and an infinitive verb have similar representations according to the dense model, whereas they are completely different when the Boolean metric is relied on.

\subsection{Minor Contribution \#3: Aspect-Based Sentiment Classification for Turkish}

In this approach, we perform aspect-based analysis on Turkish raw text data. We first extract nouns using the parsing and disambiguation tools, and consider them all to be aspects. Then, we employ a dependency parser developed for Turkish \cite{tor:14} and if we detect that an adjective or a verb modifies them, we assign the sentiment of that word to that candidate aspect. However, this approach is very similar to that in \cite{deh:15}. Our minor contributions in this respect are as follows. First, polarity lexicon is generated beforehand by using the domain-specific semi-supervised approach discussed in Section~\ref{sec:dom-spec}. This is a much more sensible alternative compared to the use of a common/generic polarity lexicon. Second, we take into account intensifiers, downtoners, and negators. For instance, when the word \textit{``most''} occurs in the text before an adjective, we increase the strength of its sentiment and assign that score to the modified aspect/noun. So as to evaluate the performance of this approach, we annotated a small set of the movie dataset as aspects and their sentiments. We achieved a success rate of around 70\% for this small set. However, since we have not performed a comprehensive analysis for the ABSA in Turkish, we do not include the experimental results here.

As future work, we plan to employ a common corpus and filter out some words as non-aspect if they appear in both the target dataset and this generic corpus. The reason is that some aspects are specific to a domain and those words occurring in generic sentences very frequently can be weighed less as aspect candidates. 

\chapter{CONCLUSIONS AND FUTURE WORK}
\label{chapter:conclusion}

\begin{figure}[h]
 
 \vspace{0.5cm} 
\includegraphics[width=\linewidth]{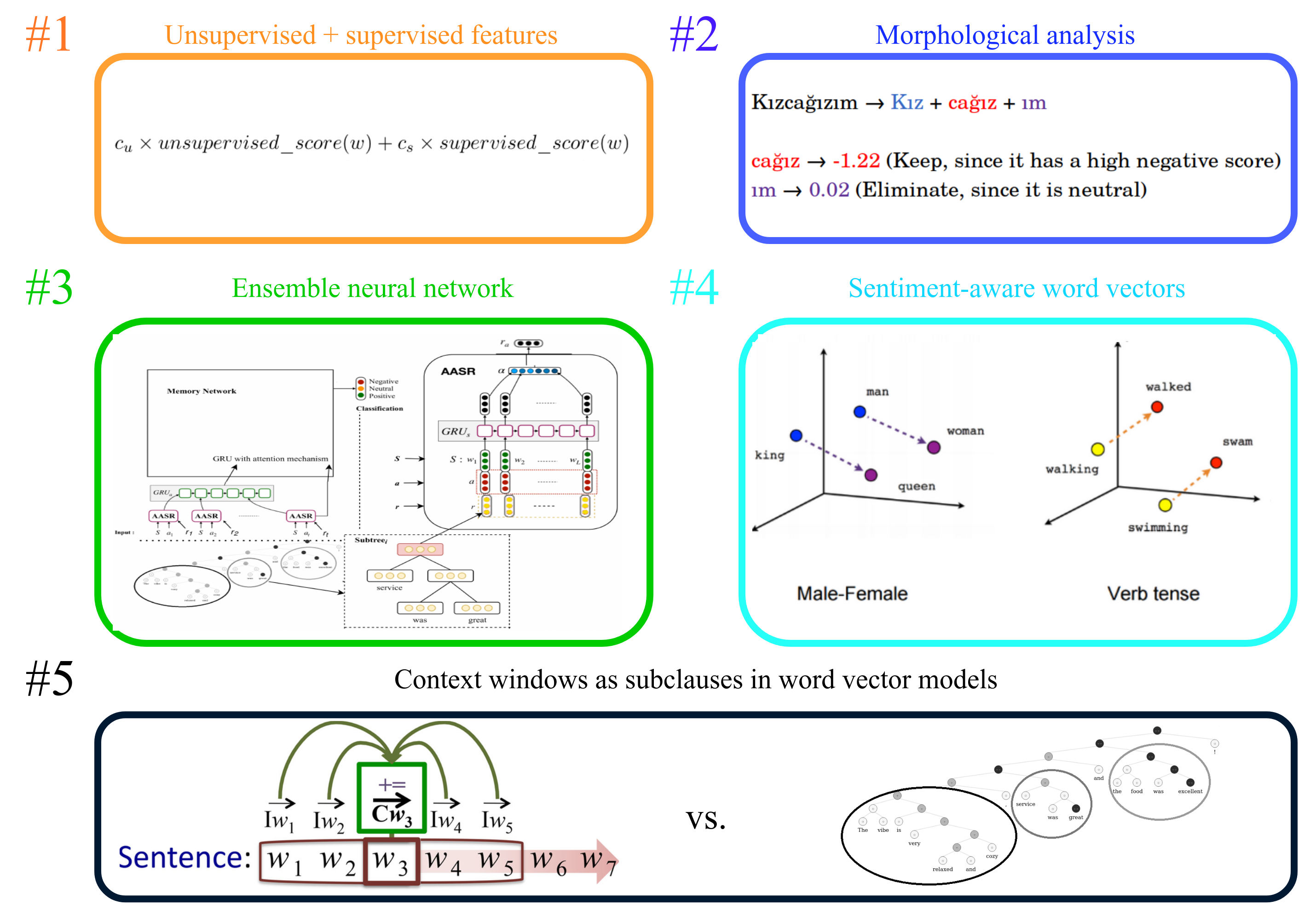} 
\caption{Visual summaries of the five major contributions.}
 \label{5-major-contr}
\end{figure}

In this thesis, we have made five major and three minor contributions on the sentiment classification task in Turkish and English, which are visually summarised in Figure~\ref{5-major-contr}. We addressed several aspects of and issues about sentiment analysis that exist for both Turkish and English. As discussed in the above separate sections, we elaborated on all our approaches. However, we summarise them all here as well as the final chapter, albeit a bit repetitively.

\begin{enumerate}
\item \textit{Major contribution \#1:} We have generated effective and novel feature engineering techniques by combining unsupervised, semi-supervised, and supervised metrics. Our research question was whether taking unsupervised characteristics into account in addition to supervised features in an ensemble form could contribute to the representations of words. After generating a polarity lexicon on a word-basis, we employed several classical machine learning models taking these scalar word sentiment scores as inputs. We thereby outperformed neural network models for several corpora in English and Turkish. That shows that classical machine learning methods still have the potential to compete with neural network models, when employing robust feature engineering techniques. This model is cross-domain and is also portable to other languages with all results being significant. 

\item \textit{Major contribution \#2:} We have performed a fine morphological analysis specific to Turkish, an agglutinative language, for sentiment analysis. In the literature, mostly only root forms of words are employed and their polarities are computed when conducting opinion mining. In our work, we have originally specified a set of rules and applied them when assigning a sentiment score to each morpheme attached to the root form of a word. We have observed that some suffixes are more expressive of sentiments and filtering out these morphemes that are not sentimentally discriminative boosts the performance for Turkish. We employed two corpora in this language to build a generic and generalisable morpheme polarity lexicon. This approach is proved to be cross-domain with significant performances. We conjecture that this approach can also be applied to the sentiment classification task for other morphologically-rich languages and even to other NLP classification tasks. The source code for this and the above approach is publicly available\cite{ayd:git-sent-an}.

\item \textit{Major contribution \#3:} Related works on ABSA in English mostly rely on either recurrent or recursive neural network models; however, only a few of them merge these two architectures and generate an ensemble classifier. In this approach, we combine recursive and recurrent neural networks in an original way by improving a model described in a baseline study. In this baseline study, the researchers only employ a recurrent neural network model exploiting the inter-aspect relations. They model the propagation effect of sentiments on aspects throughout the text. We enrich this system by incorporating a recursive neural network architecture. We first extract sub-reviews/subclauses in a novel way for each aspect in a review. Then, we generate a recursive parse tree for each aspect and train them in a distant-supervision manner using a dependency and constituency parser. Thereafter, we feed the root vectors of these trees as input into the recurrent model (baseline) per aspect. We thereby take into account temporal and grammatical characteristics at the same time in this original ensemble neural model architecture. Temporal information is modelled by the recurrent model, whereas grammatical structures represent the sentiment information more robustly. We again achieve state-of-the-art and significant results for this approach for two datasets of different genres in English. We rank first and second in a worldwide competition for both these corpora. The corresponding source code for this approach is publicly available as well \cite{ayd:git-ensemble}.

\item \textit{Major contribution \#4:} In the literature, word embeddings in general can ignore the sentiment information, and exploit only the semantic and syntactic information. Accordingly, we addressed this issue by generating word vectors that also take into account sentimental and lexical information in a novel way in addition to capturing syntactic and semantic knowledge. We have found out that we could even outperform the supervised word embedding generation approach by employing an unsupervised model when modelling word representations. We also performed better than a baseline model, which is the popular word2vec, a shallow neural network architecture, for almost all the scenarios, whereby we exploit the above-mentioned characteristics. We evaluated our word generation approaches on corpora of different genres in two languages, which are Turkish and English. This approach is also cross-domain and portable to other languages, with results being significant. The relevant source code for this approach is publicly available \cite{ayd:git-sent-embs}.

\item \textit{Major contribution \#5:} When generating word representation models, statistical information based on context windows is leveraged. In the literature, this contextual knowledge is in general extracted based mostly on windows of a fixed-size length. Researchers look up the words in the symmetrical or asymmetrical windows of a target token when modelling its embedding. However, this rigid approach may overlook several aspects about a word. There are only a few studies in the literature that define subclauses as context windows. However, in our approach, the rule set we develop for generating sub-sentences is novel. As compared to the use of windows with a fixed-size, when we employ subclauses as the context windows for target words, we arrive at more robust representations, since these can capture the relevant syntactic, semantic, and even sentimental information more sensibly. For example, even two neighbour words in a window with a fixed length could be considered far away from each other semantically and syntactically. We thereby addressed this issue in this approach. We evaluated our methods by training two word embedding models on two corpora in English. After generating word representation models, we averaged all the vectors of words appearing in documents and built document vectors. We then fed them as input into two NLP tasks, which are sentiment and spam classifications. We arrived at better results for almost all hyper-parameter values and cases as compared to the baseline method, which only uses fixed-size context windows. This approach is cross-model and cross-model. Despite the fact that this contribution is not specific to sentiment analysis unlike the other contributions, we conjecture this definition of context windows can also be applied to other NLP or machine learning problems, such as CNN and LSTM, with minor changes and adaptations. The corresponding source code for this approach is publicly available as well \cite{ayd:git-cont-win}.

\end{enumerate}

\textit{Minor contributions \#1, \#2, and \#3:} Our minor contributions for the sentiment analysis problem are listed as follows. (1) We tweaked the parameters of a semi-supervised approach so that we could generate more robust domain-specific polarity lexicons, which can be adapted to any domain and language with minor changes. (2) We extracted aspects from raw text in English using several feature engineering techniques that are not very much novel and cannot outperform the baseline by a significant margin. (3) We performed aspect-based sentiment analysis for Turkish that has a few novel characteristics inherent to it compared to a baseline approach, for which the source code is publicly available as well \cite{ayd:git-asp-sent-tr}.

In summary, in this thesis we addressed issues specific to not only Turkish, but also English for several aspects of the sentiment analysis problem. We generated several polarity lexicons, feature sets, original neural network models, and other several approaches that have thus far not been handled by researchers for both of these languages. We think our work has been the most detailed and comprehensive research conducted for sentiment analysis in Turkish as of July, 2020 \cite{boun-haber}. Some of our methods for the English language also have major contributions that fill the gap for this classification task by the mentioned novel aspects in this thesis. Several other approaches developed by us also boost the performance for the existing machine learning models. We have additionally built several methods, such as subclause and context window generations, which are specific to not only sentiment analysis domain, but also to generic machine learning models, linguistic models, and other NLP tasks. Our frameworks can be applied to several other agglutinative or non-agglutinative languages and datasets for sentiment analysis, and other classification problems.

As future work, in post-doctoral research, we plan to merge all these major and minor contributions into a single, more comprehensive opinion mining framework by performing minor changes. We also plan to develop a ``complete'' framework for the sentiment classification task in Turkish. In this scenario, we would also (1) take account of the time and opinion holder components as defined in the quintuple representation, which we have not been modelled in this thesis. Additionally, for all our approaches alone, we would (2) perform a more comprehensive error analysis. We also plan to (3) rely on neural networks when generating a morpheme polarity lexicon for the Turkish language, (4) apply our word context window definitions to other corpora, languages, and NLP tasks, (5) enhance our model for the aspect-based sentiment analysis in Turkish as discussed, and (6) improve our ensemble neural network framework by incorporating more robust sentiment information into the model as mentioned. When these would be implemented as well, we would make a broader contribution to the sentiment analysis domain for Turkish, other agglutinative languages when applied with minor changes, English, and even to other computational linguistics and NLP tasks.

\section{Publications Obtained in this Thesis}

We have three scientific papers published during the thesis stage, one of which is a long conference paper \cite{ayd:19} and two of which are journal papers \cite{ayd:20, ayd2:20}. In addition, we have submitted one of our approaches to a conference, from which the notification of acceptance is yet to be made. These publications and papers are as follows:

\begin{enumerate}
\item \textit{IEEE Access} 

Aydın, C. R. and Güngör, T., ``Combination of Recursive and Recurrent Neural Networks for Aspect-Based Sentiment Analysis Using Inter-Aspect Relations,'' in IEEE Access, vol. 8, pp. 77820-77832, 2020, doi: 10.1109/ACCESS.2020.2990306.

\item \textit{CICLing 2019}

Aydın, C. R., Güngör, T., and Erkan, A., ``Generating Word and Document Embeddings for Sentiment Analysis'', 16th International Conference on Intelligent Text Processing and Computational Linguistics (CICLing 2019), Ed. A.Gelbukh, April 2019, La Rochelle, France.

\item \textit{Natural Language Engineering (\textit{Cambridge University Press})}

Aydın, C. R. and Güngör, T. (2020). ``Sentiment Analysis in Turkish: Supervised, Semi-Supervised, and Unsupervised Techniques", Natural Language Engineering, 1-29. doi:10.1017/S1351324920000200

\item \textit{EMNLP 2020}

Submitted; the notification of acceptance is yet to be announced. (A short paper entitled ``Redefining Context Windows as Subclauses for Word Vector Models'')
\end{enumerate}

 The \textit{``IEEE Access''} journal in which our paper has been published is the third best journal in the field of Engineering \& Computer Science according to Google Scholar as of July, 2020 \cite{gs:gen}. On the other hand, the conference \textit{``CICLing''} and the journal \textit{``Natural Language Engineering''} that our works have been accepted for and published in are both among the top 20 publications on computational linguistics and NLP domains according to Google Scholar as of July, 2020 \cite{gs:nlp}. We have also submitted one short paper to EMNLP 2020, the second best publication in the field of computational linguistics and NLP according to the same Google Scholar URL reference given above, from which the notification of acceptance is yet to be made.

\bibliographystyle{styles/fbe_tez_v11}
\bibliography{references}
\end{document}